\documentclass[twoside,11pt]{article}

\usepackage{blindtext}

\usepackage{jmlr2e}
\usepackage{graphicx}
\usepackage{enumerate}
\usepackage{natbib}
\usepackage{bm}
\usepackage{mathrsfs}
\usepackage{amsmath}
\usepackage{amssymb}
\usepackage{amsfonts} 
\usepackage{chapterbib}
\usepackage{dutchcal}
\usepackage{enumerate}
\usepackage{graphicx}
\usepackage{subcaption}
\usepackage{booktabs}
\usepackage{threeparttable}

\usepackage{xcolor}
\usepackage{multirow} 
\usepackage{pifont}
\usepackage{color}
\usepackage{bbm}
\usepackage[thinc]{esdiff}
\usepackage{algorithm}
\usepackage{algorithmic}

\newcommand{\cJ}{\mathcal{J}}

\newcommand{\cC}{\mathcal{C}}
\newcommand{\cW}{\mathcal{W}}
\newcommand{\cM}{\mathcal{M}}
\newcommand{\cN}{\mathcal{N}}
\newcommand{\cT}{\mathcal{T}}
\newcommand{\cF}{\mathcal{F}}
\newcommand{\cG}{\mathcal{G}}
\newcommand{\sN}{\mathscr{N}}
\newcommand{\bbR}{\mathbb{R}}
\newcommand{\bbE}{\mathbb{E}}
\newcommand{\bbN}{\mathbb{N}}
\newcommand{\bbP}{\mathbb{P}}
\newcommand{\bbI}{\mathbb{I}}
\newcommand{\bbS}{\mathbb{S}}
\newcommand{\cL}{\mathscr{L}}

\newcommand{\bE}{\mathbf{E}}
\newcommand{\bA}{\mathbf{A}}
\newcommand{\bB}{\mathbf{B}}
\newcommand{\bI}{\mathbf{I}}

\newcommand{\bP}{\mathbf{P}}

\newcommand{\bC}{\mathbf{C}}

\newcommand{\bmX}{\bm{X}}
\newcommand{\bmC}{\bm{C}}
\newcommand{\bmY}{\bm{Y}}
\newcommand{\bmE}{\bm{E}}
\newcommand{\bmU}{\bm{U}}
\newcommand{\bmu}{\bm{\mu}}

\newcommand{\bSigma}{\bm{\Sigma}}

\newtheorem{assumption}{Assumption}
\newtheorem{theorem1}{Theorem}
\newtheorem{lemma1}{Lemma}
\newtheorem{corollary1}{Corollary}
\newtheorem{definition1}{Definition}
\newtheorem{remark1}{Remark}
\newtheorem{proposition1}{Proposition}

\newcommand*\namea{\textsc{Multi-AI }}
\newcommand*\nameb{\textsc{Max-AI }}
\newcommand*\nameaa{\textsc{Multi-AI}}
\newcommand*\namebb{\textsc{Max-AI}}

\definecolor{gray}{gray}{0.5} 

\usepackage{lastpage}

\ShortHeadings{Quickest Causal Change Point Detection}{Xu and Zhang}
\firstpageno{1}

\begin{document}

\title{Quickest Causal Change Point Detection by Adaptive Intervention}

\author{\name Haijie Xu \email xu-hj22@mails.tsinghua.edu.cn \\
       \addr Department of Industrial Engineering \\
       Tsinghua University \\
       Beijing, China
       \AND
       \name Chen Zhang \email zhangchen01@tsinghua.edu.cn \\
       \addr Department of Industrial Engineering \\
       Tsinghua University \\
       Beijing, China}
\editor{My editor}
\maketitle
\begin{abstract}
We propose an algorithm for change point monitoring in linear causal models that accounts for interventions. Through a special centralization technique, we can concentrate the changes arising from causal propagation across nodes into a single dimension. Additionally, by selecting appropriate intervention nodes based on Kullback-Leibler divergence, we can amplify the change magnitude. We also present an algorithm for selecting the intervention values, which aids in the identification of the most effective intervention nodes. Two monitoring methods are proposed, each with an adaptive intervention policy to make a balance between exploration and exploitation. We theoretically demonstrate the first-order optimality of the proposed methods and validate their properties using simulation datasets and two real-world case studies.
\end{abstract}


\section{Introduction}
Sequential change point detection has been widely studied in recent decades. Suppose a data sequence $\bmX^t, t=1,2,\ldots$ with $\bmX^t\triangleq (X^t_1,\dots,X^t_p)^T \in \bbR^p$ is sampled sequentially over time $t$. The goal is to determine whether the distribution of $\bmX^t$ exists changes \citep{lai1995sequential,keshavarz2020sequential,fellouris2022quickest,besson2022efficient,londschien2023random}, i.e., there exists an unknown but deterministic change point $\tau \in \bbN$ such that,
\begin{equation}
\label{eq:classical change}
    \bmX^t \overset{\text{i.i.d.}}{\sim} 
    \begin{cases}
    &f_0(\bmX^t),\quad  \text{for} \quad t = 1,2,\dots,\tau-1, \\
    &f_1(\bmX^t),\quad \text{for} \quad t = \tau,\tau+1,\dots,
    \end{cases}
\end{equation}
$f_0(\cdot)$ is the pre-changed distribution which is generally assumed to be known. $f_1(\cdot)$ is the post-change distribution which is generally unknown. At each time $t$, we observe the data $\bmX^t$ and need to determine whether a change has occurred, i.e., whether $\tau \leq t$. 

\begin{figure}[ht]
    \centering
    
    \begin{subfigure}[b]{0.15\textwidth} 
        \includegraphics[width=\textwidth]{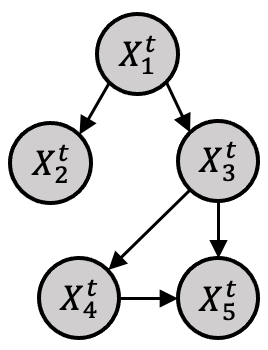}
        \caption{no change occurs}
        \label{fig:sub_1}
    \end{subfigure}
    \hfill 
    \begin{subfigure}[b]{0.15\textwidth}
        \includegraphics[width=\textwidth]{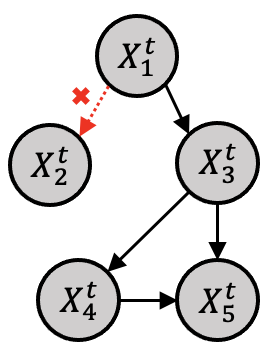}
        \caption{change pattern 1}
        \label{fig:sub_2}
    \end{subfigure}
    \hfill
    \begin{subfigure}[b]{0.15\textwidth}
        \includegraphics[width=\textwidth]{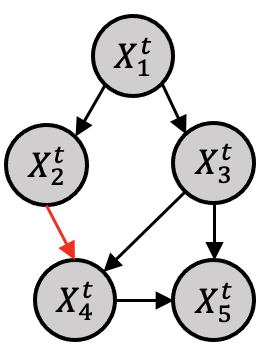}
        \caption{change pattern 2}
        \label{fig:sub_3}
    \end{subfigure}
    \hfill
    \begin{subfigure}[b]{0.15\textwidth}
        \includegraphics[width=\textwidth]{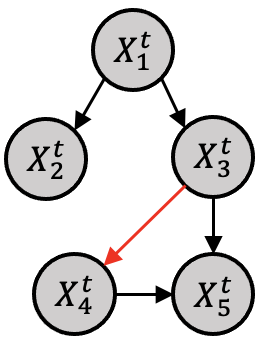}
        \caption{change pattern 3}
        \label{fig:sub_4}
    \end{subfigure}
    \hfill
    \begin{subfigure}[b]{0.15\textwidth}
        \includegraphics[width=\textwidth]{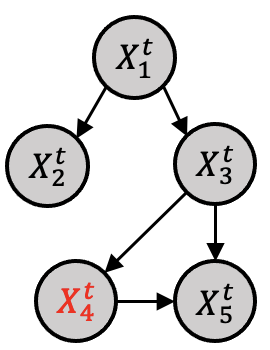}
        \caption{change pattern 4}
        \label{fig:sub_5}
    \end{subfigure}
    
    \begin{subfigure}[b]{0.95\textwidth} 
        \includegraphics[width=\textwidth]{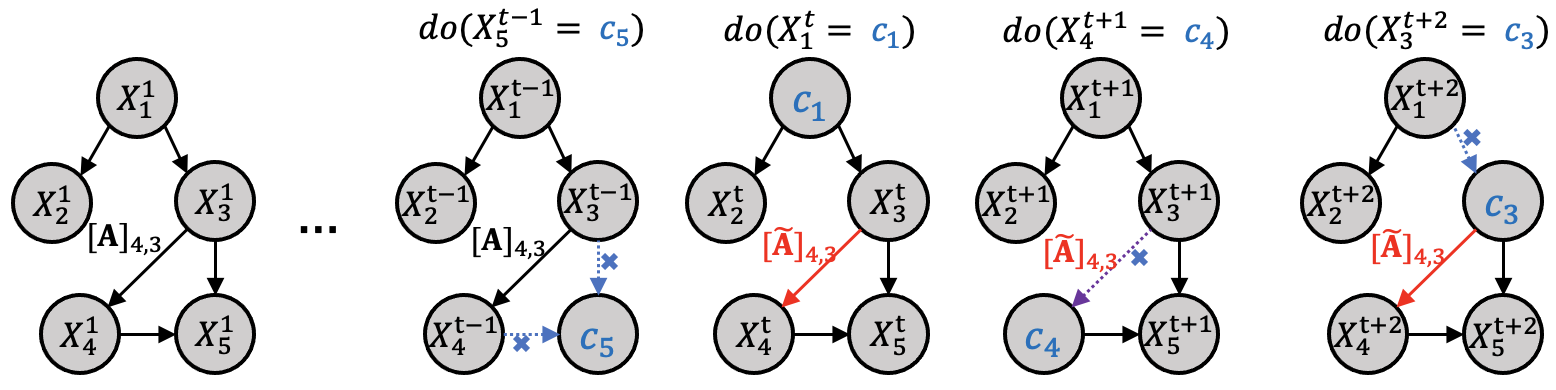}
        \caption{change occurs at $\tau = t$} 
        \label{fig:intro main} 
    \end{subfigure}
    \caption{
    (a)-(e) Different change patterns of $\cG$: (a) the pre-change causal graph, and post-change causal graphs when (b) the edge from node 1 to node 2 disappears; (c) the edge from node 2 to node 4 appears; (d) the causal effect from node 3 to node 4 changes. (e) the distribution of exogenous background variable of node 4 changes. (f) the sequential intervention: change occurs at time point $\tau = t$ and is associated with the causal effect from node $3$ to node $4$.
$c_j$ denotes the intervention value on node $j$. The blue dashed arrows represent the edges removed due to the intervention, and the purple dashed arrow indicates that the edge occurs change but is removed due to intervention. 
}
    \label{fig:combined}
\end{figure}

In this paper, we consider $\bmX^t$ to be represented through a causal graph $\mathcal{G} = (\mathcal{V},\mathcal{B})$ where $\mathcal{V}$ with $|\mathcal{V}|=p$ represents the set of nodes and $\mathcal{B}$ represents the set of edges. 
We suppose that the change influences the distribution of $\bmX^t$ by impacting the structure or parameters of the causal network, i.e., $f_0(\cdot) = f(\cdot| \mathcal{G})$ and $f_1(\cdot) = f(\cdot|\Tilde{\mathcal{G}})$. $\Tilde{\mathcal{G}}$ is different from $\mathcal{G}$ by disappearance, appearance, change in intensity of a causal effect  or change in a node as shown in Figure \ref{fig:sub_2}, Figure \ref{fig:sub_3}, Figure \ref{fig:sub_4} and Figure \ref{fig:sub_5} respectively. For example, in Figure \ref{fig:intro main}, the change occurs at time $t = \tau$, where the causal effect from node 3 to node 4 changes from $[\bA]_{4,3}$ to $[\Tilde{\bA}]_{4,3}$. 

Causal graphs have gained increasing attention and have been popularly applied to describe complex relationships between variables in biology \citep{yu2004advances}, geology \citep{luo2024causal}, social science \citep{ogburn2020causal} and so on.  Though structure learning for causal graphs has been well studied in literature \citep{huang2019causal, fujiwara2023causal, huang2020causal,glymour2019review,toth2022active}. \cite{gao2024causal,gao2024causal2} are the only works for causal change point detection. However, they assume the nodes follow discrete distributions with finite domains and also focus on off-line change detection scenarios.


In causal inference, intervention is an important tool to estimate causal relationships by actively manipulating variables. By intervening on a node, the causal effects from its ancestors to it disappear. For example, in Figure \ref{fig:intro main}, at time $t-1$, we intervene on node 5 by setting its value to $c_5$, then the causal effects from node 3 and node 4 to it disappear. 
Many studies have explored interventions for different purposes. For example, \cite{lattimore2016causal, lee2018structural,lee2019structural,varici2023causal} introduced intervention to maximize or minimize the target node's value. \cite{toth2022active,tigas2022interventions,branchini2023causal} introduced intervention to help causal discovery. However, to our best knowledge, no research has tried to integrate intervention for causal change point detection.  

Introducing interventions can improve detection power. For the previous example, if a causal effect from $X_3$ to $X_4$ has a change, we can amplify its effects by setting large intervention values on $X_{3}$ or its ancestors.
However, it is a double-edged sword, since in reality we do not know which edge has changed. If a wrong intervention node is chosen, the effect of intervention may be useless or even cause the distributions of $\bmX^t$ under pre-change and post-change conditions to become indistinguishable. For example, in Figure (\ref{fig:intro main}), at time $t+1$, intervening on node 4 will remove the causal effect with the change, causing the data distributions under both pre-change and post-change conditions to be exactly the same. Overall, introducing intervention has potential for better causal change detection, but also introduce challenges in how to select the node to intervene on and set the intervention value. 

In this paper, we propose a novel change point detection framework for causal graphs with adaptive intervention strategies, 
which can pinpoint the best node to intervene on and set its intervention value, such that the change can be magnified most, and accordingly, the alarms can be triggered quickest in the post-change condition. In particular, we summarize our contributions in the following: 
\begin{itemize}
    \item  We analyze the propagation mechanism of a change in a causal graph and the corresponding node intervention mechanism, such that after intervention, the change can be magnified most. Based on it, we propose a rule for setting the intervention value for each node. By following the rule, the optimal node to intervene on can be selected automatically.
    \item Built upon the intervention rule, we propose two detection statistics and their corresponding adaptive intervention policies, i.e., \namea and \nameb to address different types of changes. \nameb requires a shorter time window, making it suitable for quick detection or when computational resources are limited, but it assumes at most one change in $\mathcal{G}$. In contrast, \namea is more robust to multiple changes, though it needs a longer time window for accurate estimation.
    
    \item We analyze the asymptotic low bound of the expected detection delay given our intervention mechanism, and prove that \nameb and \namea can achieve the first-order optimality. We further demonstrate that this optimality can only be achieved by adopting our proposed adaptive intervention policy.
    
\end{itemize}

The paper is organized as follows: We review related work in Section \ref{sec: literature}. In Section \ref{sec: background}, we introduce the notations and problem formulation for causal change point detection. Section \ref{sec: intervention} provides a detailed definition of interventions, introduces the centralization method and its properties, and explains how it can be utilized to determine intervention values. Building on this foundation, Section \ref{sec: change point} proposes two monitoring methods with adaptive intervention. The theoretical properties of these methods are discussed in Section \ref{sec: theorem}. Simulation results and case studies are presented in Sections \ref{sec: simulation} and \ref{sec: case}, respectively. Finally, Section \ref{sec: conclusion} concludes the paper with a brief discussion.

\section{Literature Review}
\label{sec: literature}

\textbf{Online change point detection} has been extremely studied in the past decades. Numerous detection statistics have been proposed. Among them, the most popularly used detection statistics are the Cumulative Sum statistic (CUSUM) and the Generalized Likelihood Ratio statistic (GLR). 
If the post-change distribution is known, CUSUM is famous for its optimal property of achieving the smallest detection delay under a fixed false alarm constraint \citep{moustakides1986optimal}. If the post-change distribution is unknown, CUSUM loses its optimality and may not react efficiently, while GLR \citep{lai1998information} provides a common enhancement to CUSUM by estimating the unknown post-change parameters using the maximum likelihood estimation (MLE). However, the disadvantage of GLR is its high computational complexity, which increases with time $t$ due to its search for $\tau$ over all the past time points. Combining the merits of CUSUM and GLR, 
\cite{lorden2008sequential} proposes a CUSUM-type detection scheme by substituting with the estimated parameters via MLE on all the past time points. It can achieve second-order asymptotic optimality. 
Recently, \cite{xie2023window} proposes a window-limited CUSUM, which first constructs a GLR over a window of recent past time points, and then puts it as a CUSUM framework, to further reduce computational burdens. 

\textbf{Change detection with sequential decision} has been increasingly attracting attention when the distribution of the observation is influenced not only by the change point but also by a controller chosen by the observer. 
A detection scheme in this context requires specifying not only the detection statistic but also which controller to choose at each time instance. The most common application is change detection with partially-observable data, where only a subset of variables can be observed at each time, and the controller must actively decide which variable to monitor.
Many sampling policies have been designed for this problem, such as the upper confidence bound \citep{zhang2019partially}, Thompson sampling \citep{zhang2023bandit}, and the epsilon-greedy \citep{gopalan2021bandit,fellouris2022quickest}. In addition to these, \citet{chaudhuri2021sequential,xu2023asymptotic} observe a variable until it is free of changes or the observation time exceeds a threshold, then move to the next.
\citet{tsopelakos2022sequential} further allows the observable variables can vary over time and proposes a probabilistic sampling rule.



\textbf{Change detection for casual graphs} is an emerging research field not yet been well explored. 
So far to our best knowledge, very scarce works are proposed for detecting abrupt change points for a causal graph. \cite{gao2024causal}  focuses on modeling periodic changes via assuming the causal structure evolves according to a Markov process. \cite{gao2024causal2} considers non-periodic change points by evaluating the distributional distance between two causal graph structures over different time windows. Both of them make strong assumptions that the each node follows a discrete distribution with finite-domain. \cite{huang2024causal} defines a causal stability loss function and search for the change-points by binary or top-down segmentations such that each segment has the smallest loss function. However, all these existing works are offline algorithms and cannot be applied to sequential change detection. As for online scenarios, a closely related area is non-stationary causal inference for sequential data, which assumes the causal relationships can vary over time \citep{huang2019causal,huang2020causal,gao2024causal}. 
Vector autoregression model with the first-order differencing stationarity \citep{malinsky2019learning}, state space models \citep{huang2019causal}, nonlinear kernel smoothing models \citep{huang2020causal} have been applied to capture the time-varying dynamics of causal effects. The above methods can be regarded as modeling smooth causal structure changes. However, these works do not explicitly model change detection; their objectives differ fundamentally, focusing instead on tracking or estimating time-varying causal structures

\textbf{Intervention in causal graphs} is also an emerging research field that are commonly used in active Bayesian causal inference \citep{toth2022active,tigas2022interventions,branchini2023causal}. By sequentially selecting the intervention location and value, more efficient causal discovery can be achieved in limited experiment times \citep{agrawal2019abcd,sussex2021near,tigas2022interventions,toth2022active}. Besides, \cite{lattimore2016causal} specifies the output distribution of a multi-armed bandit problem as a causal graph, and treats playing a specific arm as an intervention on the corresponding node. Through sequential active interventions, the total rewards over a time horizon can be maximized \citep{lattimore2016causal,lee2018structural,varici2023causal}. These works generally only consider binary (high and low) intervention values for all the nodes, while do not consider the impact of varying intervention values for different nodes, which is one of the key focuses of our research.

\section{Problem Formulation for Causal Change Point Detection}
\label{sec: background}

In this section, we first introduce the notations, followed by the formulation of our problem based on a linear causal model and the corresponding change point setting.

\subsection{Notations}
For a vector $\bmX \in \bbR^{p}$, denote its $i$th element as $X_i$. For a matrix $\bA \in \bbR^{p\times p}$, denote its $i$th row as $[\bA]_i$ and its $ij$th element as $[\bA]_{i,j}$. For a positive integer $p$, define $[p] \triangleq \{ 1,2,\dots,p\}$. For a subset $\mathcal{S} \subset [p]$, $\bmX_{\mathcal{S}} \in \bbR^{|\mathcal{S}|}$ denotes the vector composed of the corresponding elements of $\bmX$, $[\bA]_{\mathcal{S}} \in \bbR^{|\mathcal{S}|\times p}$ denotes the matrix formed by selecting the corresponding rows of $\bA$, and $[\bA]_{\mathcal{S},\mathcal{S}} \in \bbR^{|\mathcal{S}|\times |\mathcal{S}|}$ denotes the matrix formed by selecting the corresponding rows and columns of $\bA$. $\bmE^j \in \bbR^p$ is a binary 0-1 vector with the 
$j$th element equal to 1 and all other elements equal to 0. $\bE^{i,j} \in \bbR^{p \times p}$ is a binary 0-1 matrix with the 
$ij$th element equal to 1 and all other elements equal to 0. $\bbS^p_{+}$ denotes the set of $p$-dimensional positive definite matrix. When $\lim_{n\to\infty}\frac{f(n)}{g(n)} = 0$, we say $f(n) = o(g(n))$ or $g(n) = \omega(f(n))$. When $\lim_{n\to\infty}\frac{|f(n)|}{|g(n)|} \leq C$ for some $C>0$, we say $f(n) = O(g(n))$ or $g(n) = \Omega(f(n))$. When $\lim_{n \to \infty} \frac{f(n)}{g(n)} = 1$, we say $f(n) \sim_{\infty} g(n)$.
We use $\bmX \sim N(\bmu,\bSigma) $ to denote that $\bmX$ follows the normal distribution with mean $\bmu$ and variance $\bSigma$, and use $\phi_p$ to represent the probability density function of a $p$-dimensional standard normal distribution. We use $\bmX \sim 1_{\mathbf{c}}$ to denote that $\bmX$ follows a point mass distribution, which takes the value $\mathbf{c}$ with probability 1.


\subsection{Linear Causal Model}
We consider a causal graph with known structure $(\mathcal{V}, \mathcal{B}^t)$. Here $\mathcal{V} \triangleq [p]$ is the set of nodes. $\mathcal{B}^t$ is the set of edges at time $t$ where the superscript $t$ indicates that it may vary over time. For each time $t$, the vector of random variables associated
with the nodes is denoted by $\bmX^{t} = (X_1^t, \dots, X_p^t)^T$. We refer to the set of parents, ancestors, and descendants of node $i\in \mathcal{V}$  at time $t$ by $pa^t(i), anc^t(i)$ and $des^t(i)$, respectively. For a linear causal model, we consider a linear structural equation model (SEM) as follows:
\begin{equation}
\label{eq:linSEM}
    \bmX^t = \bA^t \bmX^t + \bmU^t,
\end{equation}
where $\bA^t \in \bbR^{p \times p}$ is the edge weight matrix. It is strictly lower triangular and note that $[\bA^t]_{i,j} \neq 0$ if and only if $j \in pa^t(i)$. $\bmU^t {\sim} N(\bmu^t, \bSigma^t)$ with $\bSigma^t \triangleq diag({\sigma_1^t}^2, \dots, {\sigma_p^t}^2)$ is the exogenous background variables and $\bmU^t$ at different times are independent of each other.  In addition, we assume that there are no confounding nodes in our causal model.


\subsection{Change Point Setting}
\label{sec:change set}
According to the SEM of \eqref{eq:linSEM}, we can write the distribution of $\bmX^t \sim N(\bmu_x^t, \bSigma_x^t)$ where
\begin{equation*}
   \bmu_x^t=(\bI_p - \bA^t)^{-1}\bmu^t, \quad \bSigma_x^t=(\bI_p-\bA^t)^{-1}\bSigma^t {(\bI_p- \bA^t)^{-1}}^T,
\end{equation*}
where $\bI_p$ is the  $p$-order identity matrix. Consequently, the distribution of $\bmX^t$ can be decided by a triplet $(\bA^t, \bmu^t, \bSigma^t)$. Now we introduce our change point model.
\begin{equation}
    (\bA^t,\bmu^t,\bSigma^t) =
    \begin{cases}
    &(\bA,\bmu,\bSigma),\quad  \text{for} \quad t = 1,2,\dots,\tau-1, \\
    &(\Tilde{\bA},\Tilde{\bmu}, \Tilde{\bSigma}),\quad \text{for} \quad t = \tau,\tau+1,\dots.
    \end{cases}
    \end{equation}
Here $(\bA,\bmu,\bSigma)$ are parameters in the pre-change condition, which are assumed to be either known in advance or estimated via sufficient historical data. $\tau$ is the unknown change point, $(\Tilde{\bA},\Tilde{\bmu}, \Tilde{\bSigma})$ are unknown post-change edges. 
When there is no change, we can treat $\tau = \infty$.
Hereafter, for convenience, we use $ pa(i),anc(i),des(i)$ to denote $ pa^t(i),anc^t(i),des^t(i)$ for $t<\tau$. 

\begin{assumption}[Post-Change Condition]
\label{ass:single change}
    The triplets $(\bA,\bmu,\bSigma)$ and $(\Tilde{\bA},\Tilde{\bmu},\Tilde{\bSigma})$ differ by only one component's only one element, while all its other elements and the other two components remain the same. $\Delta$ is the difference in value between this distinct element and $|\Delta| \in [\Delta_{min},\Delta_{max}]$, where $\Delta_{min},\Delta_{max}$ are known parameters. Furthermore, $\Tilde{\bA}$ should still guarantee a directed acyclic graph. 
\end{assumption}

\section{Intervention Policy Design}
\label{sec: intervention}
In this section, we present the design principles of the intervention policy. Section \ref{sec:intervention} introduces the intervention mechanism and discusses its impact on linear SEMs. In Section \ref{sec: intervention sub2}, we propose a novel centralization method that guides the design of the optimal intervention node. Finally, in Section \ref{sec:intervention sub3}, we leverage the KL divergence to identify the optimal intervention node by setting an appropriate intervention value.


\subsection{Intervention Mechanism}
\label{sec:intervention}
According to \eqref{eq:linSEM}, the condition distribution of node $i$, $X^t_i$, given its parents, $\bmX^t_{pa^t(i)}$, is $X^t_i|\bmX^t_{pa^t(i)} \sim N(\sum_{j\in pa^t(i)} [\bA^t]_{i,j}X^t_j+ \mu^t_i, {\sigma_i^t}^2)$. 
If we intervention node $i$ at $t$ with value $c_i$, denoted as $do(X_i^t = c_i)$, this changes its conditional distribution as $X^{t,do(X^t_i = c_i)}_i|\bmX^t_{pa^t(i)} \sim 1_{c_i}$. 
From the perspective of causal graph structure, $do(X_i^t = c_i)$ indicates removing the edges in $\mathcal{B}^t$ that point to node $i$, as well as the edge from the corresponding exogenous
background variable $U^t_i$ to node $i$, and then setting its value to $c_i$. In this way, the joint distribution of $\bmX^{t,(do(X_i^t = c_i))}$ can be written as
\begin{equation}
\label{eq:intervention X}
    \begin{aligned}
    \bmX^{t,do(X_i^t=c_i)} \sim  
    N(&(\bI_p - \bA^{t,do(X_i^t=c_i)})^{-1}\bmu^{t,do(X_i^t=c_i)}, \\ &(\bI_p-\bA^{t,do(X_i^t=c_i)})^{-1}\bSigma^{t,do(X_i^t=c_i)} {(\bI_p- \bA^{t,do(X_i^t=c_i)})^{-1}}^T),
\end{aligned}
\end{equation}
where
\begin{equation}
\label{eq:intervention X 2}
\begin{aligned}
    \mu^{t,do(X_i^t = c_i)}_j &= \begin{cases}
        \mu^t_j &\quad\text{for } j\neq i, \\
        c_i &\quad\text{for } j= i, 
    \end{cases}\\
    [\bSigma^{t,do(X_i^t = c_i)}]_{j,k}& = \begin{cases}
        [\bSigma^t]_{j,j} &\quad \text{for } j=k \neq i, \\
        0 &\quad \text{for } j\neq k \  \text{or } j=k=i,\\
    \end{cases} \\
    [\bA^{t,do(X_i^t = c_i)}]_j &= 
    \begin{cases}
        [\bA^t]_j &\quad \text{for } j\neq i,\\
        \mathbf{0} &\quad \text{for } j= i.
    \end{cases}
\end{aligned}
\end{equation}

We can verify that, under this expression, the distribution of $X_i^{t,do(X_i^t = c_i)}$ degenerates to $N(c_i,0)$, which is a point mass distribution, i.e., $1_{c_i}$. This aligns with the definition of an intervention. For convenience, we use $do(X_0^t)$ to denote that we do not do any intervention, 
use $(\bA^{do(X_i=c_i)}, \bmu^{do(X_i = c_i)},\bSigma^{do(X_i=c_i)})$ for $t < \tau$ and use $(\Tilde{\bA}^{do(X_i=c_i)}, \Tilde{\bmu}^{do(X_i = c_i)},\Tilde{\bSigma}^{do(X_i=c_i)}))$ for $t \geq \tau$. We denote $\bmu_x^{do(X_i^t = c_i)} \triangleq (\bI_p - \bA^{do(X_i=c_i)})^{-1}\bmu^{do(X_i=c_i)}$.

What advantage does intervention bring to the change point detection problem? For example, when a change occurs in $[\bA]_{k,j}$, suppose the magnitude of this change is small, making it difficult to detect. However, 
if intervening on node $j$ or its ancestors, and setting their values significantly above their mean, we can amplify the initially subtle change of node $k$ and its descendants, making it easier to detect the change. In summary, we can detect the change earlier by choosing an appropriate node and intervention value. The key question is how to select the best node to intervene on and the corresponding intervention value, maximizing our detection efficiency. We will discuss this in detail in the following sections.

\subsection{Centralization and Optimal Intervention Node Design}
\label{sec: intervention sub2}
We first introduce a novel centralization method for $\bmX^{t, do(X_i^t = c_i)}$, which can concentrate change and lay the foundation of intervention policy.
\begin{definition}[Centralization]
\label{def: centrilazation}
    We say $\bmY^{t,do(X_i^t = c_i)}$ is the centralization of $\bmX^{t, do(X_i^t = c_i)}$, if the following satisfies,
    \begin{equation}
\label{eq:central}
    \begin{aligned}
    Y^{t,do(X_i^t=c_i)}_{i} &= 0 , \text{if  } i\neq 0,\\
    \bmY^{t,do(X_i^t=c_i)}_{[p]\backslash i} &= {[\bSigma^{do(X_i = c_i)}]^{-1/2}_{[p]\backslash i, [p]\backslash i}} (\bI_{|[p]\backslash i|} - [\bA^{do(X_i=c_i)}]_{[p]\backslash i, [p]\backslash i})(\bmX^{t,do(X_i^t = c_i)} _{[p]\backslash i} - \bmu^{do(X_i = c_i)}_{x,[p]\backslash i}),
\end{aligned}
\end{equation}
where $[\bSigma^{do(X_i = c_i)}]_{[p]\backslash i, [p]\backslash i}$ is a diagonal matrix, 
taking the $-\frac{1}{2}$ power only requires applying the $-\frac{1}{2}$ power to each diagonal element individually.
\end{definition}
In Definition \ref{def: centrilazation}, the case $do(X_0^t = c_0)$ (abbreviated as $do(X_0^t)$) can be considered a special case of the above equation when $[p]\backslash 0$ is treated as $[p]$.

\begin{proposition1}
\label{prop:Yt}
    For any  $i \in [p]$, $\bmY^{t,do(X_i^t=c_i)}_{[p]\backslash i}$ follows a normal distribution. Under Assumption \ref{ass:single change}, we have the following properties. \\ 
1) For $t < \tau$, $ \bbE(\bmY^{t,do(X_i^t=c_i)}_{[p]\backslash i}) =\mathbf{0}, Cov(\bmY^{t,do(X_i^t=c_i)}_{[p]\backslash i})= \bI_{|[p]\backslash i|}$. \\
2) For $t \geq \tau$, further separate discussions are needed.
        \begin{itemize}
            \item When change occurs in $[\bA]_{k,j}$, i.e. $\Tilde{\bmu} = \bmu, \Tilde{\bSigma} = \bSigma,\Tilde{\bA} = \bA + \Delta\bE^{k,j} $, \\
            \begin{align} 
             \label{eq:E_Y}
            \bbE(\bmY^{t,do(X_i^t=c_i)}_{[p]\backslash i}) &= \left(\sigma_k^{-1}\Delta\left(\mu^{do(X_i=c_i)}_j+ \sum_{l\in anc(j)} [\bB^{do(X_i = c_i)}]_{j,l}\mu^{do(X_i=c_i)}_l  \right)\bmE^k\right)_{[p]\backslash i},\\ \nonumber
            Cov(\bmY^{t,do(X_i^t=c_i)}_{[p]\backslash i}) &= \left(\bI_p +
            \sigma_k^{-1}\Delta \left(\sum_{l\in anc(j)\cup j}[\bB^{do(X_i=c_i)}]_{j,l}[\bSigma^{do(X_i = c_i)}]^{1/2}_{l,l}\bE^{k,l}  \right) + \right.\\ \nonumber
            &\quad \quad \quad \sigma_k^{-1}\Delta \left(\sum_{l\in anc(j)\cup j}[\bB^{do(X_i=c_i)}]_{j,l}[\bSigma^{do(X_i = c_i)}]^{1/2}_{l,l}\bE^{l,k}\right)+
            \\  \label{eq:Cov_Y}
             &\left.\quad \quad \quad \sigma_k^{-2}\Delta^2\left(\sum_{l\in anc(j)\cup j}([\bB^{do(X_i=c_i)}]_{j,l})^2[\bSigma^{do(X_i = c_i)}]_{l,l}\right)\bE^{k,k}\right) _{[p]\backslash i, [p]\backslash i},
            \end{align}          
    where $\bB^{do(X_i=c_i)} \triangleq (\bI_p-\bA^{do(X_i=c_i)})^{-1}$. 
             \item When change occurs in $\mu_k, k\in[p]$, i.e., $\Tilde{\bA} = \bA, \Tilde{\bSigma} = \bSigma, \Tilde{\bmu} = \bmu + \Delta \bmE^k$,  \\
           $\bbE(\bmY^{t,do(X_i^t=c_i)}_{[p]\backslash i}) = ( \sigma_k^{-1}\Delta \bmE^k)_{[p]\backslash i},  Cov(\bmY^{t,do(X_i^t=c_i)}_{[p]\backslash i})= \bI_{|[p]\backslash i|}$.
        
            \item When change occurs in $[\bSigma]_{k,k}$, i.e. $\Tilde{\bA} = \bA, \Tilde{\bmu} = \bmu, \Tilde{\bSigma} = \bSigma + \Delta \bE^{k,k}$, \\
            $\bbE(\bmY^{t,do(X_i^t=c_i)}_{[p]\backslash i}) = \mathbf{0},  Cov(\bmY^{t,do(X_i^t=c_i)}_{[p]\backslash i})= (\bI_{p} + \sigma_k^{-2}\Delta\bE^{k,k})_{[p]\backslash i, [p] \backslash i}$.
    \end{itemize}
\end{proposition1}
\begin{proof}
    The proof is shown in Appendix \ref{app: prop 1}.
\end{proof}

Proposition \ref{prop:Yt} presents the distribution of $\bmY^{t,do(X_i^t = c_i)}_{[p]\backslash\ i}$, obtained by centering $\bmX^{t,do(X_i^t = c_i)}_{[p]\backslash\ i}$ according to (\ref{eq:central}) under different change situations. From it, we can get the following important properties.
\begin{itemize}
    \item Under pre-change condition, $\bmY^{t,do(X_i^t = c_i)}_{[p]\backslash\ i}$ follows a standard normal distribution. 
    \item When a change occurs in $\mu_k$ or $[\bSigma]_{k,k}$, there are two possible intervention scenarios. In the first case, if we intervene on node $k$, the distribution of $\bmY^{t,do(X_k^t = c_k)}_{[p] \backslash k}$ remains the same as in the pre-change condition. This is because the intervention on node $k$ masks the change, preserving the distribution of the other variables. In the second case, if we do not intervene on node $k$, then intervening on any other node—with any intervention value—or not intervening at all results in the same post-change distribution. This implies that, for this type of change, intervention does not improve detection efficiency. Therefore, \textbf{in the following sections, we focus solely on cases where the change occurs in $\bA$}. 
    \item When a change occurs in $[\bA]_{k,j}$, intervening on node $k$ results in the same outcome as the first scenario where the change is in $\mu_k$ or $[\bSigma]_{k,k}$: the distribution of $\bmY^{t,do(X_k^t = c_k)}_{[p] \backslash k}$ remains identical to the pre-change condition. In subsequent algorithms, it is essential to avoid this scenario.

    \item When change occurs in $[\bA]_{k,j}$,  
    for $\forall i \notin anc(j)\cup j\cup k$, $ \forall l \in anc(j)\cup j$
    \begin{align*}
        [\bB^{do(X_i=c_i)}]_{j,l} = [\bB^{do(X_0)}]_{j,l},  \quad
        [\bSigma^{do(X_i=c_i)}]_{l,l} = [\bSigma^{do(X_0)}]_{l,l}.
    \end{align*}
    This indicates that when the intervened node is neither $j$ itself nor an ancestor of $j$, the resulting post-change distribution is the same as without intervention, providing no improvement in change detection efficiency. Such situations should be avoided. This is because we intervene on downstream nodes of the change, but the intervention values we set did not amplify the change magnitude. As a result, the outcome is effectively the same as if no intervention had occurred. 
    \item When change occurs in $[\bA]_{k,j}$, if we intervene on node $j$, for $\forall l \in anc(j)\cup j$,
    \begin{equation}
    \label{eq:B anc}
         [\bB^{do(X_j = c_j)}]_{j,l} = 0, \quad
        [\bSigma^{do(X_j = c_j)}]_{j,j} = 0,\quad
        \mu_j^{do(X_j=c_j)} = c_j.
    \end{equation}
    In this way, we can further get, 
    \begin{equation*}
        \bmY^{t,do(X_j^t = c_j)}_{[p]\backslash j} \sim N(\left(\frac{\Delta c_j}{\sigma_k}\bmE^k\right)_{[p]\backslash j}, \bI_{p-1}).
    \end{equation*}
   This indicates that the different dimensions of $\bmY^{t,do(X_j^t=c_j)}_{[p]\backslash j}$ are mutually independent, and hence we do not need to account for their covariance. The expectation $\left(\frac{\Delta c_j}{\sigma_k}\bmE^k\right)_{[p]\backslash j}$ further shows that only the exact downstream node $k$ has the mean shift and hence the change is concentrated. Compared to no intervention case,  
   by further controlling the intervention value $c_j$, we can increase the mean shift of its post-change distribution, making it easier to detect. 
   \item  When change occurs in $[\bA]_{k,j}$, if we intervene on node $i \in anc(j)$ ,  \eqref{eq:E_Y} and \eqref{eq:Cov_Y} cannot be further simplified. It indicates the covariance matrix of $\bmY^{t, do(X_j^t = c_j)}_{[p]\backslash j}$ is not an identity matrix, which is not as preferable as intervening on node $j$. 
\end{itemize}

In conclusion, when a change occurs in $[\bA]_{k,j}$, 
the optimal choice is to intervene on node $j$ itself, i.e., the origin of the changed edge, which leads to independent distributions of $\bmY^{t, do(X_j^t = c_j)}_{[p]\backslash j}$ and allows us to achieve the desired shift magnitude by adjusting $c_j$. 
The challenge is in practice, we often do not know which specific edge has undergone a change, so identifying the origin of the changed edge and selecting an appropriate intervention value $c_i$ can be challenging. To solve this challenge, we propose a rule for setting the optimal intervention value for each node in the following, which ensures that, regardless of which edge has changed, the origin node of the changed edge, will be the optimal intervention node. 

\subsection{Optimal Intervention Value Design}
\label{sec:intervention sub3}
We first define the KL divergence to measure how much the post-change distribution shifts from the pre-change distribution when intervening on node $a_t$ at time $t$, i.e.,
\begin{align}
\label{eq:KL}
    I_{a_t,c_{a_t}}^{\Delta,[k,j]} & \triangleq D(f_{a_t,c_{a_t}}^{\Delta,[k,j]}||f_{a_t,c_{a_t}}^{0}) = \int \log \left(\frac{f_{a_t,c_{a_t}}^{\Delta,[k,j]}(x)}{f_{a_t,c_{a_t}}^{0}(x)}  \right) f_{a_t,c_{a_t}}^{\Delta,[k,j]}(x) dx \\ \label{eq:Ia detal}
    &= \frac{1}{2}\sigma_k^{-2}\Delta^2 \left(\mu^{do(X_{a_t}=c_{a_t})}_j+ \sum_{l\in anc(j)} [\bB^{do(X_{a_t} = c_{a_t})}]_{j,l}\mu^{do(X_{a_t}=c_{a_t})}_l  \right)^2\\ \nonumber
    & +  \frac{1}{2}\sigma_k^{-2}\Delta^2\left(\sum_{l\in anc(j)\cup j}[\bB^{do(X_{a_t}=c_{a_t})}]_{j,l}[\bSigma^{do(X_{a_t} = c_{a_t})}]_{l,l}\right) + \frac{1}{2}.
    \end{align}
Here $f_{a_t,c_{a_t}}^{\Delta,[k,j]}$ denotes the distribution of $\bmY^{t,do(X_{a_t}^t = c_{a_t})}_{[p]\backslash a_t}$ when the change occurs on $[\bA]_{k,j}$ and the change magnitude is $\Delta$.$f_{a_t,c_{a_t}}^{0}$ denotes the pre-change distribution under the same intervention setting since the change magnitude is $0$. $f_{0,c_0}^{\Delta,[k,j]}$ and $I^{\Delta,[k,j]}_{0,c_0}$ represent their distributions with no intervention. For convenience, we omit the subscript $c_0$, and use $f_{0}^{\Delta,[k,j]}$ and $I^{\Delta,[k,j]}_{0}$ hereafter. Based on \eqref{eq:KL}, we define the best choice of $a_t$, which have the largest KL divergence $I_{a_t,c_{a_t}}^{\Delta,[k,j]}$ as
\begin{align}
\label{eq:KLselect}
a_{opt} = \arg\min_{a_t \in 0\cup[p]}I_{a_t,c_{a_t}}^{\Delta,[k,j]}.
\end{align}
As previously discussed, when a change occurs at $[\bA]_{k,j}$, node $j$ is the optimal choice because it allows the covariance matrix of $\bmY^{t,do(X_{j}^t = c_j)}_{[p]\backslash j} $ to become an identity matrix, making it easier to analysis. Thus, $j = a_{opt}$  is the result we aim to achieve. In practice, however, we have no knowledge of the change magnitude, such as the values of $\Delta$ and $[k,j]$. What we can control is only the intervention value, $\bmC \triangleq(c_1,\dots,c_p)^T$. Next, we will develop an policy to set the optimal intervention values, $\bmC$, which can guarantee that the selected $a_{opt}$ according to \eqref{eq:KLselect} always equal to the origin of the changed edge.  

Without loss of generality, we assume that all the nodes are ordered according to a topological order, i.e., $i < j \Leftrightarrow i \notin des(j)$.  Then, we sequentially consider the situation that each node is the origin of the change. Starting with the root node, i.e., node 1, if it is the origin of the change, we aim to achieve that 
\begin{equation}
\label{eq: node 1}
    I_{1,c_{1}}^{\Delta,[k,1]} > I_{j,c_{j}}^{\Delta,[k,1]} + \delta, \quad \forall j \in 0\cup[p]\backslash 1, \ k\in [p]\backslash (1\cup anc(1)),
\end{equation} 
where $\delta$ is a user-specified parameter representing intervention gap.
According to (\ref{eq:Ia detal}), we can find that $anc(1) = \emptyset$. This means that $I_{j,c_{j}}^{\Delta,[k,1]}$ for $\forall j \in 0\cup[p]\backslash 1, k\in [p]\backslash (1\cup anc(1)) $ is unrelated to $\bmC$. In this way, by combining (\ref{eq:Ia detal}) and (\ref{eq: node 1}), we can derive the range of $c_1$ should have the form $(-\infty, m_1] \cup [M_1, \infty)$, where $m_1 < 0$ and $M_1 > 0$ are two constants. In this paper, we select $M_1$ as the final value for $c_1$. Its exact value is shown in Algorithm \ref{alg: c_j}. 

Next, assume that the second node, i.e., node 2, is the origin of the changed edge, and we aim to achieve $I_{2,c_{2}}^{\Delta,[k,2]} > I_{j,c_{j}}^{\Delta,[k,2]} + \delta, \quad \forall j \in 0\cup[p]\backslash 2,  k\in [p]\backslash (2\cup anc(2)) $. Following a similar step as that for node 1, according to (\ref{eq:Ia detal}), we can find that the right side of the inequality depends at most on $ c_1$ and is independent of the other elements of $\bmC$. Since $c_1$ is already determined at this point, by combining (\ref{eq:Ia detal}) and (\ref{eq: node 1}), we can derive the range of $c_2$ should have the form $(-\infty, m_2] \cup [M_2, \infty)$, where $m_2 < 0$ and $M_2 > 0$ are two constants. Similarly, we select $M_2$ as the final value for $c_2$. 

By iterating this process until the final node, we can determine all $c_j$ values. The specific calculation details are provided in Algorithm \ref{alg: c_j}. 

\begin{algorithm}
        \caption{Set optimal intervention value $\bmC$} 
	\label{alg: c_j}
    \begin{algorithmic}
        \REQUIRE $\bmu,\bSigma,\bA,\Delta_{min},\delta$
        \ENSURE $\bmC = (c_1,\dots,c_p)^T$
        \STATE \textcolor{gray}{$\sharp$ Assume all nodes are orders by topological order.}
        \FOR{$j = 1,\dots,p$ } 
        \STATE $\bSigma^{do(X_j = c_j)}_{max} = \max_{k \notin anc(j)\cup j} [\bSigma^{do(X_j = c_j)}]_{k,k}$\\ 
        \FOR{$i \in anc(j) \cup 0$}
        \STATE \begin{equation*}
            \begin{aligned}
            \Tilde{c}_{j,i} = & ( (\sum_{m \in anc(j)}[\bB^{do(X_i = c_i)}]_{j,m}\mu_m^{do(X_i = c_i)})^2 +   \sum_{m \in anc(j)}[\bSigma^{do(X_i=c_i)}]_{m,m}([\bB^{do(X_i = c_i)}]_{j,m} )^2 \\ 
            & +   \frac{2\delta \bSigma^{do(X_j = c_j)}_{max}}{\Delta_{min}^2} )^{1/2}
        \end{aligned}
        \end{equation*}
        
        \ENDFOR
        \STATE $c_j = \max_{i\in anc(j)\cup0} \Tilde{c}_{j,i}$
        \ENDFOR
    \end{algorithmic}
\end{algorithm}

To evaluate how the change is concentrated in a particular node $l$, we denote $f_{a_t,c_{a_t}}^{\Delta,[k,j]}[l]$ as the marginal distribution of $Y^{t,do(X_{a_t}^t = c_{a_t})}_{l}$ when the change occurs on $[\bA]_{k,j}$ and the change magnitude is $\Delta$, and $f_{a_t,c_{a_t}}^{0}[l]$ as its pre-change distribution. Define the specific KL divergence of node $l$ as
\begin{equation}
\label{eq:sKL}
     I_{a_t,c_{a_t}}^{\Delta,[k,j]}[l] \triangleq D(f_{a_t,c_{a_t}}^{\Delta,[k,j]}[l]||f_{a_t,c_{a_t}}^{0}[l]). 
\end{equation}
It represents the difference in the distribution of the $l$th dimension of $\bmY^{t,do(X^t_{a_t} = c_{a_t})}$ between the pre-change and post-change conditions, when a change occurs on $[\bA]_{k,j}$ and the change magnitude is $\Delta$. Using this, we can observe our centralization technique in Proposition \ref{prop: Ia}, which consolidates the change information into a specific dimension.

\begin{proposition1}
    \label{prop: Ia}
    If the intervention values $\bmC$ are set according to Algorithm \ref{alg: c_j}, under Assumption \ref{ass:single change}, for $\forall \Delta \in [\Delta_{min},\Delta_{max}]$ and $\forall k ,j \in [p], k\neq j$ we have the follows,
    \begin{align}
    \label{eq:prop2 1}
        I_{j,c_j}^{\Delta,[k,j]} &\geq I_{i,c_i}^{\Delta,[k,j]} + \delta, \quad \forall i \in 0\cup[p]\backslash j, \\
        \label{eq:prop2 2}
        I_{j,c_j}^{\Delta,[k,j]}[k] &\geq I_{i,c_i}^{\Delta,[k,j]}[k] + \delta, \quad\forall i \in 0\cup[p]\backslash j,\\
        \label{eq:prop2 3}
        I_{i,c_i}^{\Delta,[k,j]}[l] &= 0, \quad \forall i\in0\cup[p], l\neq k, \\
        \label{eq:prop2 4}
        I_{j,c_j}^{\Delta,[k,j]} &= I_{j,c_j}^{\Delta,[k,j]}[k].
    \end{align}
\end{proposition1}
\begin{proof}
    The proof is shown in Appendix \ref{app:prop2}.
\end{proof}

Proposition \ref{prop: Ia} shows that, regardless of the location of the change, by setting intervention values for nodes according to Algorithm \ref{alg: c_j}, for the change at $[\bA]_{k,j}$, the KL divergence by intervening on node $j$ is always larger than the KL divergences by intervening on other nodes. 
Furthermore, (\ref{eq:prop2 2})-(\ref{eq:prop2 4}) demonstrate that our centralization can effectively concentrate all change information onto the node targeted by the changed edge, i.e., node $k$. This further provides rationale and theoretical support for constructing detection statistics for each node separately in the subsequent analysis. 
\begin{corollary1}
    \label{col: Ib}
 Given the intervention values $\bmC$ set according to Algorithm \ref{alg: c_j}, for any change magnitude $\Delta$ and change at $[\bA]_{k,j}$ that satisfy Assumption \ref{ass:single change}, if we select  
    \begin{equation}
        a_t \triangleq \arg\max_{i\in0\cup[p]} I_{i,c_i}^{\Delta, [k,j]} = \arg\max_{i\in0\cup[p]} I_{i,c_i}^{\Delta,[k,j]}[k],
    \end{equation}
    then $a_t$ is the optimal intervention node, i.e., the origin of the changed edge, node $j$.
\end{corollary1}
Corollary \ref{col: Ib} shows using the KL divergence, we can automatically identify the optimal node to intervene on.

For notation simplicity, hereafter we can further simplify $do(X_i^t = c_i)$ with $c_i$ set by Algorithm \ref{alg: c_j} as $do(i)$, and correspondingly simplify $f^{\Delta,[k,j]}_{a_t,c_{a_t}}$ as $f^{\Delta,[k,j]}_{a_t}$, $f^{0}_{a_t,c_{a_t}}$ as $f^{0}_{a_t}$, $f^{\Delta,[k,j]}_{a_t,c_{a_t}}[l]$ as $f^{\Delta, [k,j]}_{a_t}[l]$, and $f^{0}_{a_t,c_{a_t}}[l]$ as $f^{0}_{a_t}[l]$. 

\section{Change Point Detection with Adaptive Intervention}
\label{sec: change point}
Through our centralization and intervention value-setting policy, $\bmY^{t,do(X_j^t = c_j)}_{[p]\backslash j}$ can efficiently concentrate the change in a single node. So in this section, we use $\bmY^{t,do(X_j^t = c_j)}_{[p]\backslash j}$ to construct our detection statistic and the online adaptive intervention policies. We consider two types of detection algorithms: \nameb and \namea. As mentioned earlier, \nameb requires a shorter time window but assumes that only a single change occurs in the system. In contrast, \namea is more robust to multiple changes, with minimal impact on its performance, but it requires a longer time window for reliable estimation.

In section \ref{sec:MULTI AI},  we present our detection method \namea which combines MULTI-CUSUM detection statistic and corresponding Adaptive Intervention policy.  In section \ref{sec:MAX AI},  we present our detection method \nameb which combines MAX-CUSUM detection statistic and corresponding Adaptive Intervention policy.

\subsection{\nameaa}
\label{sec:MULTI AI}
In reality, since we do not know the true parameters of post-change $\{\Delta, [k,j] \}$, we can set a time window length $w$ and use the past $w$ observations up to $t$, i.e., $\bmY^{m,do(a_m^{\circ})}_{[p]\backslash a_m^{\circ}}, m = t-w, \ldots, t-1$,  where $A^{\circ} =  \{a_t^{\circ}| t\in \bbN\}$ denotes the intervention actions decided by \nameaa, to estimate $f^{\Delta, [k,j]}_{i}, \forall i \in 0\cup[p]$.  According to Proposition \ref{prop:Yt}, this is equivalent to estimate mean $\bmu_Y^{t,do(i)}\triangleq\bbE[\bmY^{t,do(i)}_{[p]\backslash i}]$ and covariance matrix $ \bSigma_Y^{t,do(i)}\triangleq Cov(\bmY^{t,do(i)}_{[p]\backslash i})$ for  $ i \in 0\cup[p]$, which can be achieved by
\begin{equation}
\begin{aligned}
    \label{eq:esttheta multi}
    (\hat{\bmu}_Y^{t,do(i),\circ}, \hat{\bSigma}_Y^{t,do(i),\circ}) = \arg\max_{\bm{\nu} \in \bbR^{|[p]\backslash i]|},\mathbf{S} \in \bbS^{|[p]\backslash i]|}_{+}} \sum_{m = t-w}^{t-1} \log \phi_{|[p]\backslash i|}(\mathbf{S}^{-1/2}(\bmY^{m,do(a_m^{\circ})}_{[p]\backslash a_m^{\circ}} - \bm{\nu})) \bbI\{a_m^{\circ} = i \}, \\
   \forall \quad i \in 0\cup [p].
\end{aligned}
\end{equation}
Here $\hat{\bmu}_Y^{t,do(i),\circ}$ and $ \hat{\bSigma}_Y^{t,do(i),\circ}$ are the estimation of $\bmu_Y^{t,do(i)}$ and $\bSigma_Y^{t,do(i)}$  respectively. \nameaa. $\bbI\{\cdot\}$ is the indicative function. With $\hat{\bmu}_Y^{t,do(i),\circ}$ and $\hat{\bSigma}_Y^{t,do(i),\circ}$ for $i \in 0\cup [p]$, we can get the estimated $\hat{f}_{i}^{t,\Delta,[k,j],\circ}$ for $i \in 0\cup [p]$. Then we construct the likelihood ratio test: 
 
\begin{equation}
\begin{aligned}
\label{eq:W multi}
\Lambda_{t,A^\circ}^{\Delta,[k,j],\circ} &= \log\left(\frac{\hat{f}^{t,\Delta,[k,j],\circ}_{a_t^\circ}(\bmY^{t,do(a_t^{\circ})}_{[p]\backslash a_t^\circ})}{f^{0}_{a_t^\circ}(\bmY^{t,do(a_t^\circ)}_{[p]\backslash a_t^\circ})} \right).
\end{aligned}
\end{equation}
The larger the value of $\Lambda_{t,A^\circ}^{\Delta,[k,j],\circ}$, the more likely the system is to be under the post-change condition. To further accumulate small change information, we conduct the multi-CUSUM-type detection statistic as
\begin{equation}
    \label{eq:W multi 2}
     W_{t,A^\circ} = \max\{W_{t-1,A^\circ} ,0\} + \Lambda_{t,A^\circ}^{\Delta,[k,j],\circ},\ t>w,  \quad W_{w,A^\circ} = 0.
\end{equation}
The system triggers a change alarm at $T_{b,A^{\circ}}^{multi} = \inf\{t>w| W_{t,A^{\circ}} >b\}$ where $b$ is a constant threshold.

Before introducing the corresponding adaptive intervention policy $A^{\circ} = \{a^{\circ}_t|t \in \bbN\}$, we define a sequence of deterministic time points, $\sN=\{t_{1},t_{2},\ldots\} \subset \{w+1,w+2,\dots\}$,
which contains at most $q$ and at least $q^\eta$ elements where $0< \eta\leq 1$ is a constant and $q$ is a positive integer smaller than $w$. The sequence $\sN$ is constructed such that the number of selected time points in any interval of length $w$ satisfies
\begin{equation}
    \frac{n}{w}q^{\eta} \leq \sum_{t = w+1}^{w+n} \bbI\{t\in \sN\} \leq \frac{n}{w}q,\quad \forall n\in \bbN.
\end{equation}
In conjunction with the adaptive intervention policy that will be introduced shortly, a larger value of $q$ encourages the algorithm to perform more diverse interventions, improving estimation accuracy (exploration). Conversely, a smaller $q$ causes the algorithm to rely more heavily on current estimates when selecting intervention nodes (exploitation). By tuning $q$, the algorithm balances the trade-off between exploration and exploitation through the sequence $\sN$.

Then we define the corresponding adaptive intervention policy $A^{\circ} = \{a^{\circ}_t|t \in \bbN\}$ as follows, 
\begin{enumerate}[i.]
    \item For $t = 1,\dots,w$, $a_{t}^{\circ} = V^t$ where $V^t$ is a random variable uniformly distributed in $0\cup[p]$. Additionally, $V^t$ is independent of $\{\bmU^s\}_{s=1}^{t-1}$ and is also independent across different time points $t$.
    \item For $t > w$ and $t \in \sN$, $a_t^{\circ} = V^t$ where $V^t$ is similarly defined as the last step. 
    \item For $t > w$ and $t \notin \sN$, we can use $\{\hat{\bmu}_Y^{t,do(i),\circ}, \hat{\bSigma}_Y^{t,do(i),\circ}\}_{i\in0\cup[p]}$ to estimate the KL divergence and set 
    \begin{equation}
    \label{eq:inter multi}
        a_t^{\circ} = \arg\max_{a\in 0\cup[p]} \hat{I}_{a}^{t,\Delta,[k,j],\circ} =  \arg\max_{a\in 0\cup[p]}D(\hat{f}_{a}^{t,\Delta,[k,j],\circ}||f_{a}^{0}).
    \end{equation}
\end{enumerate}

We suggest $w$ of \namea to be at least $O(p^2)$ to ensure estimaton accuracy of $\{(\hat{\bmu}_Y^{t,do(i),\circ}, \hat{\bSigma}_Y^{t,do(i),\circ} )\}_{i\in0\cup[p]}$. During exploitation, i.e., $t\notin \sN$, we estimate the joint distribution of $\bmY^{t,do(i)}$ obtained after intervening on node $i$, and select node $i$ whose distribution has the greatest KL divergence from the pre-change distribution, as the intervention node for the next time. This guarantees the selection of the node most likely to be the origin of the change, which is considered the optimal choice, as noted in Section~\ref{sec: intervention sub2}. During exploration, i.e., $t \in \sN$, we uniformly randomly select nodes for intervention, which guarantees accurate estimation of the joint distribution of $\bmY^{t,do(i)}$ for each $i\in0\cup [p]$.

However, \namea has some disadvantages. First, it requires a relatively long time window of order $O(p^2)$, which demands substantial computational resources. Second, it does not take advantage of a key insight in Proposition \ref{prop: Ia}: when we select the best intervention node $i$, the different dimensions (i.e., nodes) of $\bmY^{t,do(i)}$ are independent and the change is concentrated in only one node.

\subsection{\namebb}
\label{sec:MAX AI}

To fully utilize Proposition \ref{prop: Ia}, we further propose another detection method \namebb. The core idea of \nameb is to construct $p$ separate detection statistics, each targeting a specific node $l$ of $\bmY_{l}^{t,do(i)}$. As soon as any one of these $p$ statistics triggers an alarm, we treat the system has changed. 

In this way, we do not need to estimate the covariance of $\bmY^{t,do(i)}_{[p]\backslash i}$, i.e., $\bSigma_Y^{t,do(i)}$. Instead, we only need to estimate the means and variance of each node $l$ under different interventions, i.e., $\mu_{Y,l}^{t,do(i)}=\bbE[Y^{t,do(i)}_{l}], {\sigma^2}_{Y,l}^{t,do(i)}=Var(Y^{t,do(i)}_{l})$ as follows, 
\begin{equation}
\begin{aligned}
    \label{eq:esttheta max}
    (\hat{\mu}_{Y,l}^{t,do(i),*}, \hat{\sigma^2}_{Y,l}^{t,do(i),*} )= \arg \max_{\nu \in \bbR, s>0}\sum_{m = t-w}^{t-1} \log \phi_1(s^{-1/2} (Y_l^{m,do(a_m^*)}-\nu)) \bbI\{a_m^* = i\} \\ 
    \text{for} \quad i \in 0\cup [p], \ l\in[p] \backslash i,
\end{aligned}
\end{equation}
where $\hat{\mu}_{Y,l}^{t,do(i),*}$ and $ \hat{\sigma^2}_{Y,l}^{t,do(i),*}$ are the estimation of $\mu_{Y,l}^{t,do(i)}$ and ${\sigma^2}_{Y,l}^{t,do(i)}$ respectively. With $\hat{\mu}_{Y,l}^{t,do(i),*}$ and $ \hat{\sigma^2}_{Y,l}^{t,do(i),*}$ for $i \in 0\cup [p]$, we can get the estimated distribution of node $l$ as 
$\hat{f}_{i}^{t,\Delta,[k,j],*}[l]$ for $l \in [p]\backslash i$. 

Similar to \eqref{eq:W multi} and \eqref{eq:W multi 2}, we define the corresponding max-CUSUM type detection statistic for each node $l$ as 
\begin{equation}
\begin{aligned}
\label{eq:W max}
    \Lambda_{t,A^*}^{\Delta,[k,j],*}[l]  &= \begin{cases}
    \log\left(\frac{\hat{f}^{t,\Delta,[k,j],*}_{a_t^{*}}[l](Y_l^{t,do(a_t^{*})})}{f^{0}_{a_t^{*}}[l](Y_l^{t,do(a_t^{*})})} \right), &\text{for  } l \in 0\cup[p]\backslash a_t^{*}, \\
    0, &\text{for  } l = a_t^{*}.
    \end{cases}\\
    W_{t,A^*}^l &= \max\{W_{t-1,A^*}^l ,0\} + \Lambda_{t,A^{*}}^{\Delta,[k,j],*}[l], \ t>w \quad W_{t,A^*}^l = 0 \ \ \text{for  } l\in [p].
\end{aligned}
\end{equation}
Last we select the maximum of these test statistics as our final detection statistic,
\begin{equation}
\begin{aligned}
\label{eq:W max 1}
W_{t,A^{*}} &= \max_{l\in[p]} W_{t,A^{*}}^l.
\end{aligned}
\end{equation}
According to Proposition \ref{prop: Ia}, under Assumption \ref{ass:single change}, centralization effectively concentrates the change onto a single node. Therefore, using the maximum test statistic across all nodes as the monitoring criterion is justified.
The system triggers a change alarm at  $T_{b,A^*}^{max} = \inf\{t>w|W_{t,A^*}>b\}$, where $b$ is a constant threshold. Note that if we define a change alarm for each node at $T_{b,A^*}^l = \inf\{t>w| W_{t,A^*}^l >b\}$, we can equivalently get the same system change alarm at $T_{b,A^*}^{max} = \min_{l\in[p]} T_{b,A^*}^l$.

Using the same sequence $\sN$ defined in \nameaa,  we present the corresponding adaptive intervention policy $A^{*} = \{a^*_t|t \in \bbN\}$ as follows, 
\begin{enumerate}[i.]
    \item For $t = 1,\dots,w$, $a_t^{*} = V^t$ where $V^t$ is defined in the same way as in \nameaa.
    \item If $t>w$ and $t\in\sN$, $a_t^* = V^t$.
    \item If $t>w$ and $t\notin \sN$, we can
    use $\{\hat{\mu}_{Y,l}^{t,do(i),*}, \hat{\sigma^2}_{Y,l}^{t,do(i),*}\}_{i\in0\cup p, l\in[p]\backslash i}$ to estimate the KL divergence. Then, according to \eqref{eq:prop2 2}, we set
    \begin{equation}
    \label{eq:inter max}
         \begin{aligned}
            a_t^* &= \arg\max_{a\in0\cup[p]} \{\max_{l\in[p]}\hat{I}_a^{t,\Delta,[k,j],*}[l] \}  = \arg\max_{a\in 0\cup[p]}\{\max_{l\in[p]} D(\hat{f}_a^{t,\Delta,[k,j],*}[l]||f_a^{0}[l] ) \}
        \end{aligned}
    \end{equation}
\end{enumerate}

It is to be noted that since we do not need to estimate covariance between different nodes, we suggest $w$ of \nameb to be at least $O(p)$ to ensure the accuracy of $\{(\hat{\mu}_{Y,l}^{t,do(i),*}, \hat{\sigma^2}_{Y,l}^{t,do(i),*} )\}_{i\in0\cup[p],l\in [p]\backslash i}$. During exploitation, we estimate each node's distribution of $\bmY$ obtained after intervening on node $i$, $i\in0\cup[p]$. Then according to Proposition \ref{prop: Ia}, we select the node $i$, under which the KL divergence of a single node, can be the largest, as the intervention node for the next time.

\begin{remark1}
\label{rem: max vs multi}
 The advantage of \nameb over \namea relies on Proposition \ref{prop: Ia} that the change to be concentrated in a single node's KL divergence, based on the validity of Assumption \ref{ass:single change}. In other words, if Assumption \ref{ass:single change} does not hold, which means more than one edge in the system change, the property of concentrating change information in one node no longer applies. Consequently, the detection effectiveness of \nameb would decrease compared to \nameaa. This result is also confirmed in our simulation study of Section \ref{sec:multi_change}. In conclusion, if we are confident that Assumption \ref{ass:single change} holds, we can use \nameb which requires a shorter window. However, if we lack prior information about the number of changed edges, \namea may be a better choice to achieve a more robust detection effect.
\end{remark1}

We summarize \nameb and \namea in Algorithm \ref{alg:WLCUSUM}, using red to represent \namebb, blue for \nameaa, and black for the shared components of the two methods.

\begin{algorithm}
        \caption{\textcolor{red}{\nameb} and \textcolor{blue}{\namea}}
	\label{alg:WLCUSUM}
    \begin{algorithmic}
        \REQUIRE $\bmu,\bSigma,\bA,\Delta_{min},\bmC,\sN,w,b$
        \ENSURE \textcolor{red}{$T_{b,A^{*}}^{max}$} (\textcolor{blue}{$T_{b,A^{\circ}}^{multi}$})
        \FOR{$t = 1,2,\cdots,w$ } 
        \STATE Intervene on node $V^t$,  get observation $\bmX^{t,do(X_{V^t}^t = c_{V^t})}$.\\ 
        \STATE Centralize $\bmX^{t,do(X_{V^t}^t = c_{V^t})}$ by \eqref{eq:central} to get $\bmY^{t,do(V_t)}$.
        \ENDFOR
        \FOR{$t = w+1,w+2,\dots$}
        \STATE Estimate parameter \textcolor{red}{$\{(\hat{\mu}_{Y,j}^{t,do(i),*}, \hat{\sigma^2}_{Y,j}^{t,do(i),*} )\}_{i\in0\cup[p],j\in [p]\backslash i}$} (\textcolor{blue}{$\{(\hat{\bmu}_Y^{t,do(i),\circ}, \hat{\bSigma}_Y^{t,do(i),\circ} )\}_{i\in0\cup[p]}$}) by \textcolor{red}{\eqref{eq:esttheta max}} (\textcolor{blue}{\eqref{eq:esttheta multi}}).
        \IF{$t \in \sN$}
             \STATE Intervene node $V^t$,  get observation $\bmX^{t,do(X_{V^t}^t = c_{V^t})}$.\\ 
            \STATE Centralize $\bmX^{t,do(X_{V^t}^t = c_{V^t})}$ by \eqref{eq:central} to get $\bmY^{t,do(V_t)}$.
        \ELSE
            \STATE Choose adaptive intervention node \textcolor{red}{$a_t^{*}$} (\textcolor{blue}{$a_t^{\circ}$}) by \textcolor{red}{\eqref{eq:inter max}} (\textcolor{blue}{\eqref{eq:inter multi}}) and get observation \textcolor{red}{$\bmX^{t,do(X^t_{a_t^{*}} = c_{a_t^{*}})}$} (\textcolor{blue}{$\bmX^{t,do(X^t_{a_t^{\circ}} = c_{a_t^{\circ}})}$}). 
            \STATE Centralize the observation by \eqref{eq:central} to get \textcolor{red}{$\bmY^{t,do(a_t^{*})}$} (\textcolor{blue}{$\bmY^{t,do(a_t^{\circ})}$}).
        \ENDIF
        \STATE Calculate the monitoring statistics \textcolor{red}{$W_{t,A^{*}}$} (\textcolor{blue}{$W_{t,A^{\circ}}$}) by \textcolor{red}{\eqref{eq:W max} \eqref{eq:W max 1}} (\textcolor{blue}{\eqref{eq:W multi} \eqref{eq:W multi 2}}).
        \IF{\textcolor{red}{$W_{t,A^{*}} > b$} (\textcolor{blue}{$W_{t,A^{\circ}} > b$})}
            \STATE \textcolor{red}{$T_{b,A^*}^{max} = t$} (\textcolor{blue}{$T_{b,A^{\circ}}^{multi} = t$}).
            \STATE break.
        \ENDIF
        \ENDFOR
    \end{algorithmic}
\end{algorithm}

\section{Theoretical Analysis}
\label{sec: theorem}
In this section, we first introduce the measurement methods for false alarm and detection delay. We then provide an asymptotic lower bound on detection delay for a general intervention policy. Finally, we prove that our proposed \namea and \nameb methods are first-order asymptotically optimal, i.e., having the "quickest" property. 

\subsection{General Lower Bound}
\label{sec: thm sub1}
We first define the filtration $\cF = \{\cF_t | t\in \bbN\} $,
\begin{equation*}
    \cF_t \triangleq \sigma(\bmU^1,\dots,\bmU^t,V^1,\dots,V^t),
\end{equation*}
which is a $\sigma$-algebra that contains the randomness information of the centralized variables up to time $t$ as well as the information on our random interventions. We also define the $\cF_0$ as the trial $\sigma$-algebra. Then we give a general definition of the sequential intervention policy. 
\begin{definition1}[Sequential Intervention policy]
\label{def:policy}
    $A = \{a_t|t\in \bbN \}$ is a sequential intervention policy if it satisfies the following:\\
    1) $\bmU^t$ is conditionally independent of $\cF_{t-1}$ given the value of $a_t$; \\
    2) $a_t$ takes value in $0\cup[p]$; \\
    3) For $t>1$, $a_t$ is a $\cF_{t-1}$ measurable random variable.
\end{definition1}
 We can verify that both $A^{\circ}$ and $A^*$ satisfy Definition \ref{def:policy}. 
 Then we give the general definition of a detection method.
 \begin{definition1}[Detection method]
 \label{def:detection method}
    $(A,T)$ is a detection method if it satisfies the following:\\
    1) $A$ is the sequential intervention policy;\\
    2) $T$ is an $\{\cF_t \}$-stopping time, i.e., $\{T = t\} \in \cF_t$ for $\forall t\in \bbN$.
 \end{definition1}
 We can verify that $(A^{\circ}, T_{b,A^{\circ}}^{multi})$ and $(A^{*},T_{b,A^{*}}^{max})$ all satisfy the Definition \ref{def:detection method}. We use $\cC$ to denote the family of all possible detection methods, i.e., $(A,T) \in \cC$.
 
 To evaluate the performance of a $(A,T) \in \cC$, we define the average run length (ARL) under the pre-change condition, i.e., $\bbE_{\infty, A}[T]$, which should not be very small, to avoid too many false alarms. In our paper, we focus on the detection methods whose ARL should be larger than $\gamma$. 
\begin{definition1}[$\gamma$-false alarm class]
    Given $\gamma > 1$, define 
    \begin{equation*}
    \cC_{\gamma} \triangleq \{(A,T)\in\cC |\bbE_{\infty,A}[T] \geq \gamma\}.
\end{equation*}
where $\bbE_{\infty,A}[\cdot]$ denotes the expectation under the pre-change condition when the sequential intervention policy is $A$
\end{definition1}
We use a worst-case measure to evaluate the performance of a detection method under the post-change condition.
\begin{definition1}[Worst case delay]
Given post-change parameter $\{\Delta,[k,j]\}$ and $(A,T)\in\cC$, define 
\begin{equation*}
    \cJ_{\Delta, [k,j]}(A,T) \triangleq \sup_{\tau \in \bbN} \text{ess} \sup\bbE_{\tau, A}^{\Delta, [k,j]}[\max\{T-\tau+1,0 \} | \cF_{\tau-1} ]
\end{equation*}
where $\bbE_{\tau, A}^{\Delta, [k,j]}[\cdot]$ is the expectation with the sequential intervention policy $A$ under the post-change condition whose change time is $\tau$ and the post-change parameter is $\{\Delta, [k,j]\}$. 
\end{definition1}
Then Theorem \ref{thm:general} gives the lower bound of $\cJ_{\Delta, [k,j]}(A,T)$ for a detection method $(A, T)\in \cC_{\gamma}$ when its intervention values of $A$ are set according to Algorithm \ref{alg: c_j}. 
\begin{theorem1}
\label{thm:general}
    For any post-change $\{\Delta, [k,j]\}$ with $\Delta \in [\Delta_{min},\Delta_{max}]$, by setting the intervention value $\bmC$ via Algorithm \ref{alg: c_j}, under Assumption \ref{ass:single change},
when $\gamma \to \infty$, we have
    \begin{equation}
        \inf_{(A,T) \in \cC_{\gamma}} \cJ_{\Delta, [k,j]} (A,T) \geq \frac{\log\gamma}{I^{\Delta,[k,j]}_{j,c_j}}(1+o(1)).
    \end{equation}
    
\end{theorem1}
\begin{proof}
    The proof is shown in Appendix \ref{app: lower bound}.
\end{proof}
Note that according to Proposition \ref{prop: Ia}, $I_{j,c_j}^{\Delta,[k,j]}$ is the max KL divergence through all intervention nodes, i.e., $I_{j,c_j}^{\Delta,[k,j]} = \max_{i\in0\cup[p]} I_{i,c_i}^{\Delta,[k,j]} $.  Theorem \ref{thm:general} describes the lower bound on detection delay when the false alarm rate for a given detection method is fixed. 
We observe that this lower bound is directly proportional to the logarithm of the run length under the false alarm scenario, and inversely proportional to the maximum KL divergence between the post-change and pre-change distributions, which is intuitively reasonable.

We subsequently prove that $(A^{\circ},T_{b,A^{\circ}})$ and $(A^*, T_{b,A^*})$ can achieve the lower bound in a first-order asymptotic sense as $\gamma, w, q \to \infty$. 
\subsection{First-order Optimality of \namea}
\label{sec: thm sub2}
\begin{lemma1}
\label{lem:multi IC}
For any $\gamma > 1$, by setting $b = \log \gamma$, we have that $(A^{\circ},T_{b,A^{\circ}}^{multi}) \in \cC_{\gamma}$, i.e., $\bbE_{\infty,A^{\circ}}[T_{b, A^{\circ}}^{multi}] > \gamma$.
\end{lemma1}
\begin{proof}
The proof is shown in Appendix \ref{app:multi IC}.
\end{proof}
Lemma \ref{lem:multi IC} theoretically guides the selection of the threshold $b$ to achieve the desired run length for \namea under the pre-change condition.

\begin{lemma1}
\label{lem:multi OC}
For any post-change $\{\Delta, [k,j]\}$ with $\Delta \in [\Delta_{min},\Delta_{max}]$, by setting the intervention value $\bmC$ via Algorithm \ref{alg: c_j}, under Assumption \ref{ass:single change}, for any change-point $\tau \in \bbN$ and threshold $b > 0$, we have
\begin{equation}
    \cJ_{\Delta, [k,j]}(A^{\circ}, T_{b,A^\circ}^{multi}) \leq \sup_{a_1^{\circ},\dots,a_{w}^{\circ}}\bbE_{1,A^{\circ}}^{\Delta, [k,j]}[T_{b,A^\circ}^{multi}| A^{\circ}_{[w]}],
\end{equation}
where $A^{\circ}_{[w]} \triangleq \{a_t^{\circ}|t\in[w] \}$.
\end{lemma1}
\begin{proof}
    The proof is shown in Appendix \ref{app:multi OC}.
\end{proof}
Lemma \ref{lem:multi OC} shows that $\cJ_{\Delta, [k,j]}(A^{\circ}, T_{b,A^\circ}^{multi})$ is no larger than the supremum of expected detection delay under $\tau = 1$. Hence in the subsequent analysis of the upper bound of $\cJ_{\Delta, [k,j]}(A^{\circ}, T_{b,A^\circ}^{multi})$, it suffices to examine the upper bound of the expected detection delay under $\tau = 1$.

It is noted that the above two lemmas are straightforward extensions of \cite{fellouris2022quickest}. However, in \cite{fellouris2022quickest}, the post-change parameters can only take a finite number of values, while in our case the post-change mean and covariance of $\bmY^{t,do(i)}$ can be any value in a continuous bounded space since we assume $\Delta\in[\Delta_{min},\Delta_{max}]$. This distinction renders other key results in \cite{fellouris2022quickest} for finite post-change parameter set no longer applicable in our case, and 
generalizing them to continuous bounded space is nontrival. The following Lemma \ref{lem:multi arm j} and lemma \ref{lem:multi liminf} are our core lemmas. 

We first show that when $q,w \to \infty$, the sequential intervention policy for \namea can find the optimal intervention node, i.e., the origin of the changed edge.

\begin{lemma1}
    \label{lem:multi arm j}
For any post-change $\{\Delta, [k,j]\}$ with $\Delta \in [\Delta_{min},\Delta_{max}]$, by setting the intervention value $\bmC$ via Algorithm \ref{alg: c_j}, under Assumption \ref{ass:single change}, for $t>w, t\notin \sN$ and $q = o(w)$ we have
\begin{equation}
    \lim_{q,w\to\infty}\bbP_{1,A^{\circ}}^{\Delta,[k,j]}(a_t^{\circ} = j|\cF_{t-w-1}, A^\circ_{[w]}) = 1,
\end{equation}
where $\bbP_{1, A^{\circ}}^{\Delta, [k,j]}[\cdot]$ is the probability measure with sequential intervention policy $A^{\circ}$ under the post-change condition $\{\Delta, [k,j]\}$ and $\tau=1$.  
\end{lemma1}
\begin{proof}
    The proof is shown in Appendix \ref{app:multi arm j}.
\end{proof}

Then we give the asymptotic property for $ \Lambda_{t,A^\circ}^{\Delta,[k,j],\circ}$. 

\begin{lemma1}
\label{lem:multi liminf}
For any post-change $\{\Delta, [k,j]\}$ with $\Delta \in [\Delta_{min},\Delta_{max}]$, 
by setting the intervention value $\bmC$ via Algorithm \ref{alg: c_j}, under Assumptions \ref{ass:single change}, for $t > w$ we have
\begin{equation}
    \liminf_{q,w\to\infty}\bbE_{1,A^{\circ}}^{\Delta, [k,j]}\left[\Lambda_{t,A^{\circ}}^{\Delta, [k,j],\circ}| \cF_{t-w-1}, A^{\circ}_{[w]}\right] \geq \begin{cases}
        I^{\Delta,[k,j]}_{j,c_j} > 0, & \text{for } t \notin \sN \\
        \frac{1}{p+1}\sum_{i=0}^p I^{\Delta,[k,j]}_{i,c_i} > 0, & \text{for } t \in \sN 
    \end{cases}
\end{equation}
\end{lemma1}
\begin{proof}
    The proof is shown in Appendix \ref{app:multi liminf}.
\end{proof}

Lemma \ref{lem:multi liminf}  states that, as $q\to \infty$, the expected value of the likelihood ratio statistic calculated using the estimated parameters $\{\hat{\bmu}_Y^{t,do(i),\circ}, \hat{\bSigma}_Y^{t,do(i),\circ}\}_{i\in0\cup[p]}$ is at least as large as the maximum expected value of the likelihood ratio computed based on the true parameter $\{(\bbE[\bmY^{t,do(i)}_{[p]\backslash i}], Cov(\bmY^{t,do(i)}_{[p]\backslash i}))\}_{i\in0\cup[p]}$ since $I_{j,c_j}^{\Delta,[k,j]} = \max_{i\in0\cup[p]} \bbE\left[\log\left(\frac{f_i^{\Delta, [k,j]}(\bmY^{t,do(i)})}{f_i^{0}(\bmY^{t,do(i)})} \right)\right]$. 

Based on Lemmas \ref{lem:multi IC}-\ref{lem:multi liminf}, we can prove \namea is first-order optimal, as shown in Theorem \ref{thm:multi}.

\begin{theorem1}
\label{thm:multi}
For any post-change $\{\Delta, [k,j]\}$ with $\Delta \in [\Delta_{min},\Delta_{max}]$, 
by setting the intervention value $\bmC$ via Algorithm \ref{alg: c_j}, under Assumption \ref{ass:single change}, if $b = \log\gamma$, then as $q,w,\gamma \to \infty$ so that $q = o(w)$ and $w = o(\log\gamma)$, we have
\begin{equation}
    \cJ_{\Delta, [k,j]}(A^{\circ}, T_{b,A^\circ}^{multi}) \sim_{\infty}  \inf_{(A,T) \in \cC_{\gamma}} \cJ_{\Delta, [k,j]} (A,T) \sim_{\infty} \frac{\log\gamma}{I_{j,c_j}^{\Delta,[k,j]}}.
\end{equation}
\end{theorem1}
\begin{proof}
    The proof is shown in Appendix \ref{app:multi thm}.
\end{proof}

\subsection{First-order Optimality of \nameb}
\label{sec: thm sub3}
The procedure to prove \nameb is first-order optimal is similar to that of \namea. The difference is that \nameb takes the maximum of each node's statistic when constructing the final detection statistic. In the proof process, we need to handle this aspect separately.

\begin{lemma1}
\label{lem:max IC}
    For any $\gamma > 1$, by setting $b = \log\gamma + \log p$, we have that $(A^*, T_{b,A^*}^{max}) \in \cC_{\gamma}$, i.e., $\bbE_{\infty,A^*}[T_{b,A^*}^{max}] > \gamma$.
\end{lemma1}
\begin{proof}
    The proof is shown in Appendix \ref{app:max IC}.
\end{proof}
Note that in Lemma \ref{lem:max IC}, $b=\log\gamma + \log p$ differs from the value of $b$ in Lemma \ref{lem:multi IC}. This discrepancy arises from the additional ``max'' operation. 

To discuss properties of the detection delay of \namebb, given that $T_{b,A^*}^{max} = \min_{l\in[p]}T_{b,A^*}^l$, we can first examine the upper bound of $T_{b,A^*}^l$, to characterize the upper bound of $T_{b,A^*}^{max}$. Based on Proposition \ref{prop: Ia}, we know that when the change occurs at $[k, j]$, the centralized change is concentrated at $Y_{k}^{t,do(a_t^*)}$. Therefore we first discuss the properties of $T_{b,A^*}^k$.

\begin{lemma1}
\label{lem:max OC}
For any post-change $\{\Delta, [k,j]\}$ with $\Delta \in [\Delta_{min},\Delta_{max}]$, by setting the intervention value $\bmC$ via Algorithm \ref{alg: c_j}, under Assumption \ref{ass:single change}, for any change-point $\tau \in \bbN$ and threshold $b > 0$, we have
    \begin{equation}
        \cJ_{\Delta,[k,j]}(A^{*}, T_{b,A^*}^{k}) \leq \sup_{a_1^{*},\dots,a_w^{*}}\bbE_{1,A^*}^{\Delta,[k,j]}[T_{b,A^*}^{k}| A^{*}_{[w]} ],
    \end{equation}
    where $A^*_{[w]} \triangleq \{a_t^*|t\in[w] \}$.
\end{lemma1}
\begin{proof}
    The proof is shown in Appendix \ref{app:max OC}.
\end{proof}
Similar to Lemma \ref{lem:multi OC}, Lemma \ref{lem:max OC} also shows that $ \cJ_{\Delta,[k,j]}(A^{*}, T_{b,A^*}^{k})$ is no larger than the supremum of expected detection delay under $\tau = 1$. Hence in the subsequent analysis of the upper bound of $ \cJ_{\Delta,[k,j]}(A^{*}, T_{b,A^*}^{k})$, it also suffices to examine the upper bound of the expected detection delay under $\tau = 1$.

Similar to the last section, we first show that when $q,w \to \infty$, the sequential intervention policy for \nameb can find the optimal intervention node, i.e., the origin of the changed edge.
\begin{lemma1}
\label{lem:max arm j}
For any post-change $\{\Delta, [k,j]\}$ with $\Delta \in [\Delta_{min},\Delta_{max}]$, by setting the intervention value $\bmC$ via Algorithm \ref{alg: c_j}, under Assumption \ref{ass:single change}, for $t>w, t\notin \sN$ and $q = o(w)$ we have
\begin{equation}
    \lim_{q,w\to\infty}\bbP_{1,A^{*}}^{\Delta,[k,j]}(a_t^{*} = j|\cF_{t-w-1}, A^*_{[w]}) = 1,
\end{equation}
where $\bbP_{1, A^{*}}^{\Delta, [k,j]}[\cdot]$ is the probability measure with sequential intervention policy $A^{*}$ under the post-change condition $\{\Delta, [k,j]\}$ and $\tau=1$.  
\end{lemma1}
\begin{proof}
    The proof is shown in Appendix \ref{app:max arm j}.
\end{proof}
Then we give the asymptotic property for $ \Lambda_{t,A^*}^{\Delta,[k,j],*}[l]$. 

\begin{lemma1}
\label{lem:max liminf}
For any post-change $\{\Delta, [k,j]\}$ with $\Delta \in [\Delta_{min},\Delta_{max}]$, 
by setting the intervention value $\bmC$ via Algorithm \ref{alg: c_j}, under Assumption \ref{ass:single change}, for $t > w$, $l\in[p]$ and $q = o(w)$, we have
\begin{equation}
    \label{eq:max liminf}\liminf_{q,w\to\infty}\bbE_{1,A^{*}}^{\Delta,[k,j]}\left[\Lambda_{t,A^{*}}^{\Delta,[k,j],*}[l]| \cF_{t-w-1}, A^*_{[w]}\right] \geq \begin{cases}
        I^{\Delta,[k,j]}_{j,c_j}[l], & \text{for } t \notin \sN \\
        \frac{1}{p+1}\sum_{i=0}^p I^{\Delta,[k,j]}_{i,c_i}[l], & \text{for } t \in \sN 
    \end{cases}
\end{equation}
\end{lemma1}
\begin{proof}
The proof is shown in Appendix \ref{app:max liminf} .
\end{proof}


Lemma \ref{lem:max liminf}  states that, when a change occurs in $[\bA]_{k,j}$, for $t\notin \sN$ and node $k$,   as $q,w\to \infty$, the expected value of the likelihood ratio statistic calculated using the estimated parameters $\{\hat{\mu}_{Y,k}^{t,do(i),*}, \hat{\sigma^2}_{Y,k}^{t,do(i),*}\}_{i\in0\cup[p]}$ is at least as large as the maximum expected value of the likelihood ratio computed based on the true parameter $\{(\bbE[Y^{t,do(i)}_{k}], Cov(Y^{t,do(i)}_{k}))\}_{i\in0\cup[p]}$ since $I_{j,c_j}^{\Delta,[k,j]} = I_{j,c_j}^{\Delta,[k,j]}[k] = \max_{i\in0\cup[p]} \bbE\left[\log\left(\frac{f_i^{\Delta, [k,j]}[k](Y^{t,do(i)}_k)}{f_i^{0}[k](Y^{t,do(i)}_k)} \right)\right]$. 

Based on Lemmas \ref{lem:max IC}-\ref{lem:max liminf}, we can prove \nameb is first-order optimal, as shown in Theorem \ref{thm:max}.

\begin{theorem1}
\label{thm:max}
For any post-change $\{\Delta, [k,j]\}$ with $\Delta \in [\Delta_{min},\Delta_{max}]$, 
by setting the intervention value $\bmC$ via Algorithm \ref{alg: c_j}, under Assumption \ref{ass:single change}, if $b = \log\gamma$, then as $q,w,\gamma \to \infty$ so that $q = o(w)$ and $w = o(\log\gamma)$, we have
\begin{equation}
    \cJ_{\Delta,[k,j]}(A^{*}, T_{b,A^*}^{max}) \sim_{\infty}  \inf_{(A,T) \in \cC_{\gamma}} \cJ_{\Delta,[k,j]} (A,T) \sim_{\infty} \frac{\log\gamma}{I_{j,c_j}^{\Delta,[k,j]}}.
\end{equation}
\end{theorem1}
\begin{proof}
    The proof is shown in Appendix \ref{app:max thm}.
\end{proof}

\begin{remark1}
    In Lemma \ref{lem:max liminf}, when $t \notin \sN$, for node $k$, the lower bound on the right-hand side of \eqref{eq:max liminf} is essentially $I_{j,c_j}^{\Delta,[k,j]}[k]$. Since \eqref{eq:prop2 4} holds, we have $I_{j,c_j}^{\Delta,[k,j]} = I_{j,c_j}^{\Delta,[k,j]}[k]$.
    However, suppose changes occur in multiple edges (i.e., Assumption \ref{ass:single change} does not hold). We cannot represent all the change information with a single dimension, leading to $I_{j,c_j}^{\Delta,[k,j]}[k] < I_{j,c_j}^{\Delta,[k,j]}$. Consequently, the first-order optimality of \nameb in Theorem \ref{thm:max} is no longer guaranteed. This issue does not arise for \namea. So as mentioned in Remark \ref{rem: max vs multi},  \namea is more robust than \nameb.
\end{remark1}

\subsection{Discussions of Theorem \ref{thm:multi} and Theorem \ref{thm:max}}
Recall the definition $I_{j,c_j}^{\Delta,[k,j]}$ in \eqref{eq:Ia detal}. It is associated with the intervention value $c_j$. As long as we increase the value of $c_j$, the intervention gap $\delta$ in Algorithm \ref{alg: c_j} can be increased. Consequently, $I_{j,c_j}^{\Delta,[k,j]}$ can be increased and the expected detection delay in Theorems \ref{thm:multi} and \ref{thm:max} can be reduced. However, if larger intervention values typically incur higher intervention costs, it may introduce a trade-off between the cost of intervention and the cost of achieving shorter detection delays, which deserves further study in the future. 

If the adaptive intervention policies $A^{\circ}$ in \namea and $A^{*}$ in \nameb are replaced by either the no-intervention policy $A^0 \triangleq \{ a_t^0 = 0| t\in\bbN\}$ or the random intervention policy $A^r \triangleq \{a^r_t = V^t|t\in\bbN \}$, 
we can get four counterparts multi-no-intervention (MULTI-NI), multi-random-intervention (MULTI-RI), max-no-intervention (MAX-NI) and max-random-intervention (MAX-RI).
Using similar derivations in Theorem \ref{thm:multi} and Theorem \ref{thm:max}, we can only get $\cJ_{\Delta,[k,j]}(A^0, T_{b,A^0}^{multi}) \leq \frac{\log\gamma}{I_{0}^{\Delta,[k,j]}}(1+o(1))$ when $(A^0,T_{b,A^0}^{multi}) \in \cC_{\gamma}$ as $\gamma \to \infty$; $\cJ_{\Delta,[k,j]}(A^r, T_{b,A^r}^{multi}) \leq \frac{\log\gamma}{\frac{1}{p+1}\sum_{i=0}^p I^{\Delta,[k,j]}_{i,c_i}}(1+o(1))$ when $(A^r,T_{b,A^r}^{multi}) \in \cC_{\gamma}$ as $\gamma \to \infty$;
$\cJ_{\Delta,[k,j]}(A^0, T_{b,A^0}^{max}) \leq \frac{\log\gamma}{I_{0}^{\Delta,[k,j]}[k]}(1+o(1))$ when $(A^0,T_{b,A^0}^{max}) \in \cC_{\gamma}$ as $\gamma \to \infty$ and $\cJ_{\Delta,[k,j]}(A^r, T_{b,A^r}^{max}) \leq \frac{\log\gamma}{\frac{1}{p+1}\sum_{i=0}^p I^{\Delta,[k,j]}_{i,c_i}[k]}(1+o(1))$ when $(A^r,T_{b,A^r}^{max}) \in \cC_{\gamma}$ as $\gamma \to \infty$.
Since we know $I_{i,c_i}^{\Delta,[k,j]} + \delta < I_{j,c_j}^{\Delta,[k,j]}$ and $I_{i,c_i}^{\Delta,[k,j]}[k] + \delta < I_{j,c_j}^{\Delta,[k,j]}[k] = I_{j,c_j}^{\Delta,[k,j]}$ for $i\in0\cup[p]\backslash j$, we can find MULTI-NI, MULTI-RI, MAX-NI and MAX-RI cannot achieve the lower bound in Theorem \ref{thm:general}. This means that our adaptive intervention policies, $A^{\circ}$ and $A^*$ are useful compared with no intervention policy $A^0$, and random intervention policy $A^r$. We will also discuss and compare these methods in the next section using simulations. 


\section{Simulation}
\label{sec: simulation}
In this section, we validate the performance of our methods through simulation results under different scenarios. We also compare our proposed two methods, \nameb and \namea with four baseline methods: (1) MAX-NI, which replaces the adaptive intervention policy $A^{*}$ in \nameb with $A^0 = \{a_t^0 = 0| t\in \bbN\}$; (2) MAX-RI, which replaces the adaptive intervention policy $A^{*}$ in \nameb with a random policy $A^r = \{a_t^r = V^t| t\in \bbN \}$; (3) MULTI-NI, which replaces the adaptive intervention policy $A^{\circ}$ in \namea with $A^0$; (4) MULTI-RI, which replaces the adaptive intervention policy $A^{\circ}$ in \namea with a random policy $A^r$. In this section, we first describe the data generation process, and discuss the performance of different methods in the following scenarios: (1) different number of nodes $p$, (2) different change magnitude $\Delta$, (3) different exogenous background variance $\bSigma$, (4) different intervention gap $\delta$, and (5) cases where the single-change assumption in Assumption \ref{ass:single change} is not satisfied.

\subsection{Data Generation}
Define the outward degree of a node $j$, $d_{j}^{out}$, as the number of edges that point out from it and the inward degree $d_j^{in}$ as the number of edges that point into it. Then, define $d = \max_{j\in[p]} \max \{d_j^{in},d_j^{out} \}$ to characterize the sparsity of the causal graph. First, we discuss how to generate $\bA$. For a given number of nodes $p$, we consider all $p(p-1)$ possible edges. When generating a specific edge, $[k,j]$, we create the edge with probability $\Tilde{p} = 0.5$, under the condition that adding the edge neither creates a cycle nor increases $d$ beyond a specified value. The edge weight $[\bA]_{k,j}$ is drawn from a uniform distribution over $[1,2]$. Then each element in $\bmu$ is independently generated from a uniform distribution over $[-1,1]$, and each diagonal element of $\bSigma$ is independently generated from a uniform distribution $[1/2,2]$ unless otherwise specified. We set $\Delta_{min} = 0.1$ and $\Delta_{max} = 2$. For $\sN$, we set $\eta = 1$ and randomly generate sequential $\sN$ for each run. The change magnitude is set to be a given value $\Delta$, and the location of the change is randomly generated, ensuring no cycles are introduced, i.e., $\Tilde{\bA} = \bA + \Delta \bE^{k,j}$. We set $\Tilde{\bmu} = \bmu$ and $\Tilde{\bSigma} = \bSigma$ since we only consider the change in $\bA$ as we discuss the properties of Proposition \ref{prop:Yt} earlier. We set the intervention value $\bmC$ via Algorithm \ref{alg: c_j}. Then we can generate $X^{t,do(X_i^t = c_i)}$ according to \eqref{eq:intervention X} and \eqref{eq:intervention X 2} for different intervention nodes under pre-change or post-change conditions. According to Lemma \ref{lem:multi OC} and Lemma \ref{lem:max OC}, we consider the worst-case for the post-change condition, i.e., $\tau = 1$. For each parameter setting, we perform 1000 independent runs and use the detection delays of the 1000 runs to calculate the expected detection delay (EDD). We can control the value of $b$ to observe how the EDD increases with ARL. This allows us to evaluate the detection efficiency under different false alarm conditions.

\subsection{Varying Number of Nodes $p$} 
\label{sec:node_number}
We set the number of nodes as $p = 4$, $6$, and $12$. For the case  $p = 4$, we examine two settings for $d$ (i.e., $d=2$ and $d=3$) to explore different sparsity levels. Additionally, for each parameter setting, we configure two time-window settings, i.e., a short window with $w = 20$ and $q = 10$,
and a long window with $w = 40$ and $q = 20$—to study the algorithm’s performance under varying time frames. Similarly, for  $p = 6$, we set $d = 2$ or $d = 3$, with a short window of $w = 30$ and $q = 15$, and a long window of $w = 80$ and $q = 40$. Finally, for $p = 12$, we set $ d = 3 $ or $d = 4$, with a short window of $w = 60$ and $q = 30$, and a long window of $w = 300$ and $q = 150$. 
Therefore, for each $p$, we have four parameter settings, resulting in a total of 12 cases. In each case, we set $\Delta = 0.1$ and $\delta = 1$. 

\begin{figure}[h]
    \centering
    \includegraphics[width=\linewidth]{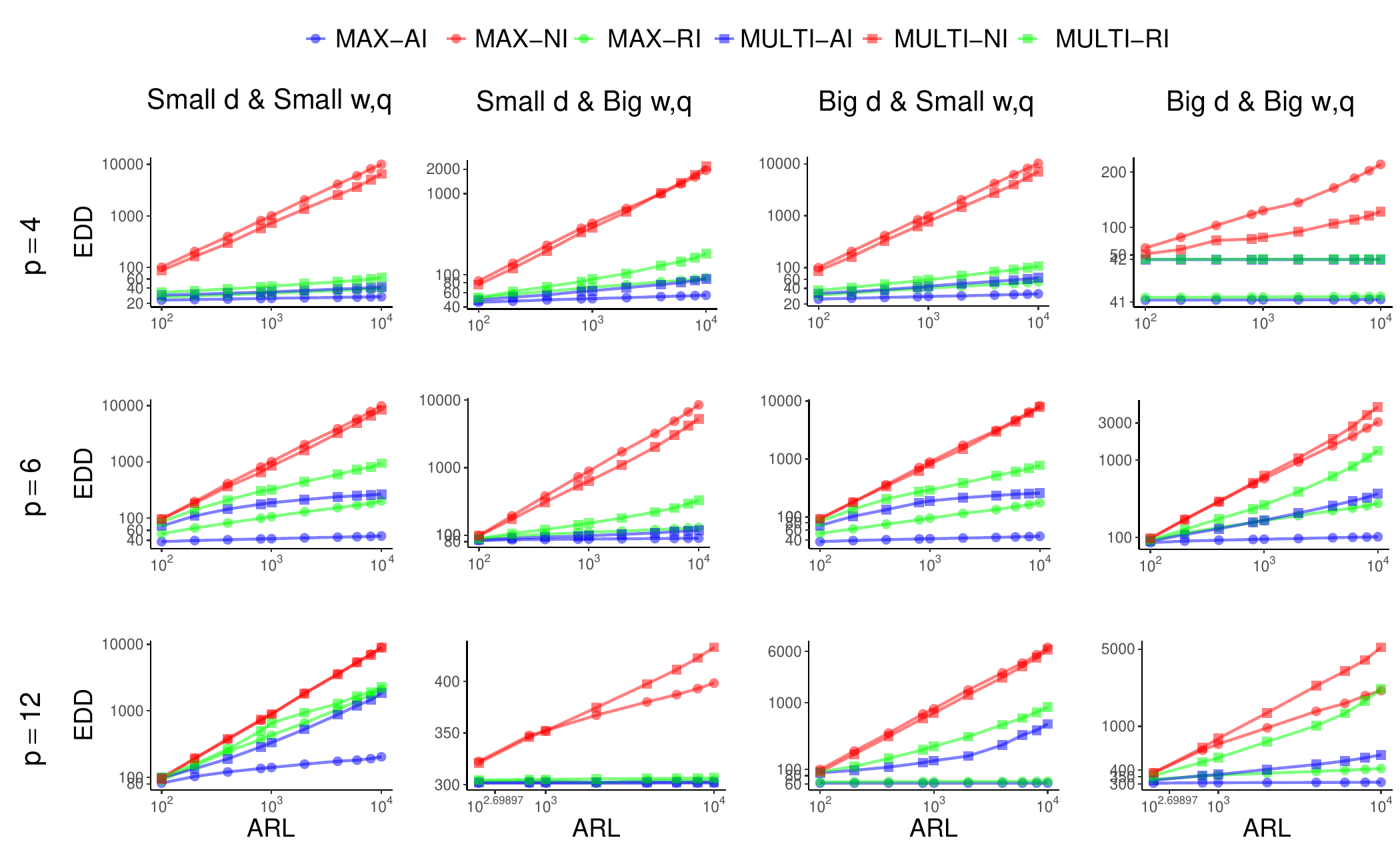}
    \caption{EDD v.s. ARL for varying number of nodes $p$. Different rows represent different $p$. Different columns represent different graph sparsity $d$ and time window $w$.}
    \label{fig:sim p}
\end{figure}

We plot the EDD versus ARL of each case in Figure \ref{fig:sim p}. \nameb outperforms MAX-NI and MAX-RI across all the settings, and \namea also demonstrates superior performance compared to MULTI-NI and MULTI-RI. This is because, when the change magnitude is relatively small, detecting such subtle change without intervention is challenging, and the random intervention policy $A^r$ is unable to identify the optimal intervention nodes, leading to poorer performance. Furthermore, when the time window is short, the MAX-type methods significantly outperform the MULTI-type methods. However, as $w$ increases, this difference is remarkably reduced, and the effectiveness of the MULTI-type methods improves considerably. This improvement occurs because the MULTI-type methods involve estimating more parameters, which requires a larger $w$ for reliable parameter estimation $\theta$, as previously discussed in Remark \ref{rem: max vs multi}.

\subsection{Varying Change Magnitude $\Delta$}
\label{sec:magnitude}
We set the change magnitude as $\Delta = 0.1, 0.2$, and $0.5$. For other parameters, we set $p=6$, and $d=2$ or $3$. We still consider two time-window settings, i.e., a short window with $w=30$ and $q =15$, and a long time window with $w=80$ and $q = 40$. We set $\delta = 1$. 

\begin{figure}[h]
    \centering
    \includegraphics[width=\linewidth]{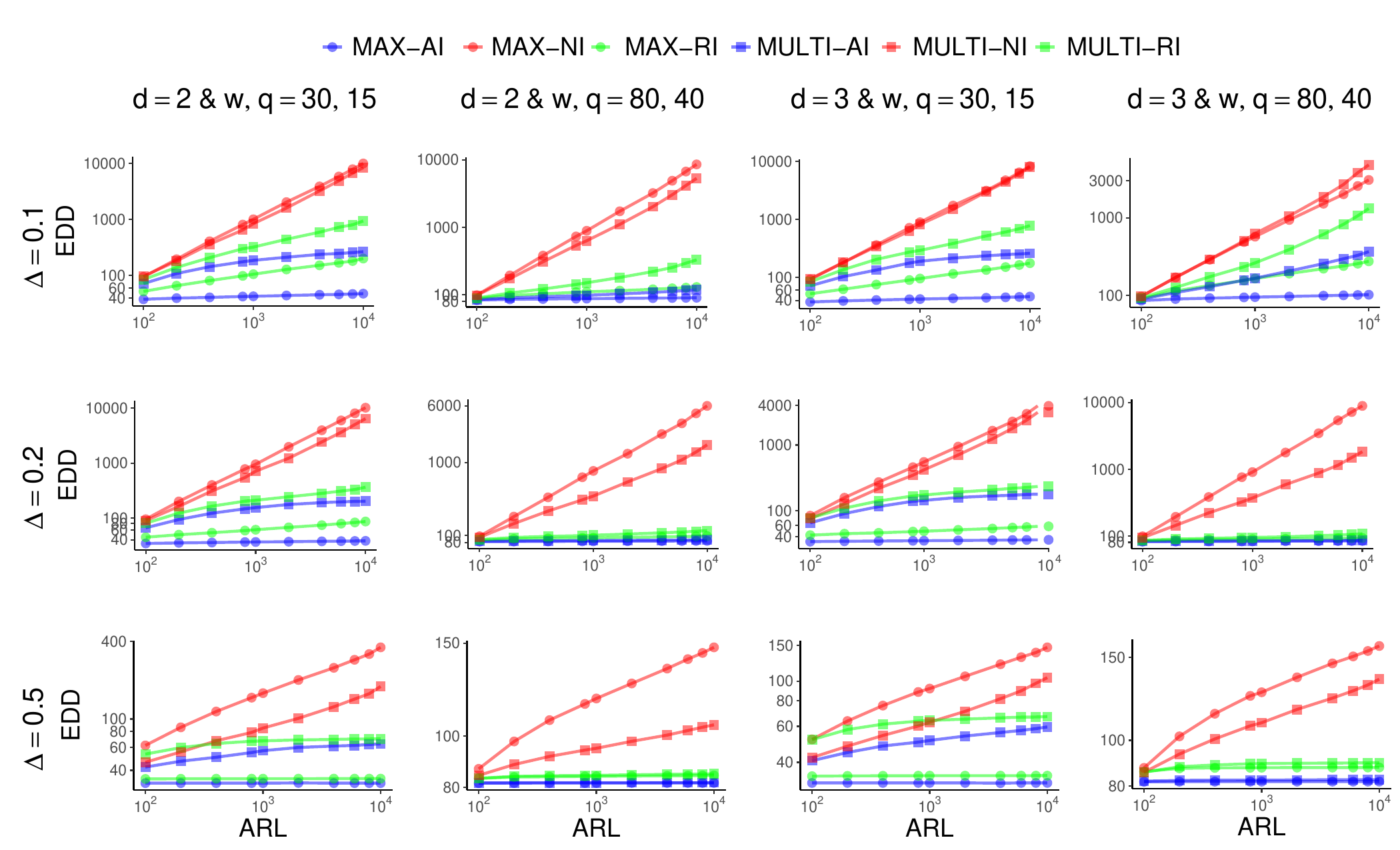}
    \caption{EDD v.s. ARL for varying change magnitude $\Delta$. Different rows represent different $\Delta$. Different columns represent different graph sparsity $d$ and time window $w$.}
    \label{fig:sim Delta}
\end{figure}

We plot the EDD versus ARL of each case in Figure \ref{fig:sim Delta}. we observe that the relative performance of the different methods remains consistent with the descriptions in Section \ref{sec:node_number}. As $\Delta$ increases, the EDDs for all the methods decrease, and \nameb still demonstrates the best performance. This result is intuitive, as larger changes are easier to detect.

\boldmath
\subsection{Varying Exogenous Background Variance $\Sigma$}
\unboldmath
\label{sec:vary_sigma}
We design two settings for $\bSigma$. In the first one, each diagonal element of $\bSigma$ is drawn from the uniform distribution over $[1/2,2]$. In the second one, each diagonal element of $\bSigma$ is drawn from a uniform distribution over $[2,4]$. For other parameters the same as Section \ref{sec:magnitude}, we set $p,d, w$ and $q$ in the same way as Section \ref{sec:magnitude}. In each case, we set $\Delta = 0.1$ and $\delta = 1$.
\begin{figure}[h]
    \centering
    \includegraphics[width=\linewidth]{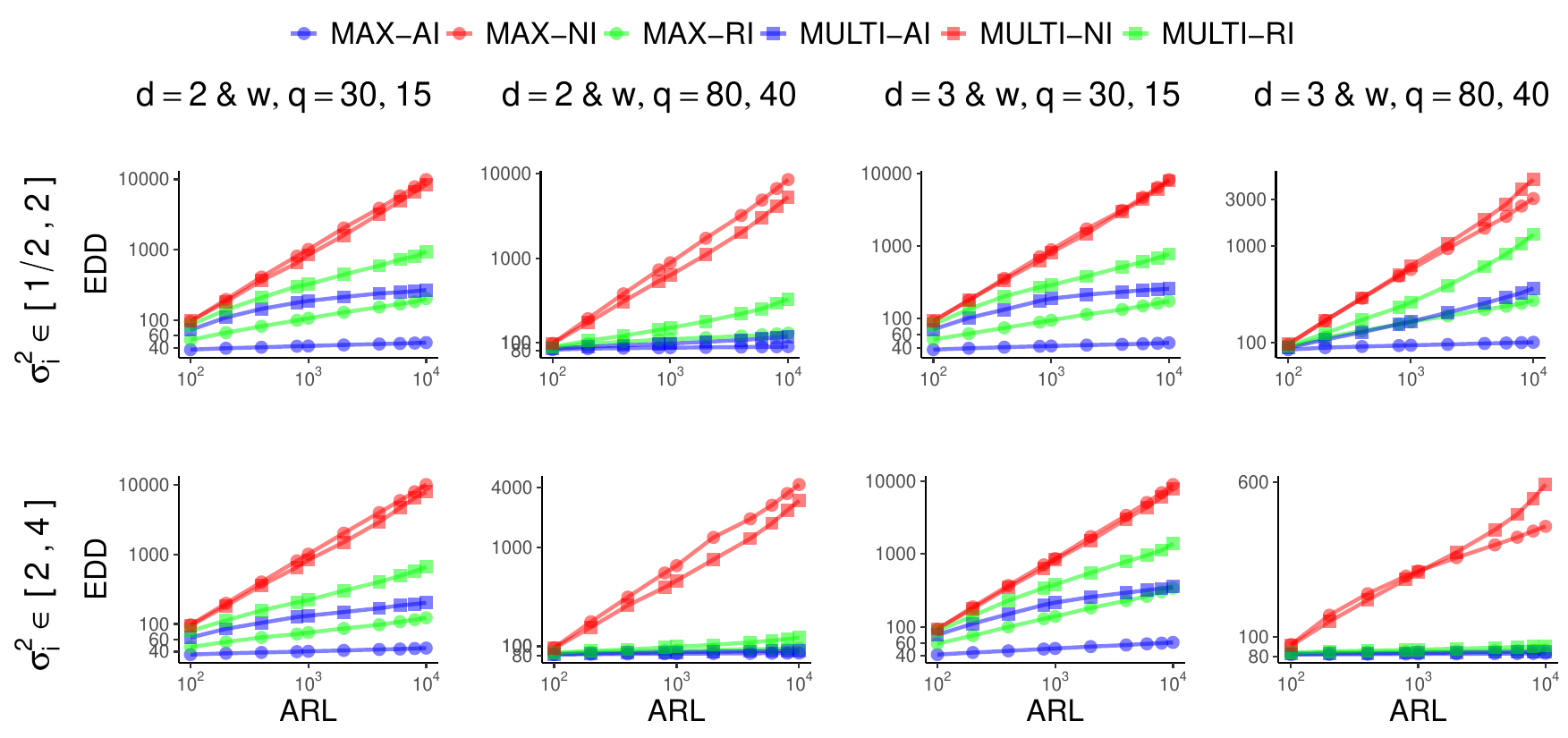}
    \caption{EDD v.s. ARL for varying exogenous background variance $\bSigma$. Different rows represent different $\bSigma$. Different columns represent different graph sparsity $d$ and time window $w$.}
    \label{fig:sim Sigma}
\end{figure}

 We plot the EDD versus ARL of each case in Figure \ref{fig:sim Sigma}. \nameb remains the best among all the methods, and \namea performs best among all the MULTI-type methods, whose performance becomes better with larger $w$, consistent with findings from the previous subsection. When the variance $\bSigma$ is increasing, we observe that EDD either increases or decreases, with no clear trend. According to Proposition \ref{prop:Yt} and \eqref{eq:Ia detal}, we can get $I_{j,c_j}^{\Delta,[k,j]} = \frac{1}{2}(1+ \frac{c_j\Delta^2}{\sigma_k^2})$. However, since $c_j$ is set by Algorithm \ref{alg: c_j}, it may also increase when $\sigma_k$ increases. Therefore, the ratio of $c_j$ to $\sigma_k^2$ may have no clear increase or decrease trend, so as to EDD.

\subsection{Varying Intervention Gap $\delta$}
\label{sec:multi_change}
Theoretically, as $\delta$ increases, the intervention values $\bC$ generally increase, and the gap between the KL divergence of the optimal intervention node and those of other intervention nodes becomes larger, making a better detection performance. We set $\delta$ as $0.5, 1$, and $1.5$. For other parameters the same as Section \ref{sec:magnitude}, we set $p,d, w$ and $q$ in the same way as Section \ref{sec:magnitude}. In each case, we set $\Delta = 0.1$. 

\begin{figure}[h]
    \centering
    \includegraphics[width=\linewidth]{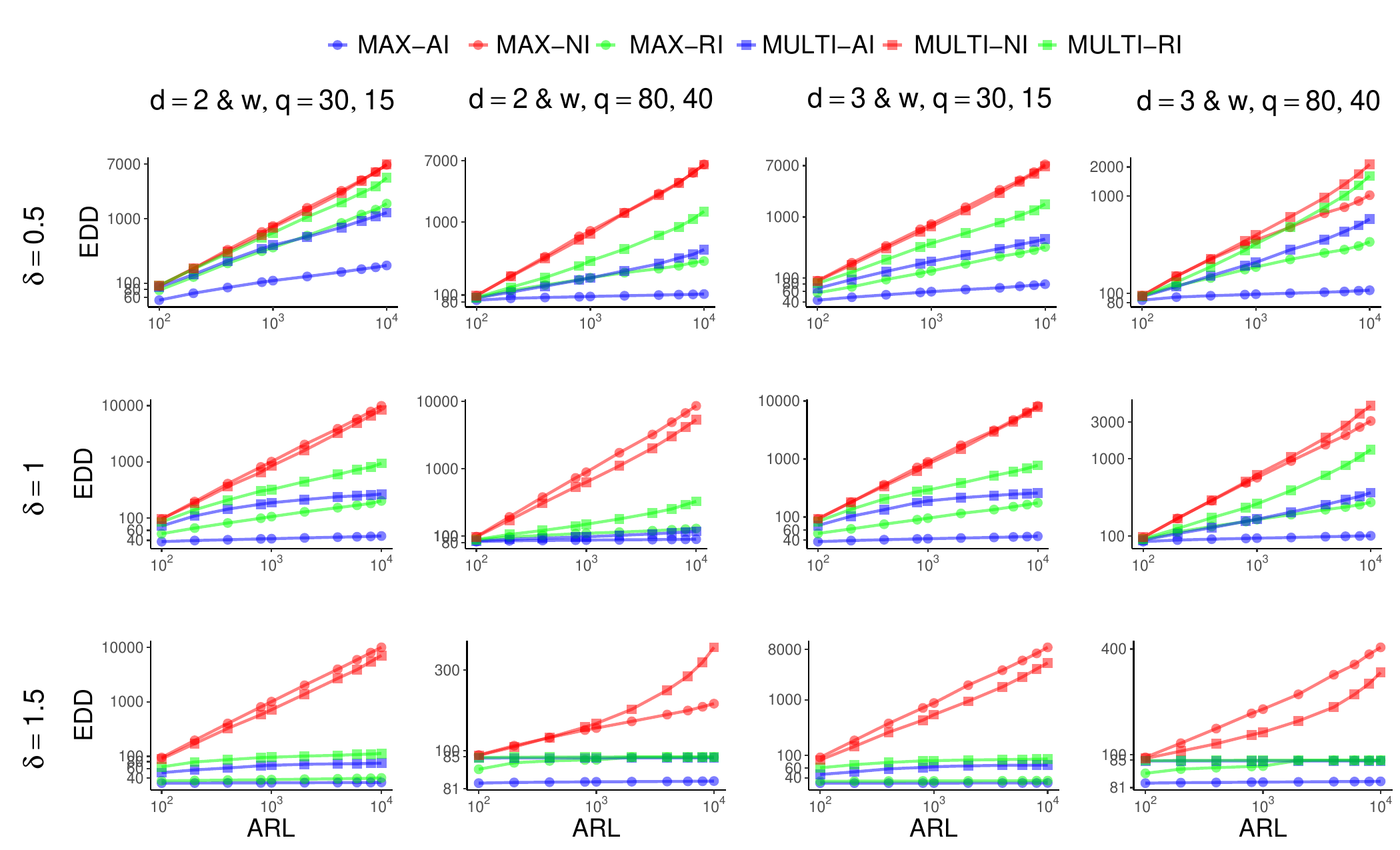}
    \caption{EDD v.s. ARL for varying intervention gap $\delta$. Different rows represent different $\delta$. Different columns represent different graph sparsity $d$ and time window $w$.}
    \label{fig:sim delta}
\end{figure}

We plot the EDD versus ARL of each case in Figure \ref{fig:sim delta}. The relative performance of all the methods remains consistent with the previous subsections. As expected, when $\delta$ increases, the EDDs of \nameb, \namea, MAX-RI and MULTI-RI decrease, while the EDDs of MAX-NI and MULTI-NI remain unchanged, as they do not involve any interventions.

\subsection{Multiple Changes}
\label{sec:multi_change}
In this subsection, we consider two specific cases which have more than one changed edges, i.e., Assumption \ref{ass:single change} does not hold. In the first case,  we consider a causal graph with four nodes, and set $\bmu = \Tilde{\bmu} = (1,1,1,1)^T$, $\bSigma = \Tilde{\bSigma} = \bI_4$, and 
\begin{equation*}
    \bA = \begin{pmatrix} 
0 &0&0&0\\
1&0&0&0\\
0&1&0&0\\
0&0&1&0
\end{pmatrix}, \Tilde{\bA} = \begin{pmatrix} 
0 &0&0&0\\
1.2&0&0&0\\
0&1.14&0&0\\
0&0&1.1&0
\end{pmatrix}.
\end{equation*} 
For other parameters, we set $w = 40, q=20$. In the second case, we consider a causal graph with five nodes, and set $\bmu = \Tilde{\bmu} = (1,1,1,1,1)^T$, $\bSigma = \Tilde{\bSigma} = \bI_5$, and 
\begin{equation*}
    \bA = \begin{pmatrix} 
0 &0&0&0&0\\
1&0&0&0&0\\
0&1&0&0&0\\
1&0&0&0&0\\
0&0&0&1&0
\end{pmatrix}, \Tilde{\bA} = \begin{pmatrix} 
0 &0&0&0&0\\
1.15&0&0&0&0\\
0&1.1&0&0&0\\
1.15&0&0&0&0\\
0&0&0&1.1&0
\end{pmatrix}.
\end{equation*}
For other parameters, we set $w = 60, q=30$. For both cases, we consider two values of $\delta=0.5,1$. 

\begin{figure}[h]
    \centering
    \includegraphics[width=0.95\linewidth]{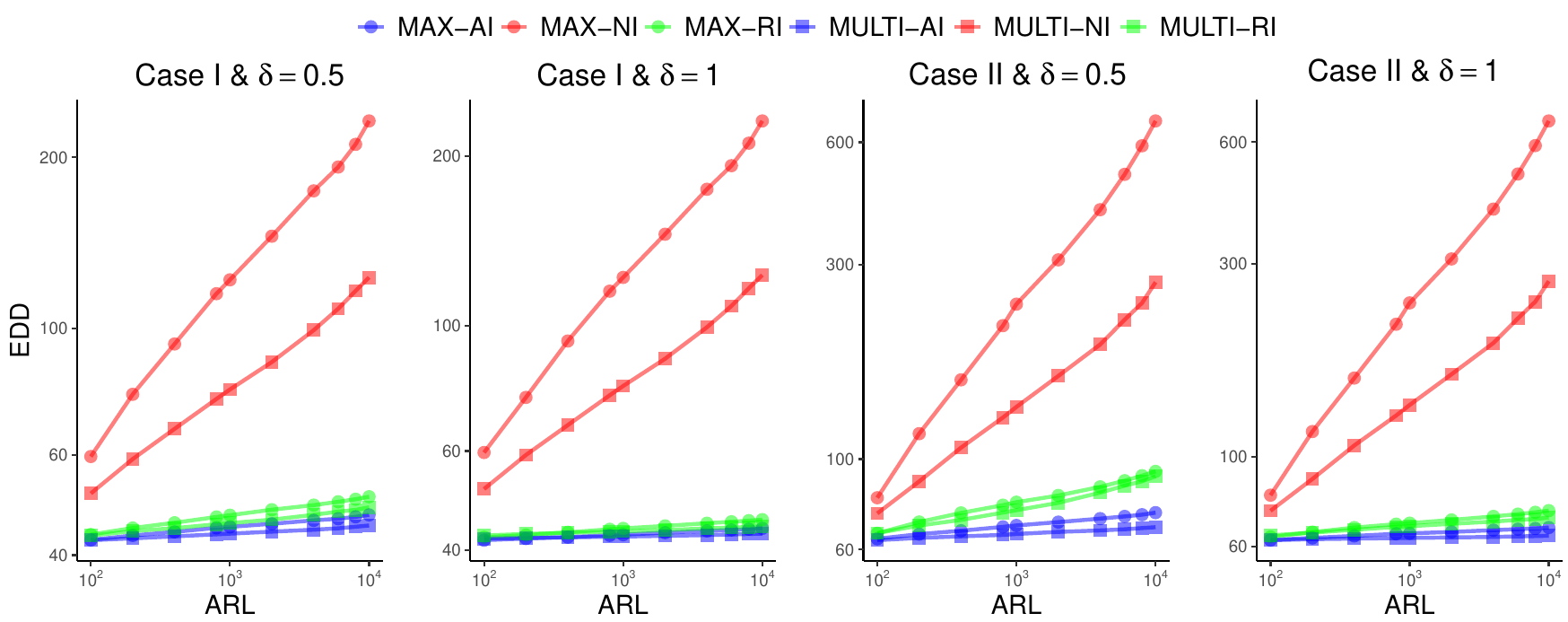}
    \caption{EDD v.s. ARL of multiple changes setting. Different rows represent different change cases and different $\delta$.}
    \label{fig:sim multi change}
\end{figure}

We plot EDD versus ARL in Figure \ref{fig:sim multi change}. \namea performs the best among all the methods, followed by \namebb. This is because Assumption \ref{ass:single change} is not satisfied, 
and the property of Proposition \ref{prop: Ia}, which assumes changes are concentrated in a single dimension, no longer holds. So \nameb fails to accurately select the optimal intervention node, leading to poorer monitoring performance. Yet the difference between \namea and \nameb is not very significant. 
\section{Case Studies}
\label{sec: case}
\subsection{Case in Ecology}
\label{sec:ecology}
This case study investigates the causal relationship between 11 environmental variables (including seafloor temperature and seawater pH) and net ecosystem calcification (NEC), using a linear SEM model\citep{courtney2017environmental}. The associated DAG structure and information about these environmental variables are shown in Appendix \ref{app:ecology}. By intervening on certain environmental variables, we aim to detect whether the system occurs anomalies and the causal relationships between different variables have changed. A similar case study is also used in \cite{aglietti2020causal}.
We primarily consider two types of changes: 1) the causal effect from seafloor temperature to NEC; 2) the causal effect from seawater pH to NEC. 
We generate the data based on the parameters provided in \citep{courtney2017environmental}. For each change location, we consider three different change magnitudes, $\Delta = 0.1,0.2,0.5$, and two time window settings, a short time window $w=60$ and $q=30$ and a long time window $w=300$ and $q= 150$. For other parameters, we set $\delta= 1$ and $\eta = 1$.

\begin{figure}[h]
    \centering
    \includegraphics[width=\linewidth]{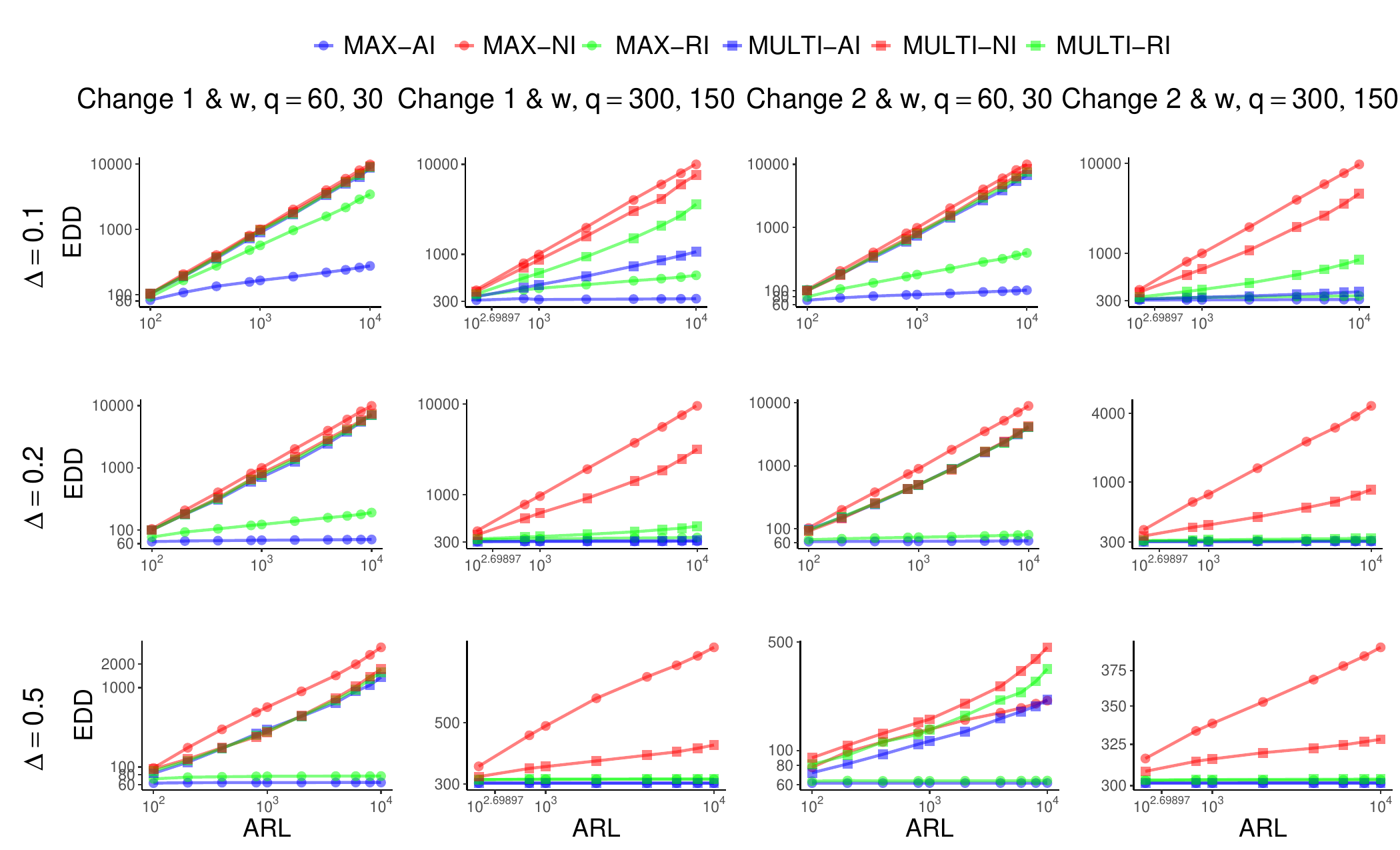}
    \caption{EDD v.s. ARL of the ecological case. Different rows represent different change magnitude $\Delta$. Different columns represent different change locations and time window $w$.}
    \label{fig:case 1}
\end{figure}
The results are shown in Figure \ref{fig:case 1}.  Under different parameter settings, \nameb consistently performs the best, which is in line with the previous simulation results. Although \namea does not perform well under the short time window setting, this issue is significantly improved when the time window is extended. Overall, \nameb and \namea significantly outperform other baselines with no-intervention policies and random intervention policies.

\subsection{Case in Psychology}
\label{sec:psychology}
This case study uses a casual graph to describe the five-facet mindfulness model, which suggests that mindfulness includes five facets, i.e., observing, describing, nonjudging, nonreactivity, and acting with awareness \citep{heeren2021network}. The associated causal graph is shown in Appendix \ref{app:psychology} and the specific parameters can be found in the \cite{heeren2021network}. In this case, these five facets are measured through scores from a relevant questionnaire. We can treat the intervention as psychological counseling, aiming to increase their score in a specific facet and then observe the scores of other facets. This allows us to assess whether a change has occurred in any causal relationships among the five facets. We consider two types of changes: 1) the causal effect from acting with awareness to nonjudging; 2) the causal effect from nonreactivity to observing. For each type of change, we consider three change magnitudes $\Delta = 0.1, 0.2, 0.3$, and two time window settings, a short time window with $w=25$ and $q=12$, a long time window with $w=60$ and $q=30$. For other parameters, we set $\delta = 1$ and $\eta = 1$. 
\begin{figure}[h]
    \centering
    \includegraphics[width=\linewidth]{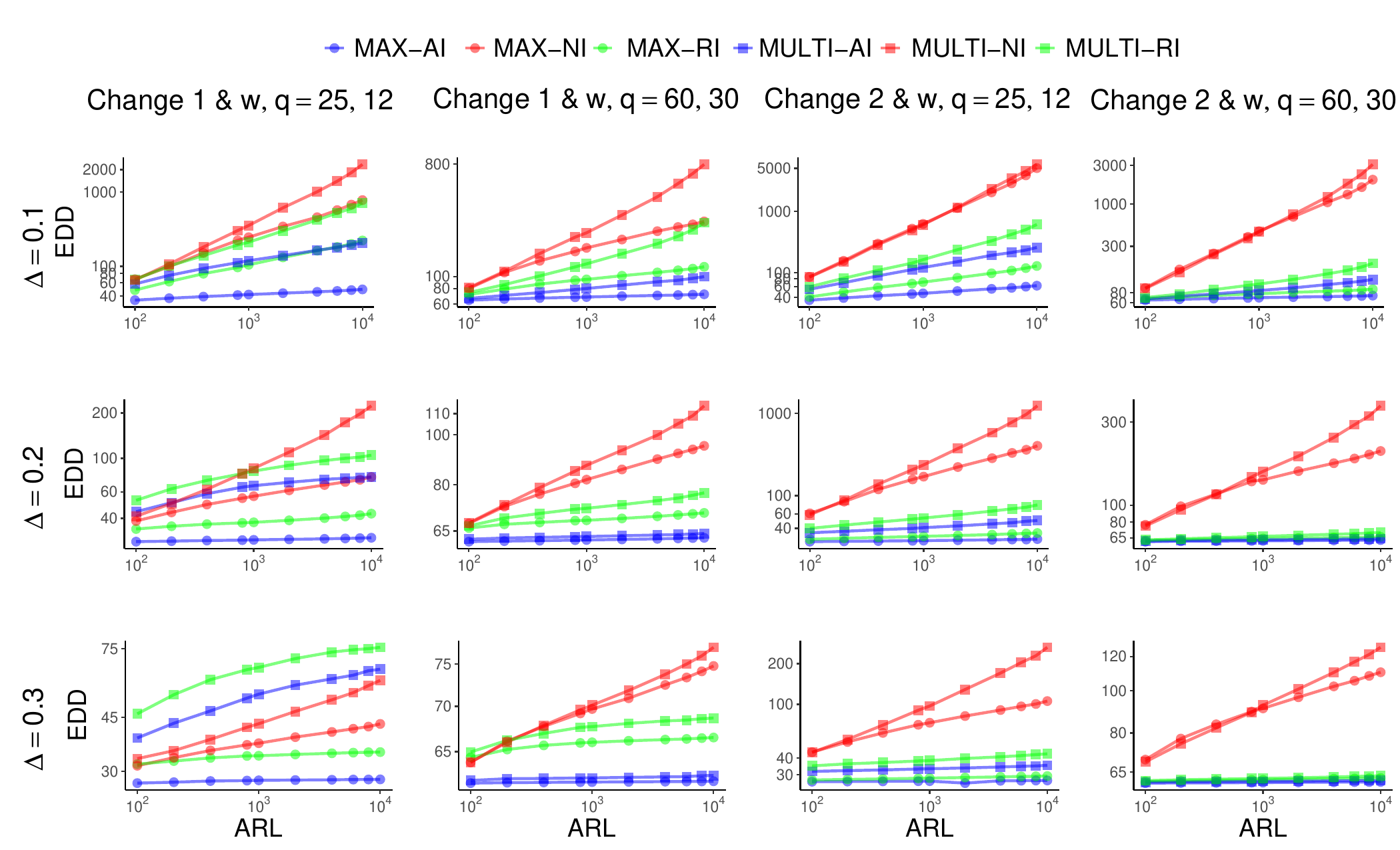}
    \caption{EDD v.s. ARL of the psychological case. Different rows represent different  change magnitude $\Delta$. Different columns represent different change locations and time window $w$.}
    \label{fig:case 2}
\end{figure}

The results are shown in Figure \ref{fig:case 2}. The performance of different methods are consistent with those from previous sections. Both \nameb and \namea outperform the other baselines, demonstrating the effectiveness of our proposed intervention policies.
\section{Conclusion}
\label{sec: conclusion}
This paper targets sequential change point detection in linear causal models, and is the first work to consider introducing intervention for increasing detection efficiency. We first novelly propose an intervention mechanism. Its core contribution is by delivering a rule for setting the intervention values for each node, the origin of the change edge, which proves to be the best node to intervene on, can be selected naturally, even though in reality we do not know which edge has changed or which node is the origin of the changed edge. Based on it, we further design two detection methods, \namea and \nameb, to address different types of changes. For each detection method, the detection statistic and the adaptive intervention policy are introduced, and its theoretical expected detection delay are analyzed. 

In future, many extensions deserve further study, such as when multiple nodes can be intervened on simultaneously, when different interventions have different costs while the total budget is limited, etc, when the causal graph has time-lagged edges and intervention on one node will influence other nodes' future data.

\newpage

\appendix
\section{Proof of Proposition \ref{prop:Yt}}
\label{app: prop 1}



In this section, we prove Proposition \ref{prop:Yt} in Section \ref{sec: intervention sub2}.

For convenient, we use $\bB^{do(X_i = c_i)}$ to denote $(\bI_p - \bA^{do(X_i = c_i)})^{-1}$ and use $\Tilde{\bB}^{do(X_i = c_i)}$ to denote $(\bI_p - \Tilde{\bA}^{do(X_i = c_i)})^{-1}$.
Recall that $X^{t,do(X_i^t = c_i)} \sim N(\bmu_x^{do(X_i =c_i)},\bSigma_x^{do(X_i=c_i)})$ when $t < \tau$ where $\bmu_x^{do(X_i = c_i)} = \bB^{do(X_i = c_i)} \bmu^{do(X_i = c_i)}$ and $\bSigma_x^{do(X_i = c_i)} = \bB^{do(X_i = c_i)} \bSigma^{do(X_i = c_i)} (\bB^{do(X_i = c_i)})^T$.  $X^{t,do(X_i^t = c_i)} \sim N(\Tilde{\bmu}_x^{do(X_i =c_i)},\Tilde{\bSigma}_x^{do(X_i=c_i)})$ when $t \geq \tau$ where $\Tilde{\bmu}_x^{do(X_i = c_i)} = \Tilde{\bB}^{do(X_i = c_i)} \Tilde{\bmu}^{do(X_i = c_i)}$ and $\Tilde{\bSigma}_x^{do(X_i = c_i)} = \Tilde{\bB}^{do(X_i = c_i)}\Tilde{\bSigma}^{do(X_i = c_i)} (\Tilde{\bB}^{do(X_i = c_i)})^T$. For convenient, we use $do(i)$ as an abbreviation for $do(X_i = c_i)$. 

\begin{proof}

Since $\bmY^{t,do(i)}_{[p]\backslash i}$ is a linear combination of $\bmX^{t,do(i)}_{[p]\backslash i}$, we have that $\bmY^{t,do(i)}_{[p]\backslash i}$ also follows the multivariate normal distribution. Next, we only need to prove that its mean and covariance matrix are consistent with those in Proposition \ref{prop:Yt}.

When $t <\tau$, since $\bbE[\bmX^{t,do(i)}] = \bmu_x^{do(i)}$ and $Cov(\bmX^{t,do(i)}) = \bSigma_x^{do(i)}$,we have 
\begin{equation*}
    \begin{aligned}
        \bbE[\bmY^{t,do(i)}_{[p]\backslash i}] = [\bSigma^{do(i)}]^{-1/2}_{[p]\backslash i, [p]\backslash ,i}  [(\bB^{do(i)})^{-1}]_{[p]\backslash i, [p]\backslash ,i} (\bbE(\bmX^{t,do(i)})_{[p]\backslash i} - \bmu^{do(i)}_{x,[p]\backslash i}) = 0, 
    \end{aligned}
\end{equation*} and
\begin{equation*}
    \begin{aligned}
        &Cov(\bmY^{t,do(i)}_{[p]\backslash i}) = [\bSigma^{do(i)}]^{-1/2}_{[p]\backslash i, [p]\backslash ,i} [(\bB^{do(i)})^{-1}]_{[p]\backslash i, [p]\backslash ,i} [\bSigma_x^{do(i)}]_{[p]\backslash i, [p]\backslash ,i} 
        [(\bB^{do(i)})^{-1}]^T_{[p]\backslash i, [p]\backslash ,i} 
        ([\bSigma^{do(i)}]^{-1/2}_{[p]\backslash i, [p]\backslash ,i} )^T \\
        &\quad=[\bSigma^{do(i)}]^{-1/2}_{[p]\backslash i, [p]\backslash ,i} [(\bB^{do(i)})^{-1}]_{[p]\backslash i, [p]\backslash ,i} 
        [\bB^{do(i)}\bSigma^{do(i)}(\bB^{do(i)})^T]_{[p]\backslash i, [p]\backslash ,i} 
        [(\bB^{do(i)})^{-1}]^T_{[p]\backslash i, [p]\backslash ,i} 
        ([\bSigma^{do(i)}]^{-1/2}_{[p]\backslash i, [p]\backslash ,i} )^T \\
       &\quad = \bSigma^{do(i)}]^{-1/2}_{[p]\backslash i, [p]\backslash ,i} [(\bB^{do(i)})^{-1}]_{[p]\backslash i, [p]\backslash ,i} 
        [\bB^{do(i)}]_{[p]\backslash i, [p]\backslash ,i} [\bSigma^{do(i)}]_{[p]\backslash i, [p]\backslash ,i} \\
        & \qquad [(\bB^{do(i)})^T]_{[p]\backslash i, [p]\backslash ,i} 
        [(\bB^{do(i)})^{-1}]^T_{[p]\backslash i, [p]\backslash ,i} 
        ([\bSigma^{do(i)}]^{-1/2}_{[p]\backslash i, [p]\backslash ,i} )^T \\
        &\quad =\bI_{|[p]\backslash i|}
    \end{aligned}
\end{equation*}
where the third equality holds because $[\bB^{do(i)}]_i = \bmE^i$ and $[\bSigma^{do(i)}]_{i,i} = 0$ for $
i \neq 0$.

When $t \geq \tau$ and change occurs in $\mu_k$, i.e., $\Tilde{\bA} = \bA, \Tilde{\bSigma} = \bSigma, \Tilde{\bmu} = \bmu + \Delta \bmE^k$, we can find that $\bSigma_x^{do(i)} = \Tilde{\bSigma}_x^{do(i)}$ since $\bSigma = \Tilde{\bSigma}$ and $\bA = \Tilde{\bA}$. So the covariance matrix of $\bmY^{t,do(i)}_{[p]\backslash i}$ is the same as under pre-change condition. For the mean vector, we have 
\begin{equation*}
    \begin{aligned}
        \bbE[\bmY^{t,do(i)}_{[p]\backslash i}] &= [\bSigma^{do(i)}]^{-1/2}_{[p]\backslash i, [p]\backslash ,i}  [(\bB^{do(i)})^{-1}]_{[p]\backslash i, [p]\backslash ,i} (\Tilde{\bmu}^{do(i)}_{x,[p]\backslash i} - \bmu^{do(i)}_{x,[p]\backslash i})  \\ 
         &= [\bSigma^{do(i)}]^{-1/2}_{[p]\backslash i, [p]\backslash ,i}  [(\bB^{do(i)})^{-1}]_{[p]\backslash i, [p]\backslash ,i} [\bB^{do(i)}]_{[p]\backslash i, [p]\backslash ,i} (\Tilde{\bmu}^{do(i)} - \bmu^{do(i)})_{[p] \backslash i}\\
          &= [\bSigma^{do(i)}]^{-1/2}_{[p]\backslash i, [p]\backslash ,i} (\Delta \bmE^k)_{[p]\backslash i} = (\sigma_k^{-1}\Delta \bmE^k)_{[p]\backslash i},
    \end{aligned}
\end{equation*}
where the second equality holds because $[\bB^{do(i)}]_i = \bmE^i$ and $\Tilde{\mu}_i^{do(i)} - \mu_i^{do(i)} = 0$ for $i \neq 0$.

When $t \geq \tau$  and change occurs in $[\bSigma]_{k,k}$, i.e. $\Tilde{\bA} = \bA, \Tilde{\bmu} = \bmu, \Tilde{\bSigma} = \bSigma + \Delta \bE^{k,k}$,  we can find that $\bmu_x^{do(i)} = \Tilde{\bmu}_x^{do(i)}$ since $\bmu = \Tilde{\bmu}$ and $\bA = \Tilde{\bA}$. So the mean of $\bmY^{t,do(i)}_{[p]\backslash i}$ is the same as under pre-change condition. For the covariance matrix, we have
\begin{equation*}
    \begin{aligned}
        &Cov(\bmY^{t,do(i)}_{[p]\backslash i}) = [\bSigma^{do(i)}]^{-1/2}_{[p]\backslash i, [p]\backslash ,i} [(\bB^{do(i)})^{-1}]_{[p]\backslash i, [p]\backslash ,i} [\Tilde{\bSigma}_x^{do(i)}]_{[p]\backslash i, [p]\backslash ,i} 
        [(\bB^{do(i)})^{-1}]^T_{[p]\backslash i, [p]\backslash ,i} 
        ([\bSigma^{do(i)}]^{-1/2}_{[p]\backslash i, [p]\backslash ,i} )^T \\
        &\quad=[\bSigma^{do(i)}]^{-1/2}_{[p]\backslash i, [p]\backslash ,i} [(\bB^{do(i)})^{-1}]_{[p]\backslash i, [p]\backslash ,i} 
        [\bB^{do(i)}\Tilde{\bSigma}^{do(i)}(\bB^{do(i)})^T]_{[p]\backslash i, [p]\backslash ,i} 
        [(\bB^{do(i)})^{-1}]^T_{[p]\backslash i, [p]\backslash ,i} 
        ([\bSigma^{do(i)}]^{-1/2}_{[p]\backslash i, [p]\backslash ,i} )^T \\
       &\quad = [\bSigma^{do(i)}]^{-1/2}_{[p]\backslash i, [p]\backslash ,i}  [\bSigma^{do(i)} +\Delta \bE^{k,k}]_{[p]\backslash i, [p]\backslash ,i} 
        ([\bSigma^{do(i)}]^{-1/2}_{[p]\backslash i, [p]\backslash ,i} )^T \\
        &\quad =(\bI_{p} + \sigma_k^{-2}\Delta\bE^{k,k})_{[p]\backslash i, [p] \backslash i}.
    \end{aligned}
\end{equation*}

When $t \geq \tau$ and change occurs in $[\bA]_{k,j}$, i.e. $\Tilde{\bmu} = \bmu, \Tilde{\bSigma} = \bSigma,\Tilde{\bA} = \bA + \Delta\bE^{k,j} $, we first consider the mean vector, 
\begin{equation}
\label{eq:prop1 pf1}
    \begin{aligned}
        \bbE[\bmY^{t,do(i)}_{[p]\backslash i}] &= [\bSigma^{do(i)}]^{-1/2}_{[p]\backslash i, [p]\backslash ,i}  [(\bB^{do(i)})^{-1}]_{[p]\backslash i, [p]\backslash ,i} (\Tilde{\bmu}^{do(i)}_{x,[p]\backslash i} - \bmu^{do(i)}_{x,[p]\backslash i})  \\
        &=[\bSigma^{do(i)}]^{-1/2}_{[p]\backslash i, [p]\backslash ,i}  [(\bB^{do(i)})^{-1}]_{[p]\backslash i, [p]\backslash ,i} ((\Tilde{\bB}^{do(i)} - \bB^{do(i)}) \bmu)_{[p]\backslash i} \\
        & = [\bSigma^{do(i)}]^{-1/2}_{[p]\backslash i, [p]\backslash ,i} \left((\bB^{do(i)})^{-1}\Tilde{\bB}^{do(i)} - \bI_p \right)_{[p]\backslash i, [p]} \bmu
    \end{aligned}
\end{equation}
where the third equation holds because $[\bB^{do(i)}]_i = [\Tilde{\bB}^{do(i)}]_i$ for $i\neq 0$.  Then we only need to discuss the gap between $\bI_p$ and $(\bB^{do(i)})^{-1}\Tilde{\bB}^{do(i)}$. Through some algebraic calculations, we find that $(\bB^{do(i)})^{-1}$ and $(\Tilde{\bB}^{do(i)})^{-1}$ differ only in the $(k,j)$th element, while the differences between $\bB^{do(i)}$ and $\Tilde{\bB}^{do(i)}$  are all located in the $l$th row where $l \in des(k)$. In this way, we have that 
\begin{equation}
\label{eq:prop1 pf2}
    [(\bB^{do(i)})^{-1}\Tilde{\bB}^{do(i)}]_{l,m} = \begin{cases}
        1 & \text{for } l = m \\
        \Delta [\bB^{do(i)}]_{j,m} &\text{for } l = k \text{ and } m \in anc(j)\cup j\\
        0 & \text{otherwise}.
    \end{cases}
\end{equation}

Combine \eqref{eq:prop1 pf1} and \eqref{eq:prop1 pf2}, we have 
\begin{equation*}
    \bbE[\bmY^{t,do(i)}_{[p]\backslash i}] = \left(\sigma_k^{-1}\Delta \left(\mu^{do(i)}_j+ \sum_{l\in anc(j)} [\bB^{do(i)}]_{j,l}\mu^{do(i)}_l  \right)\bmE^k\right)_{[p]\backslash i}.
\end{equation*}

Then we discuss the covariance matrix of $Y_{[p]\backslash i}^{t,do(i)}$,
\begin{equation}
\label{eq:prop1 pf3}
    \begin{aligned}
        &Cov(\bmY^{t,do(i)}_{[p]\backslash i}) = [\bSigma^{do(i)}]^{-1/2}_{[p]\backslash i, [p]\backslash ,i} [(\bB^{do(i)})^{-1}]_{[p]\backslash i, [p]\backslash ,i} [\Tilde{\bSigma}_x^{do(i)}]_{[p]\backslash i, [p]\backslash ,i} 
        [(\bB^{do(i)})^{-1}]^T_{[p]\backslash i, [p]\backslash ,i} 
        ([\bSigma^{do(i)}]^{-1/2}_{[p]\backslash i, [p]\backslash ,i} )^T \\
        &\quad=[\bSigma^{do(i)}]^{-1/2}_{[p]\backslash i, [p]\backslash ,i} [(\bB^{do(i)})^{-1}]_{[p]\backslash i, [p]\backslash ,i} 
        [\Tilde{\bB}^{do(i)}\bSigma^{do(i)}(\Tilde{\bB}^{do(i)})^T]_{[p]\backslash i, [p]\backslash ,i} 
        [(\bB^{do(i)})^{-1}]^T_{[p]\backslash i, [p]\backslash ,i} 
        ([\bSigma^{do(i)}]^{-1/2}_{[p]\backslash i, [p]\backslash ,i} )^T \\
        &\quad =[\bSigma^{do(i)}]^{-1/2}_{[p]\backslash i, [p]\backslash ,i} [(\bB^{do(i)})^{-1}]_{[p]\backslash i, [p]\backslash ,i} [\Tilde{\bB}^{do(i)}]_{[p]\backslash i, [p]\backslash ,i} [\bSigma^{do(i)}]^{1/2}_{[p]\backslash i, [p]\backslash ,i}  \\
        & \qquad \left([\bSigma^{do(i)}]^{-1/2}_{[p]\backslash i, [p]\backslash ,i} [(\bB^{do(i)})^{-1}]_{[p]\backslash i, [p]\backslash ,i} [\Tilde{\bB}^{do(i)}]_{[p]\backslash i, [p]\backslash ,i} [\bSigma^{do(i)}]^{1/2}_{[p]\backslash i, [p]\backslash ,i} \right)^T,
    \end{aligned}
\end{equation}
Note that $[(\bB^{do(i)})^{-1}]_{[p]\backslash i, [p]\backslash ,i} [\Tilde{\bB}^{do(i)}]_{[p]\backslash i, [p]\backslash ,i}  = [(\bB^{do(i)})^{-1}\Tilde{\bB}^{do(i)}]_{[p]\backslash i, [p]\backslash ,i}$, then combine \eqref{eq:prop1 pf2} and \eqref{eq:prop1 pf3}, we have that
\begin{equation*}
    \begin{aligned}
        Cov(\bmY^{t,do(i)}_{[p]\backslash i}) &= \left(\bI_p +
            \sigma_k^{-1}\Delta \left(\sum_{l\in anc(j)\cup j}[\bB^{do(i)}]_{j,l}[\bSigma^{do(i)}]^{1/2}_{l,l}\bE^{k,l}  \right) + \right.\\
            &\quad \quad \quad \sigma_k^{-1}\Delta \left(\sum_{l\in anc(j)\cup j}[\bB^{do(i)}]_{j,l}[\bSigma^{do(i)}]^{1/2}_{l,l}\bE^{l,k}\right)+
            \\
             &\left.\quad \quad \quad \sigma_k^{-2}\Delta^2\left(\sum_{l\in anc(j)\cup j}([\bB^{do(i)}]_{j,l})^2[\bSigma^{do(i)}]_{l,l}\right)\bE^{k,k}\right) _{[p]\backslash i, [p]\backslash i}
    \end{aligned}
\end{equation*}

Now we have completed the discussion of all exceptional cases in Proposition \ref{prop:Yt}; thus, the proof is complete.
\end{proof}

\section{Proof of Proposition \ref{prop: Ia}}
\label{app:prop2}
In this section, we prove Proposition \ref{prop: Ia} in Section \ref{sec:intervention sub3}. We use the same notation as Appendix \ref{app: prop 1}.

\begin{proof}
We consider the change magnitude is $\Delta$ and the change location is $[k,j]$, i.e., node $j$ is the origin of the changed edge.

We begin with the KL divergence for the multivariate normal distribution 
\begin{equation}
\label{eq: multi norm KL}
    D(N(\bmu,\bSigma)||N(0,\bI_p)) = \frac{1}{2}\left( Tr(\bSigma)-p + \bmu^T\bmu - \log|\bSigma |\right),
\end{equation}
where $Tr(\cdot)$ denote the trace of matrix.

To prove \eqref{eq:prop2 1}, we only need to show 
\begin{equation*}
\begin{aligned}
    Tr(Cov(\bmY^{t,do(j)}_{[p]\backslash j})) &+ ||\bbE[\bmY^{t,do(j)}_{[p]\backslash j}]||_2^2 - \log |Cov(\bmY^{t,do(j)}_{[p]\backslash j})|  \\ &> Tr(Cov(\bmY^{t,do(i)}_{[p]\backslash i})) + ||\bbE[\bmY^{t,do(i)}_{[p]\backslash i}]||_2^2 - \log |Cov(\bmY^{t,do(i)}_{[p]\backslash i})| + 2\delta
\end{aligned}
\end{equation*}
holds for $\forall i \in 0\cup[p] \backslash j$. Through algebraic calculations, we find that $\log |Cov(\bmY^{t,do(i)}_{[p]\backslash i})| = 0\quad \forall i \in 0\cup[p]$. Then combining Proposition \ref{prop:Yt}, we have that the above equation is equivalent to 
\begin{equation*}
\begin{aligned}
    &\sigma_k^{-2}\Delta^2 \left( \sum_{m\in anc(j)\cup j} [\bB^{do(j)}]_{j,m} \mu^{do(j)}_m  \right)^2 +   \sigma_k^{-2}\Delta^2\left(\sum_{m\in anc(j)\cup j}([\bB^{do(j)}]_{j,m})^2[\bSigma^{do(j)}]_{m,m}\right) \\
    &> \sigma_k^{-2}\Delta^2 \left( \sum_{m\in anc(j)\cup j} [\bB^{do(i)}]_{j,m} \mu^{do(i)}_m  \right)^2 +   \sigma_k^{-2}\Delta^2\left(\sum_{m\in anc(j)\cup j}([\bB^{do(i)}]_{j,m})^2[\bSigma^{do(i)}]_{m,m}\right) + 2\delta.
\end{aligned}
\end{equation*}
Then by \eqref{eq:B anc}, the above expression can be further simplified to
\begin{equation}
\label{eq:prop2 pf1}
    c_j > \sqrt{\left(\sum_{m \in anc(j)}[\bB^{do(i)}]_{j,m}\mu_m^{do(i)}\right)^2 + \sum_{m \in anc(j)}[\bSigma^{do(i)}]_{m,m}\left([\bB^{do(i)}]_{j,m} \right)^2 + \frac{2\delta \sigma_k^2}{\Delta^2}}.
\end{equation} 

\eqref{eq:prop2 pf1} is already very close to the one in Algorithm \ref{alg: c_j}. Since the value of $\Delta$ is unknown in the actual computation process, we take $\Delta_{min}$ to ensure that \eqref{eq:prop2 pf1} holds. Moreover, although we do not know which node $k$ is pointed to by the changed edge, we know that when the origin node is $j$, based on Assumption \ref{ass:single change}, no additional cycles are introduced, and $k$ will not belong to the set $anc(j)\cup j$. Therefore, we can replace $\sigma_k^2$ with $\bSigma^{do(X_j = c_j)}_{max}$. 
Finally we only need to take the maximum value for $i \in anc(j)\cup 0$. 

Thus, we have proven that when the origin node of the changed edge is node $j$, the value $c_j$ obtained through Algorithm \ref{alg: c_j} ensures that \eqref{eq:prop2 1} holds. Furthermore, by iteratively solving for each $c_l$ (note that solving for subsequent $c_l$ does not affect the validity of the preceding properties), we can guarantee that \eqref{eq:prop2 1} holds for any change location.

Then we prove Equations (\ref{eq:prop2 2})-(\ref{eq:prop2 4}).

Using Proposition \ref{prop:Yt} and \eqref{eq: multi norm KL}, we have 
\begin{equation}
\label{eq:prop2 pf2}
    I_{i,c_i}^{\Delta,[k,j]}[k] =\begin{cases}
         I_{i,c_i}^{\Delta,[k,j]} - \log( [Cov(\bmY^{t,do(i)})]_{k,k}) &\text{ for } i\neq k,\\
         0 &\text{ for } i =  k.
    \end{cases}
\end{equation}
Based on Proposition \ref{prop:Yt}, we also have that 
\begin{equation}
\label{eq:prop2 pf3}
    [Cov(\bmY^{t,do(i)})]_{k,k}  \begin{cases}
         = 1 &\text{ for } i = j, \\
         = 0 &\text{ for } i = k,\\
         \geq 1 &\text{ otherwise}.
    \end{cases}
\end{equation}

Then combine \eqref{eq:prop2 pf2}, \eqref{eq:prop2 pf3} and \eqref{eq:prop2 1}, we can conclude that \eqref{eq:prop2 2} and \eqref{eq:prop2 4} hold. 

As for \eqref{eq:prop2 3}, it is easy to check that when $t \geq \tau$, $Y^{t,do(i)}_l \sim N(0,1)$ for $i\in0\cup[p],l\in [p]\backslash (k\cup i)$, and $Y_i^{t,do(i)}\sim 1_{0}$ if $i\neq 0$ which remains consistent with the case when $t<\tau$, then \eqref{eq:prop2 3} holds.
\end{proof}

\section{Additional Notation in Appendices \ref{app: lower bound}-\ref{app:max}}
\label{app:notation}
\textbf{To facilitate the presentation, we use the following notations in the subsequent Appendices \ref{app: lower bound}-\ref{app:max}.}

When $t < \tau$, we denote $\bmu_{Y,[p]\backslash i}^{0,do(i)} \triangleq \bbE(\bmY^{t,do(X_i^t = c_i)}_{[p]\backslash i} ) = \mathbf{0}$, $[\bSigma_Y^{0,do(i)}]_{[p]\backslash i,[p]\backslash i} \triangleq Cov(Y^{t,do(X_i^t = c_i)}_{[p]\backslash i} ) =  \bI_{|[p]\backslash i|}$, $\mu_{Y,j}^{0,do(i)} \triangleq \bbE(Y^{t,do(X_i^t = c_i)}_{j}) = 0$ and $\sigma_{Y,j}^{2,0,do(i)} \triangleq Var(Y^{t,do(X_i^t = c_i)}_{j} ) = 1$.  When $t \geq \tau$, we denote  $\bmu_{Y,[p]\backslash i}^{\Delta,[k,j],do(i)} \triangleq \bbE(\bmY^{t,do(X_i^t = c_i)}_{[p]\backslash i} )$, $[\bSigma_Y^{\Delta,[k,j],do(i)}]_{[p]\backslash i,[p]\backslash i} \triangleq Cov(\bmY^{t,do(X_i^t = c_i)}_{[p]\backslash i} )$, $\mu_{Y,j}^{\Delta,[k,j],do(i)} \triangleq \bbE(Y^{t,do(X_i^t = c_i)}_{j})$ and $\sigma_{Y,j}^{2,\Delta,[k,j],do(i)} \triangleq Var(Y^{t,do(X_i^t = c_i)}_{j} )$ and they are totally decided by the changed edge $[k,j]$ and the change magnitude $\Delta$ according to Proposition \ref{prop:Yt}. Then define $\theta^0 = \{(\bmu_{Y,[p]\backslash i}^{0,do(i)}, [\bSigma_Y^{0,do(i)}]_{[p]\backslash i,[p]\backslash i})\}_{i=0}^p$ and $\theta = \theta^{\Delta,[k,j]} = \{(\bmu_{Y,[p]\backslash i}^{\Delta,[k,j],do(i)},[\bSigma_Y^{\Delta,[k,j],do(i)}]_{[p]\backslash i,[p]\backslash i}) \}_{i=0}^p$. They represent the mean and variance of the multivariate normal distribution that $\bmY^t_{[p]\backslash i}$ follows, given the intervention on node $i$ under the pre-change and post-change cases for $i \in 0\cup[p]$. 
If we use $f_{l}^{\theta}$ to denote the distribution function of a normal distribution with parameters as the $l$th pair in $\theta$ and use $f_{l}^{\theta}[k]$ denotes  distribution function of its $k$th dimension, we can find that
 $f^{\theta}_{a_t} = f^{\Delta,[k,j]}_{a_t,c_{a_t}}$, $f^{\theta^0}_{a_t} = f^{0,[k,j]}_{a_t,c_{a_t}}$, $f^{\theta}_{a_t}[l] = f^{\Delta,[k,j]}_{a_t,c_{a_t}}[l]$ and $f^{\theta^0}_{a_t}[l] = f^{0,[k,j]}_{a_t,c_{a_t}}[l]$ (The subscript $c_{a_t}$ is removed because, as mentioned earlier, we assume that $\bmC$ is obtained through Algorithm \ref{alg: c_j}).  This means that we can regard $\theta$ as our post-change distribution parameter, which we do not know, and regard $\theta^0$ as our pre-change distribution parameter. 
 We define the post-change parameter space as $\Theta \triangleq \{\theta^{\Delta,[k,j]}| \Delta,[k,j] \text{ satisfies Assumption \ref{ass:single change}} \}$. In this way, we transform the original problem into a change point detection problem with an unknown post-change parameter $\theta$. 
 
 Define $\Tilde{\Theta} \triangleq \{\Tilde{\theta} = \{(\Tilde{\bmu}^i, \Tilde{\bSigma}^i)\}_{i=0}^p | \Tilde{\bmu}^i \in \bbR^{|[p]\backslash i|}, ||\Tilde{\bmu}^i ||_2^2 <M, \Tilde{\bSigma}^i\in\bbR^{|[p]\backslash i| \times|[p]\backslash i| } \text{ is non-negative definite},\\ 1/M<|\Tilde{\bSigma}^i|<M\} \backslash \theta^0$ and $M$ is a bounded constant. According to Assumption \ref{ass:single change}, we have $|\Delta| < \Delta_{max}$ which means that we can choose $M$ large enough to make $\Theta \subset \Tilde{\Theta}$. In this way, $\Tilde{\Theta}$ can be viewed as an expansion of $\Theta$. This approach simplifies calculations by allowing us to choose large enough constant $M$ that ensure the estimate $\hat{\theta}_t^{\circ}$ and $\hat{\theta}_t^{*}$ is well-represented.

 In this way, we can rewrite \eqref{eq:esttheta multi} as 
 \begin{equation*}
    \hat{\theta}_t^{\circ} = \arg \max_{\Tilde{\theta}\in\Tilde{\Theta}} \sum_{m=t-w}^{t-1}\log f_{a_t^{\circ}}^{\Tilde{\theta}}(\bmY^{t,do(a_t^{\circ})}).
\end{equation*} 
where $\hat{\theta}_t^{\circ} = \{(\hat{\bmu}_Y^{t,do(i),\circ}, \hat{\bSigma}_Y^{t,do(i),\circ} )\}_{i\in0\cup[p]}$ is the estimated parameters in \namea. Rewrite \eqref{eq:esttheta max} as 
\begin{equation*}
     \hat{\theta}_t^{*} = \arg \max_{\Tilde{\theta}\in\Tilde{\Theta}} \sum_{m=t-w}^{t-1}\log f_{a_t^{*}}^{\Tilde{\theta}}(\bmY^{t,do(a_t^{*})}).
 \end{equation*}
 where $\hat{\theta_t^{*}} = \{(\hat{\bmu}_Y^{t,do(i),*}, \hat{\bSigma}_Y^{t,do(i),*} )\}_{i\in0\cup[p]}$ and in Algorithm \ref{alg:WLCUSUM}, we do not require the covariance estimates. Instead, $\{\hat{\mu}_{Y,j}^{t,do(i),*}, \hat{\sigma^2}_{Y,j}^{t,do(i),*}  = [\hat{\bSigma}_Y^{t,do(i),*} ]_{j,j}\}_{i\in0\cup[p],j\in [p]\backslash i}$ represents our estimated parameter in \nameb.

 Then we simplify $\hat{f}_{a}^{t,\Delta,[k,j],\circ}, \hat{I}_a^{t,\Delta,[k,j],\circ},\Lambda_{t,A^{\circ}}^{\Delta,[k,j],\circ}$ as $f_a^{\hat{\theta}_t^{\circ}}, I_{a}^{\hat{\theta}_t^{\circ}}, \Lambda_{t,A^{\circ}}^{\hat{\theta}_t^{\circ}}$ respectively and  simplify $\hat{f}_{a}^{t,\Delta,[k,j],*}[l], \hat{I}_a^{t,\Delta,[k,j],*}[l],\Lambda_{t,A^{\circ}}^{\Delta,[k,j],*}[l]$ as $f_a^{\hat{\theta}_t^{*}}[l], I_{a}^{\hat{\theta}_t^{*}}[l], \Lambda_{t,A^{\circ}}^{\hat{\theta}_t^{*}}[l]$ respectively. Further more, we simplify $\bbE_{\tau,A}^{\Delta,[k,j]}[\cdot]$ as $\bbE_{\tau,A}^{\theta}[\cdot]$ and simplify $\cJ_{\Delta,[k,j]}(A,T)$ as $\cJ_{\theta}(A,T)$
.

 \section{Proof of Theorem \ref{thm:general}}
\label{app: lower bound}
\textbf{In this section, we use the additional notations defined in Appendix \ref{app:notation}}
, and we prove Theorem \ref{thm:general} in Section \ref{sec: thm sub1}.

Before starting the proof, we first present a conclusion that will be used in all the subsequent proofs. For any $\theta\in \Tilde{\Theta}$ and any $j\in 0\cup[p]$, we have 
\begin{equation}
    \label{eq:bounded V}
    \int \log(\frac{f_j^{\theta}(y)}{f_j^{\theta^0}(y)}) ^2 f_{j}^{\theta}(y) dy < V < \infty.
\end{equation}
where $\theta^0$ is the pre-change parameter, as we defined earlier. This is easy to check since $\Tilde{\Theta}$ is a bounded set, and the $|\Tilde{\bSigma}^i|$ in it has a lower bound from zero. 

\begin{proof}
Suppose $(A,T) \in \cC_{\gamma}$. For the post-change $\{\Delta, [k,j]\}$ with $\theta \in \Theta$ as its post-change parameter, according to Theorem 1 in \citet{lai1998information}, we only need to show that, for every $\xi > 0$, the sequence
\begin{equation}
\label{eq:pf thm1}
    \sup_{\tau \in \bbN} \text{ess}\sup \bbP_{\tau,A}^{\theta}\left(\max_{1\leq t\leq n}\sum_{m=\tau}^{\tau+t} \Lambda_{m,A}^{\theta} > I^{\Delta,[k,j]}_{j,c_j}(1+\xi) n | \cF_{\tau -1}\right)
\end{equation}
converges to 0 as $n\to \infty$. $\bbP_{\tau,A}^{\theta}(\cdot)$ denotes the probability measure with sequential intervention policy $A$ under the post-change condition whose change time is $\tau$ and post-change parameter is $\theta$. $\Lambda_{m,A}^{\theta} = \log(\frac{f^{\theta}_{a_m}(Y^{m,do(a_m)})}{f^{\theta^0}_{a_m}(Y^{m,do(a_m)})})$ has the similar definition in Section \ref{sec: change point}. 

Then for every $n, \tau \in \bbN$, we have
\begin{equation*}
    \begin{aligned}
        \bbP_{\tau,A}^{\theta}&\left(\max_{1\leq t\leq n}\sum_{m=\tau}^{\tau+t} \Lambda_{m,A}^{\theta} > I^{\Delta,[k,j]}_{j,c_j}(1+\xi) n | \cF_{\tau -1}\right) \\
        &\leq\bbP_{\tau,A}^{\theta}\left(\max_{1\leq t\leq n}\sum_{m=\tau}^{\tau+t} \left( \Lambda_{m,A}^{\theta}-I^{\Delta,[k,j]}_{a_m,c_{a_m}} \right) > I^{\Delta,[k,j]}_{j,c_j}\xi n | \cF_{\tau -1}\right) \\
        & \leq \frac{nV}{(I_{j,c_j}^{\Delta,[k,j]}\xi n)^2} = \frac{V}{(I_{j,c_j}^{\Delta,[k,j]}\xi)^2 n}.
    \end{aligned}
\end{equation*}
The first inequality follows from \eqref{eq:prop2 1}, the second inequality follows from a conditional version of Doob's submartingale inequality. We can apply the latter because 
\begin{equation*}
     \left\{Y^{\theta,\tau:t}_{A}\triangleq \sum_{m=\tau}^{\tau+t} \left( \Lambda_{m,A}^{\theta}-I^{\Delta,[k,j]}_{a_m,c_{a_m}} \right), \cF_{\tau+t}, t\in\bbN \right\}
\end{equation*}
is a $\bbP_{\tau,A}^{\theta}$-martingale and according to \eqref{eq:bounded V}, it follows that 
\begin{equation}
    Var_{\tau,A}^{\theta}(Y^{\theta,\tau:t}_{A} | \cF_{\tau-1}) \leq Vt \quad \text{for all } t\in\bbN.
\end{equation}

Thus we prove \eqref{eq:pf thm1} converges to $0$ as $n\to\infty$, and complete the proof.
\end{proof}

\section{Proof of Lemmas \ref{lem:multi IC}-\ref{lem:multi liminf} and Theorem \ref{thm:multi}}
\label{app:multi}
\textbf{In this section, we use the additional notations defined in Appendix \ref{app:notation}}
and we  prove Lemmas \ref{lem:multi IC}-\ref{lem:multi liminf} and Theorem \ref{thm:multi} in Section \ref{sec: thm sub2}.

\subsection{Proof of Lemma \ref{lem:multi IC}}
\label{app:multi IC}
\begin{proof}
We can rewrite the test statics in \eqref{eq:W multi} as follows,
\begin{equation*}
    e^{W_{t,A^{\circ}}} = \max\{e^{W_{t-1,A^{\circ}}} ,1\} \frac{f^{\hat{\theta}_t^\circ}_{a_t^\circ}(\bmY^{t,do(a_t^{\circ})})}{f^{\theta^0}_{a_t^\circ}(\bmY^{t,do(a_t^\circ)})}, \ t > w, \ e^{W_{w,A^{\circ}}} = 1.
\end{equation*}

Then we can define a new statistic as follows,
\begin{equation*}
    \cW_{t,A^{\circ}} = (\cW_{t-1,A^{\circ}} + 1) \frac{f^{\hat{\theta}_t^\circ}_{a_t^\circ}(\bmY^{t,do(a_t^{\circ})})}{f^{\theta^0}_{a_t^\circ}(\bmY^{t,do(a_t^\circ)})}, \ t > w,\ \cW_{w,A^{\circ}} = 1.
\end{equation*}
It is easy to find that $\cW_{t,A^{\circ}} \geq e^{W_{t,A^{\circ}}}$ for $\forall \ t \geq w $.

Because $\hat{\theta}_t^{\circ}$ is $\cF_{t-1}$-measurable, we have
\begin{equation*}
    \bbE_{\infty,A^{\circ}}[\cW_{t,A^{\circ}} - t| \cF_{t-1}] = \cW_{t-1,A^{\circ}} + 1-t = \cW_{t-1,A^{\circ}} - (t-1).
\end{equation*}
This means that $\{\cW_{t,A^{\circ}} - t| t \geq w\}$ is a martingale under $\bbP_{\infty,A^{\circ}}$, with respect to the filtration $\{\cF_t |t\in\bbN \}$. $\bbP_{\infty,A}(\cdot)$ denote the probability measure under pre-change condition with sequential intervention policy $A$.

Then using the Optional Sampling Theorem \citep{tartakovsky2014sequential}, we have 
\begin{equation*}
    \bbE_{\infty,A^{\circ}}[\cW_{T_{b,A^{\circ}}^{multi}, A^{\circ}} -T_{b,A^{\circ}}^{multi} ] = \cW_{w,A^{\circ}} - w = 1-w,
\end{equation*}
which implies that 
\begin{equation*}
    \bbE_{\infty,A^{\circ}}[ T_{b,A^{\circ}}^{multi}] = \bbE_{\infty,A^{\circ}}[\cW_{T_{b,A^{\circ}}^{multi}, A^{\circ}}] + w -1 \geq \bbE_{\infty,A^{\circ}}[e^{W_{T_{b,A^{\circ}}^{multi},A^{\circ}}}] \geq e^b.
\end{equation*}
We can set $b = \log\gamma$ to ensure $(A^{\circ},T_{b,A^{\circ}}^{multi} ) \in \cC_{\gamma}$.

\end{proof}

\subsection{Proof of Lemma \ref{lem:multi OC}}
\label{app:multi OC}
\begin{proof}
For each $\tau \in \bbN$, we can define a new stopping time $T_{b,A^{\circ},\tau}^{multi}$ as follows
\begin{equation*}
    T_{b,A^{\circ},\tau}^{multi} = \inf\{t>\tau +w-1 | W_{t,A^{\circ}}^{\tau} > b \}, 
\end{equation*}
where 
\begin{equation*}
    W_{t,A^{\circ}}^{\tau} = \max\{W_{t-1,A^\circ}^{\tau} ,0\} + \Lambda_{t,A^\circ}^{\hat{\theta}_t^\circ}, \ t>\tau+w-1, \quad W_{\tau+w-1,A^\circ}^{\tau} = 0.
\end{equation*}

Define $x^+ = \max\{x,0\}$, for any $\tau \in \bbN$ and $\theta \in \Theta$, we have
\begin{equation*}
    \begin{aligned}
        \bbE_{\tau,A^{\circ}}^{\theta}\left[(T_{b,A^{\circ}}^{multi} - \tau + 1)^+ | \cF_{\tau-1}\right] &\leq w + \bbE_{\tau,A^{\circ}}^{\theta}\left[(T_{b,A^{\circ}}^{multi} - \tau + 1 + w)^+ | \cF_{\tau-1}\right] \\
        & \leq w +   \bbE_{\tau,A^{\circ}}^{\theta}\left[T_{b,A^{\circ},\tau}^{multi} - \tau + 1 + w | \cF_{\tau-1}\right] \\
        & = \bbE_{\tau,A^{\circ}}^{\theta}\left[ \bbE_{\tau,A^{\circ}}^{\theta}\left[T_{b,A^{\circ},\tau}^{multi} - \tau + 1  |A^{\circ}_{[\tau,\tau+w-1]}, \cF_{\tau-1}\right]  | \cF_{\tau-1}\right] \\
        &= \bbE_{\tau,A^{\circ}}^{\theta}\left[ \bbE_{\tau,A^{\circ}}^{\theta}\left[T_{b,A^{\circ},\tau}^{multi} - \tau + 1  |A^{\circ}_{[\tau,\tau+w-1]}\right]  | \cF_{\tau-1}\right] \\
        &\leq \sup_{a_{\tau}^{\circ},\dots,a_{\tau+w-1}^{\circ}} \bbE_{\tau,A^{\circ}}^{\theta}\left[T_{b,A^{\circ},\tau}^{multi} - \tau + 1  |A^{\circ}_{[\tau,\tau+w-1]}  \right]\\
        & = \sup_{a_1^{\circ},\dots,a_w^{\circ}} \bbE_{1,A^{\circ}}^{\theta}\left[T_{b,A^{\circ}}^{multi}  |A^{\circ}_{[w]} \right],
    \end{aligned}
\end{equation*}
where $A^{\circ}_{[c,d]} \triangleq \{a_t^{\circ}| c\leq t\leq d \}$. The first inequality holds because $x^+ < y+(x-y)^+$ for $x,y>0$. The second inequality holds because $T_{b,A^{\circ}}^{multi} \leq T_{b,A^{\circ},\tau}^{multi}$ since $W_{t,A^{\circ}} \geq W_{t,A^{\circ}}^\tau$. The first equality follows from the law of iterated expectation. The  second holds because $T_{b,A^{\circ},\tau}^{multi}$ only dependents on $A^{\circ}_{[\tau,\tau+w-1]}$. The last equality follows from the definition of $T_{b,A^{\circ},\tau}^{multi}$.

\end{proof}

\subsection{Proof of Lemma \ref{lem:multi arm j}}
\label{app:multi arm j}
\begin{proof}
We first recall the notation. Fix the change magnitude $\Delta$ and change location $[k,j]$, $\theta = \theta^{\Delta,[k,j]} = \{(\bmu_{Y,[p]\backslash i}^{\Delta,[k,j],do(i)},[\bSigma_Y^{\Delta,[k,j],do(i)}]_{[p]\backslash i,[p]\backslash i}) \}_{i=0}^p \in \Theta$ is the post-change parameter. $\bmu_{Y,[p]\backslash i}^{\Delta,[k,j],do(i)} \in \bbR^{|[p]\backslash i|}$ is the mean of $\bmY^{t,do(X_i^t=c_i)}_{[p]\backslash i}$ for $t \geq \tau$ and for convenience, we use $\bmu_Y^{\theta,do(i)}$ as its abbreviation. $[\bSigma_Y^{\Delta,[k,j],do(i)}]_{[p]\backslash i,[p]\backslash i} \in \bbR^{|[p]\backslash i| \times |[p]\backslash i|}$ is the covariance matrix of $\bmY^{t,do(X_i=c_i)}_{[p]\backslash i}$ for $t \geq \tau$, and for convenience, we use $\bSigma_Y^{\theta,do(i)}$ as its abbreviation. Since the dimensions of $\bSigma_Y^{\theta,do(i)}$ and $\bSigma_Y^{\Delta,[k,j],do(i)}$ may differ when \( i \neq 0 \), we by default index $\bSigma_Y^{\theta,do(i)}$  based on its corresponding indices in $\bSigma_Y^{\Delta,[k,j],do(i)}$ in subsequent steps. The same applies to $\bmu_Y^{\theta,do(i)}$ and $\bmu_Y^{\Delta,[k,j],do(i)}$ as well.  $\Tilde{\theta} = \{(\Tilde{\bmu}^i,\Tilde{\bSigma}^i )_{i=0}^p \} \in \Tilde{\Theta}$ is a parameter in the extend post-change parameter space. In the computation, we use MLE to find an optimal $\Tilde{\theta}\in\Tilde{\Theta}$ for estimating $\theta$, where $\Tilde{\bmu}^i$ is the estimate of $\bmu_Y^{\theta,do(i)}$, and $\Tilde{\bSigma}^i$ is the estimate of $\bSigma_Y^{\theta,do(i)}$. Similarly, we index their dimensions using the same way as $\bmu_Y^{\theta,do(i)}$ and $\bSigma_Y^{\theta,do(i)}$.

In this section, we consider $\bmC$ is the outcome of Algorithm \ref{alg: c_j}. Additionally, for $\Tilde{\theta} \in \Tilde{\Theta}$, we set $I_{i}^{\Tilde{\theta}} \triangleq D(f_{i}^{\Tilde{\theta}}||f_{i}^{\theta^0} )$, $i\in 0\cup[p]$.  Then we can use $I_i^{\theta}$ as the abbreviation of $I_{i,c_i}^{\Delta, [k,j]}$ and $I^{\theta} = \max_{i\in0\cup[p]} I_i^{\theta}$. According to Proposition \ref{prop: Ia}, we have $I^{\theta} = I_{j,c_j}^{\Delta,[k,j]}$ when change occurs at location $[k,j]$.

Then for any $\theta \in \Theta, \zeta \in \Tilde{\Theta}$ and $a\in 0\cup[p]$, we set 
\begin{equation}
    B_i(\theta,\zeta)\triangleq \int \log\left(\frac{f_i^{\zeta}(x)}{f_i^{\theta^0}(x)} \right)f_i^{\theta}(x)dx = D(f_a^{\theta}||f_a^{\theta^0}) - D(f_i^{\theta}||f_i^{\zeta}) = I_i^{\theta} - D(f_i^{\theta}||f_i^{\zeta}).
\end{equation}
It is clear that $B_i(\theta, \theta) = I_i^{\theta}$ and $I_i^\theta \geq B_i(\theta,\zeta)$.
We also set $a^\circ(\zeta)$ is the selected intervention node according to \eqref{eq:inter multi} by replacing $\hat{f}_i^{t,\Delta,[k,j],\circ} = f_i^{\hat{\theta}_t^{\circ}}$ by $f_i^{\zeta}$ for $t \notin \sN$.

For a fixed $q\in \bbN$, we define $\epsilon = \epsilon^q = \min\{\delta/(2p+1), q^{-\eta/4} \}$. Next we define $\cM_{\epsilon}^{\theta}$ is the $\epsilon$-neighbor set of $\theta$ which consist of $\Tilde{\theta}= \{(\Tilde{\bmu}^i,\Tilde{\bSigma}^i )_{i=0}^p \} $  that satisfies the following conditions,
\begin{align}
\label{eq:lem3 pf1}
    |\Tilde{\mu}^i_j - \mu_{Y,j}^{\theta,do(i)}| < \epsilon&\text{ for }i \in 0\cup[p]\text{ and }j \in [p]\backslash i. \\
    |(\Tilde{\mu}^i_j)^2 - (\mu_{Y,j}^{\theta,do(i)})^2| < \epsilon&\text{ for }i \in 0\cup[p]\text{ and }j \in [p]\backslash i.\\
    |[\Tilde{\bSigma}^i]_{j,j}-[\bSigma_{Y}^{\theta,do(i)}]_{j,j}|<\epsilon &\text{ for }i \in 0\cup[p]\text{ and }j \in [p]\backslash i.\\
    |\log|\Tilde{\bSigma}^i | - \log|\bSigma_{Y}^{\theta,do(i)}| |< \epsilon&\text{ for }i \in 0\cup[p]. \\
    \label{eq:lem3 pf5}|Tr((\Tilde{\bSigma}^i)^{-1}\bSigma_{Y}^{\theta,do(i)}) - |[p]\backslash i| | < \epsilon&\text{ for }i \in 0\cup[p].
\end{align}
Consider the bounded constant $M$ in $\Tilde{\Theta}$ is large enough, we can say $\cM_\epsilon^\theta \subset \Tilde{\Theta}$. Obviously, $\cM_{\epsilon}^{\theta}$ is not empty since at least $\theta \in \cM_{\epsilon}^{\theta}$.

Then we provide two important properties of $\cM_q^\theta$.

This first is, if $\zeta \in \cM_{\epsilon}^\theta$,  we can find that,
\begin{equation}
\label{eq:neighbor multi 1}
     |I_i^{\theta} - I_i^{\zeta}| < \delta/2, 
\end{equation}
hols for any $i \in 0\cup[p]$.
Combine this with \eqref{eq:prop2 1}, we can further get $a^{\circ}(\zeta) = j = \arg \max_{i \in 0\cup[p]} I_{i}^{\theta}$ when the change location is $[k,j]$. 

The second property is, for any $i\in0\cup[p]$
\begin{equation}
\label{eq:neighbor multi 2}
     \lim_{\epsilon \to 0}\max_{\zeta\in\cM_\epsilon^{\theta}} D(f_i^\theta||f_i^\zeta) \to 0, 
\end{equation}
which can be directly derived from the definition of the KL divergence between two normal distributions.

By \eqref{eq:neighbor multi 1}, to prove Lemma \ref{lem:multi arm j}, we only need to prove 
\begin{equation}
\label{eq:lim P}
    \lim_{q,w\to \infty} \bbP_{1,A^\circ}^\theta \left(\hat{\theta}_t^{\circ} \in \cM_{\epsilon}^\theta| \cF_{t-w-1}, A^{\circ}_{[1:w]} \right) \to 1.
\end{equation}

Since $\cM_\epsilon^\theta \in \Tilde{\Theta}$, according to the MLE property of normal distribution, we can write the $i$th pare $(\hat{\bmu}_{Y}^{t,do(i),\circ}, \hat{\bSigma}_{Y}^{t,do(i),\circ})$ in $\hat{\theta}_t^{\circ}$ as,
\begin{equation}
    \begin{aligned}
        \hat{\bmu}_{Y}^{t,do(i),\circ} &= \frac{\sum_{n=t-w}^{t-1} \bmY^{n,do(a_n^\circ)}_{[p]\backslash a_n^\circ}\bbI\{a_n^\circ = i\}}{\sum_{n=t-w}^{t-1} \bbI\{a_n^{\circ} = i\}}, \\
        \hat{\bSigma}_{Y}^{t,do(i),\circ} &= \frac{\sum_{n=t-w}^{t-1}(\bmY^{n,do(a_n^\circ)}_{[p]\backslash a_n^\circ} - \hat{\bmu}_{Y}^{t,do(i),\circ}) (\bmY^{n,do(a_n^\circ)}_{[p]\backslash a_n^\circ} - \hat{\bmu}_{Y}^{t,do(i),\circ})^T  \bbI\{a_n^\circ = i\}}{\sum_{n=t-w}^{t-1} \bbI\{a_n^{\circ} = i\}},
    \end{aligned}
\end{equation}
 For convenient, we denote $n_i = \sum_{n=t-w}^{t-1} \bbI\{a_n^{\circ} = i\}$ for $i\in0\cup[p]$. Since when $t\in\sN$, we uniformly random select intervention node, we have $n_i = \Omega(q^{\eta})$ as $q\to \infty$.

Combine \eqref{eq:lem3 pf1} to (\ref{eq:lem3 pf5}), we have that 
\begin{equation}
\begin{aligned}
     \bbP_{1,A^\circ}^\theta \left(\hat{\theta}_t^{\circ} \in \cM_{\epsilon}^\theta| \cF_{t-w-1}, A^{\circ}_{[1:w]} \right) \geq 1 &- \sum_{i=0}^p\sum_{j=1,j\neq i}^p \underbrace{\bbP_{1,A^\circ}^\theta \left( |\hat{\mu}_{Y,j}^{t,do(i),\circ} - \mu_{Y,j}^{\theta,do(i)}| \geq \epsilon\right)}_{I} \\
     & - \sum_{i=0}^p\sum_{j=1,j\neq i}^p\underbrace{\bbP_{1,A^\circ}^\theta \left( |(\hat{\mu}_{Y,j}^{t,do(i),\circ})^2 - (\mu_{Y,j}^{\theta,do(i)})^2| \geq \epsilon\right)}_{II} \\ 
     & - \sum_{i=0}^p\sum_{j=1,j\neq i}^p\underbrace{\bbP_{1,A^\circ}^\theta \left( |[\hat{\bSigma}_{Y}^{t,do(i),\circ}]_{j,j}-[\bSigma_{Y}^{\theta,do(i)}]_{j,j}|\geq\epsilon\right)}_{III} \\ 
     & -\sum_{i=0}^p\underbrace{\bbP_{1,A^\circ}^\theta \left(|\log|\hat{\bSigma}_{Y}^{t,do(i),\circ} | - \log|\bSigma_{Y}^{\theta,do(i)}| |\geq \epsilon\right)}_{IV} \\ 
     & -\sum_{i=0}^p\underbrace{\bbP_{1,A^\circ}^\theta \left(|Tr((\hat{\bSigma}_{Y}^{t,do(i),\circ})^{-1}\bSigma_{Y}^{\theta,do(i)}) - |[p]\backslash i| | \geq \epsilon\right)}_{V}.
\end{aligned}
\end{equation}
Next, we will prove term $I$ to term $V$ converge to $0$ as $q \to \infty$ using the centralization properties of the sample mean and sample covariance matrix of the normal distribution.

Consider term $I$, using Markov inequality, we have 
\begin{equation*}
    \bbP_{1,A^\circ}^\theta \left( |\hat{\mu}_{Y,j}^{t,do(i),\circ} - \mu_{Y,j}^{\theta,do(i)}| \geq \epsilon\right) \leq \frac{1}{\epsilon^2} Var\left(\hat{\mu}_{Y,j}^{t,do(i),\circ} \right)  = \frac{[\bSigma_Y^{\theta,do(i)}]_{j,j}}{n_i \epsilon^2} = O(q^{-\eta/2})
\end{equation*}
converges to $0$ as $q\to \infty$.

Consider term $II$, using Markov inequality, we have
\begin{equation*}
\begin{aligned}
    \bbP_{1,A^\circ}^\theta \left( |(\hat{\mu}_{Y,j}^{t,do(i),\circ})^2 - (\mu_{Y,j}^{\theta,do(i)})^2| \geq \epsilon\right) &\leq \frac{1}{\epsilon^2}\left(Var\left((\hat{\mu}_{Y,j}^{t,do(i),\circ})^2\right) + \left(\bbE_{1,A^\circ}^\theta[(\hat{\mu}_{Y,j}^{t,do(i),\circ})^2] - (\mu_{Y,j}^{\theta,do(i)})^2\right)^2 \right) \\
    & =\frac{1}{\epsilon^2} \left(\frac{2[\bSigma_{Y}^{\theta,do(i)}]_{j,j}^2 + 4n_i[\bSigma_{Y}^{\theta,do(i)}]_{j,j} (\mu_{Y,j}^{\theta,do(i)})^2 }{n_i^2} +\frac{[\bSigma_Y^{\theta,do(i)}]_{j,j}^2}{n_i^2} \right)\\
    &= O(q^{-\eta/2})
\end{aligned}
\end{equation*}
converges to $0$ as $q\to \infty$.

Consider term $III$, using Markov inequality, we have
\begin{equation*}
    \begin{aligned}
        \bbP_{1,A^\circ}^\theta \left( |[\hat{\bSigma}_{Y}^{t,do(i),\circ}]_{j,j}-[\bSigma_{Y}^{\theta,do(i)}]_{j,j}|\geq\epsilon\right) & \leq \frac{1}{\epsilon^2} \left(Var\left([\hat{\bSigma}_{Y}^{t,do(i),\circ}]_{j,j} \right) + \left(\bbE_{1,A^{\circ}}^{\theta}\left[[\hat{\bSigma}_{Y}^{t,do(i),\circ}]_{j,j} \right] - [\bSigma_{Y}^{\theta,do(i)}]_{j,j} \right)^2   \right)  \\
        & = \frac{1}{\epsilon^2} \left( \frac{(2n_i-2)[\bSigma_Y^{\theta,do(i)}]_{j,j}^2}{n_i^2}+\frac{[\bSigma_Y^{\theta,do(i)}]_{j,j}^2}{n_i^2}\right) \\
        & = O(q^{-\eta/2})
    \end{aligned}
\end{equation*}
converges to $0$ as $q\to \infty$.

Consider term $IV$, using Markov inequality, we have
\begin{equation}
\label{eq:IV 1}
    \begin{aligned}
    \bbP_{1,A^\circ}^\theta &\left(|\log|\hat{\bSigma}_{Y}^{t,do(i),\circ} | - \log|\bSigma_{Y}^{\theta,do(i)}| |  
    \geq \epsilon\right)\\ 
    &\leq \frac{1}{\epsilon^2} \left(Var\left(\log|\hat{\bSigma}_{Y}^{t,do(i),\circ}| \right) + \left(\bbE_{1,A^\circ}^\theta\left[\log|\hat{\bSigma}_{Y}^{t,do(i),\circ} | \right] - \log|\bSigma_{Y}^{\theta,do(i)}| \right)^2\right).
    \end{aligned}
\end{equation}
Since $n_i\hat{\bSigma}_{Y}^{t,do(i),\circ} \sim Wishart_{|[p]\backslash i|}(n_i-1,\bSigma_Y^{\theta,do(i)})$, we have 
\begin{equation*}
\begin{aligned}
    \bbE_{1,A^\circ}^\theta\left[\log|\hat{\bSigma}_{Y}^{t,do(i),\circ} | \right] &= -|[p]\backslash i|\log(n_i) + |[p]\backslash i| \log2 + \log|\bSigma_{Y}^{\theta,do(i)}| + \sum_{k=1}^{|[p]\backslash i |}\psi(\frac{n_i-k}{2}), \\
    Var\left(\log|\hat{\bSigma}_{Y}^{t,do(i),\circ}|  \right) &= \sum_{k=1}^{|[p]\backslash i|} \psi^{'}(\frac{n_i - k}{2}),
\end{aligned}
\end{equation*}
where $\psi$ is the degamma function and $\psi(x) = \log x - \frac{1}{2x} + o(\frac{1}{x})$. $\psi^{'}$is the trigamma function and $\psi^{'}(x) = \frac{1}{x} + o(\frac{1}{x})$. Thus, we have 
\begin{equation}
\begin{aligned}
    \label{eq:IV 2}
\bbE_{1,A^\circ}^\theta\left[\log|\hat{\bSigma}_{Y}^{t,do(i),\circ} | \right] - \log|\bSigma_{Y}^{\theta,do(i)}| &= O(\frac{1}{n_i}). \\ 
Var\left(\log|\hat{\bSigma}_{Y}^{t,do(i),\circ}| \right) = O(\frac{1}{n_i}).
\end{aligned}
\end{equation}
Then combine \eqref{eq:IV 1} and \eqref{eq:IV 2}, we get 
\begin{equation*}
    \bbP_{1,A^\circ}^\theta \left(|\log|\hat{\bSigma}_{Y}^{t,do(i),\circ} | - \log|\bSigma_{Y}^{\theta,do(i)}| |\geq \epsilon\right) \leq O(q^{-\eta/2})
\end{equation*}
converges to $0$ as $q \to \infty$.

Consider term $V$, using Markov inequality, we have 
\begin{equation}
\label{eq:V 1}
    \begin{aligned}
        \bbP_{1,A^\circ}^\theta &\left(|Tr((\hat{\bSigma}_{Y}^{t,do(i),\circ})^{-1}\bSigma_{Y}^{\theta,do(i)}) - |[p]\backslash i| | \geq \epsilon\right) \\ &\leq \frac{1}{\epsilon^2}\left(Var\left(Tr((\hat{\bSigma}_{Y}^{t,do(i),\circ})^{-1}\bSigma_{Y}^{\theta,do(i)})\right) + \left(\bbE_{1,A^\circ}^\theta\left[Tr((\hat{\bSigma}_{Y}^{t,do(i),\circ})^{-1}\bSigma_{Y}^{\theta,do(i)}) \right] - |[p]\backslash i| \right)^2 \right).
    \end{aligned}
\end{equation}
Since $n_i\hat{\bSigma}_{Y}^{t,do(i),\circ} \sim Wishart_{|[p]\backslash i|}(n_i-1,\bSigma_Y^{\theta,do(i)})$, we have 
\begin{equation*}
    \begin{aligned}
        \bbE_{1,A^\circ}^\theta \left[Tr((\hat{\bSigma}_{Y}^{t,do(i),\circ})^{-1}\bSigma_{Y}^{\theta,do(i)}) \right] &= \frac{n_i  |[p]\backslash i|}{n_i -|[p]\backslash i| -2}, \\ Var\left(Tr((\hat{\bSigma}_{Y}^{t,do(i),\circ})^{-1}\bSigma_{Y}^{\theta,do(i)}) \right) & = n_i^2 \left(\frac{2|[p]\backslash i|}{(n_i-|[p]\backslash i| -2)^2(n_i-|[p]\backslash i|-4)}\right. \\ 
        &\qquad\left. + \frac{2|[p]\backslash i|(|[p]\backslash i|-1)}{(n_i - |[p]\backslash i| -1)(n_i - |[p]\backslash i| -2)^2 (n_i - |[p]\backslash i| -4)} \right).
    \end{aligned}
\end{equation*}
Thus, we have 
\begin{equation}
\label{eq:V 2}
    \begin{aligned}
        \bbE_{1,A^\circ}^\theta \left[Tr((\hat{\bSigma}_{Y}^{t,do(i),\circ} )^{-1}\bSigma_{Y}^{\theta,do(i)}) \right] - |[p]\backslash i | = O(\frac{1}{n_i}), \\
        Var\left(Tr((\hat{\bSigma}_{Y}^{t,do(i),\circ})^{-1}\bSigma_{Y}^{\theta,do(i)}) \right) = O(\frac{1}{n_i}).
    \end{aligned}
\end{equation}
Then combine \eqref{eq:V 1} and \eqref{eq:V 2}, we get 
\begin{equation*}
    \bbP_{1,A^\circ}^\theta \left(|Tr((\hat{\bSigma}_{Y}^{t,do(i),\circ})^{-1}\bSigma_{Y}^{\theta,do(i)}) - |[p]\backslash i| | \geq \epsilon\right) \leq O(q^{-\eta/2})
\end{equation*}
converges to $0$ as $q \to \infty$.

Now we have \eqref{eq:lim P}, and complete the proof.

\end{proof}

\subsection{Proof of Lemma \ref{lem:multi liminf}}
\label{app:multi liminf}
\begin{proof}
In this section, we use the same notation as the last section.

For $t \notin \sN$ and fixed $q\in\bbN$, by \eqref{eq:neighbor multi 1} and \eqref{eq:neighbor multi 2}, we have
\begin{equation}
\label{eq:lem3 mainpf}
    \begin{aligned}
        &\quad \bbE_{1,A^{\circ}}^{\theta}[\Lambda_{t,A^{\circ}}^{\hat{\theta}_{t}^{\circ}}| \cF_{t-w-1}, A^{\circ}_{[w]}] \\
        & = \int_{\Tilde{\Theta}} \bbE_{1,A^\circ}^\theta\left[\log\left(\frac{f^{\zeta}_{a^\circ(\zeta)}(\bmY^{t,do(a^\circ(\zeta))}_{[p]\backslash a^\circ(\zeta)})}{f^{\theta^0}_{a^\circ(\zeta)}(\bmY^{t,do(a^\circ(\zeta))}_{[p]\backslash a^\circ(\zeta)})} \right) |\hat{\theta}_t^{\circ} = \zeta, \cF_{t-w-1}, A^{\circ}_{[1:w]}\right]\bbP_{1,A^\circ}^\theta \left(\hat{\theta}_t^{\circ} = \zeta| \cF_{t-w-1}, A^{\circ}_{[1:w]} \right) d\zeta\\
        &= \int_{\Tilde{\Theta}} B_{a^\circ(\zeta)}(\theta,\zeta) \bbP_{1,A^\circ}^\theta \left(\hat{\theta}_t^{\circ} = \zeta| \cF_{t-w-1}, A^{\circ}_{[1:w]} \right) d\zeta\\
        &=  \int_{\cM_\epsilon^{\theta}} \left(I^{\theta} - D(f^{\theta}_{j} ||f^\zeta_{j}) \right) \bbP_{1,A^\circ}^\theta \left(\hat{\theta}_t^{\circ} = \zeta| \cF_{t-w-1}, A^{\circ}_{[1:w]} \right) d\zeta \\
        &\quad + \int_{\Tilde{\Theta}\backslash\cM_\epsilon^{\theta}} 
        B_{a^\circ(\zeta)}(\theta,\zeta)
        \bbP_{1,A^\circ}^\theta \left(\hat{\theta}_t^{\circ} = \zeta| \cF_{t-w-1}, A^{\circ}_{[1:w]} \right) d\zeta \\
        &\geq \left(I^{\theta} - \max_{\zeta \in \cM_\epsilon^\theta} D(f^{\theta}_{j} ||f^\zeta_{j}) \right) \int_{\cM_\epsilon^{\theta}}  \bbP_{1,A^\circ}^\theta \left(\hat{\theta}_t^{\circ} = \zeta| \cF_{t-w-1}, A^{\circ}_{[1:w]} \right) d\zeta \\ 
        &\quad + \min_{\zeta\in\Tilde{\Theta}\backslash\cM_{\epsilon}^{\theta}}B_{a^\circ(\zeta)}(\theta,\zeta)\int_{\Tilde{\Theta}\backslash\cM_\epsilon^{\theta}} 
        \bbP_{1,A^\circ}^\theta \left(\hat{\theta}_t^{\circ} = \zeta| \cF_{t-w-1}, A^{\circ}_{[1:w]} \right) d\zeta \\
        &= I^\theta - \max_{\zeta \in \cM_\epsilon^\theta} D(f^{\theta}_{j} ||f^\zeta_{j}) - \left( I^\theta - \max_{\zeta \in \cM_\epsilon^\theta} D(f^{\theta}_{j} ||f^\zeta_{j}) - \min_{\zeta\in\Tilde{\Theta}\backslash\cM_{\epsilon}^{\theta}}B_{a^\circ(\zeta)}(\theta,\zeta)\right) \bbP_{1,A^\circ}^\theta \left(\hat{\theta}_t^{\circ} \in \Tilde{\Theta}\backslash\cM_{\epsilon}^\theta| \cF_{t-w-1}, A^{\circ}_{[1:w]} \right).
    \end{aligned}
\end{equation}

Since $\epsilon \to 0$ as $q\to \infty$, we have $\lim_{q\to\infty}\max_{\zeta \in \cM_\epsilon^\theta} D(f^{\theta}_{j} ||f^\zeta_{j}) \to 0$ and $I^{\theta} \geq B_a(\theta,\zeta)$ holds for any $\zeta \in \Tilde{\Theta}$ and $j\in 0\cup[p]$. Combine \eqref{eq:lem3 mainpf} and \eqref{eq:lem3 mainpf}, we get that for $t \notin\sN $, 
\begin{equation*}
     \liminf_{q\to\infty}\bbE_{1,A^{\circ}}^{\theta}[\Lambda_{t,A^{\circ}}^{\hat{\theta}_{t}^{\circ}}| \cF_{t-w-1}, A^{\circ}_{[w]}] \geq I^\theta.
\end{equation*}
Recall that $I^\theta = I_{j,c_j}^{\Delta,[k,j]}$ according to Proposition \ref{prop: Ia}, we complete the proof of Lemma \ref{lem:multi liminf} when $t\notin \sN$. 

For $t\in\sN$, similar to \eqref{eq:lem3 mainpf}, we have 
\begin{equation}
    \begin{aligned}
           &\quad \bbE_{1,A^{\circ}}^{\theta}[\Lambda_{t,A^{\circ}}^{\hat{\theta}_{t}^{\circ}}| \cF_{t-w-1}, A^{\circ}_{[w]}] \\
        & = \int_{\Tilde{\Theta}} \bbE_{1,A^\circ}^\theta\left[\log\left(\frac{f^{\zeta}_{a^\circ_t}(\bmY^{t,do(a^\circ_t)}_{[p]\backslash a^\circ_t})}{f^{\theta^0}_{a^\circ_t}(\bmY^{t,do(a^\circ_t)}_{[p]\backslash a^\circ_t})} \right) |\hat{\theta}_t^{\circ} = \zeta, \cF_{t-w-1}, A^{\circ}_{[1:w]}\right]\bbP_{1,A^\circ}^\theta \left(\hat{\theta}_t^{\circ} = \zeta| \cF_{t-w-1}, A^{\circ}_{[1:w]} \right) d\zeta\\
        &= \int_{\Tilde{\Theta}} \sum_{i=0}^p\frac{1}{p+1}B_{i}(\theta,\zeta) \bbP_{1,A^\circ}^\theta \left(\hat{\theta}_t^{\circ} = \zeta| \cF_{t-w-1}, A^{\circ}_{[1:w]} \right) d\zeta\\
        &= \frac{1}{p+1}\sum_{i=0}^p \left( \int_{\cM_\epsilon^{\theta}} \left(I^{\theta}_i - D(f^{\theta}_{i} ||f^\zeta_{i}) \right) \bbP_{1,A^\circ}^\theta \left(\hat{\theta}_t^{\circ} = \zeta| \cF_{t-w-1}, A^{\circ}_{[1:w]} \right) d\zeta \right.\\
        &\quad \left. + \int_{\Tilde{\Theta}\backslash\cM_\epsilon^{\theta}} 
        B_{i}(\theta,\zeta)
        \bbP_{1,A^\circ}^\theta \left(\hat{\theta}_t^{\circ} = \zeta| \cF_{t-w-1}, A^{\circ}_{[1:w]} \right) d\zeta \right)\\
        &\geq\frac{1}{p+1}\sum_{i=0}^p\left(  \left(I^{\theta}_i - \max_{\zeta \in \cM_\epsilon^\theta} D(f^{\theta}_{i} ||f^\zeta_{i}) \right) \int_{\cM_\epsilon^{\theta}}  \bbP_{1,A^\circ}^\theta \left(\hat{\theta}_t^{\circ} = \zeta| \cF_{t-w-1}, A^{\circ}_{[1:w]} \right) d\zeta \right.\\ 
        &\quad \left.+ \min_{\zeta\in\Tilde{\Theta}\backslash\cM_{\epsilon}^{\theta}}B_{i}(\theta,\zeta)\int_{\Tilde{\Theta}\backslash\cM_\epsilon^{\theta}} 
        \bbP_{1,A^\circ}^\theta \left(\hat{\theta}_t^{\circ} = \zeta| \cF_{t-w-1}, A^{\circ}_{[1:w]} \right) d\zeta\right) \\
        &= \frac{1}{p+1}\sum_{i=0}^p \left(I^\theta_i - \max_{\zeta \in \cM_\epsilon^\theta} D(f^{\theta}_{i} ||f^\zeta_{i}) \right.\\ 
        &\left.\qquad \qquad\qquad - \left( I^\theta_i - \max_{\zeta \in \cM_\epsilon^\theta} D(f^{\theta}_{a} ||f^\zeta_{a}) - \min_{\zeta\in\Tilde{\Theta}\backslash\cM_{\epsilon}^{\theta}}B_{i}(\theta,\zeta)\right) \bbP_{1,A^\circ}^\theta \left(\hat{\theta}_t^{\circ} \in \Tilde{\Theta}\backslash\cM_{\epsilon}^\theta| \cF_{t-w-1}, A^{\circ}_{[1:w]} \right) \right).
    \end{aligned}
\end{equation}

We already know that $\lim_{\epsilon \to 0} \max_{\zeta \in \cM_\epsilon^\theta} D(f^{\theta}_{i} ||f^\zeta_{i}) \to 0$ and combine \eqref{eq:lim P}, we get that for $t\in \sN$
\begin{equation*}
    \liminf_{q\to\infty}\bbE_{1,A^{\circ}}^{\theta}[\Lambda_{t,A^{\circ}}^{\hat{\theta}_{t}^{\circ}}| \cF_{t-w-1}, A^{\circ}_{[w]}] \geq \frac{1}{p+1}\sum_{a=0}^p I^\theta_a.
\end{equation*}
Recall that $I_i^\theta$ is the abbreviation for $I_{i,c_i}^{\Delta,[k,j]}$. So Lemma \ref{lem:multi liminf} holds when $t \in \sN$.

\end{proof}

\subsection{Proof of Theorem \ref{thm:multi}}
\label{app:multi thm}
Before starting the proof, we first introduce an important lemma.

\begin{lemma1}[Proposition 1 in \cite{fellouris2022quickest}]
\label{lem:ref}
Let $(\Omega, \cF, \bP)$ be a be a probability space, and let $\bE$ denote expectation with respect to $\bP$. Let $\{\cG_t| t\geq1 \}$ be a filtration on this space, where $\cG_0$ is the trivial  $\sigma$-algebra. Let $\{Z_t|t \geq1 \}$ a sequence of random variables in $\cL^2$ that is adapted to $\{\cG_t| t\geq1 \}$. Let $w \in \bbN$, $\sN \subset\{w+1,w+2,dots \}$, $\rho^*,\rho_*,\rho,\nu > 0$ and $q\in \bbN $ and suppose that, for every $t > w$,
\begin{align}
    \bE\left[Z_t|\cG_{t-w-1}\right] &\geq \begin{cases}
        \rho^*, \quad & t\notin\sN\\
        \rho_*, \quad & t \in \sN
    \end{cases}
    \text{   a.s.} \\
    \bE\left[Z_t|\cG_{t-1}\right] & \leq \rho \text{   a.s.} \\
    \bE\left[Z_t^2|\cG_{t-1}\right] & \leq \nu \text{   a.s.} 
\end{align}
and
\begin{equation*}
    \sum_{t = w+1}^{w+n} \bbI\{ t\in \sN\} \leq \frac{n}{w} q.
\end{equation*}
For any $b > 0$, set
\begin{equation}
    T_b\triangleq\inf \{t>w| S_t\geq b\}, \text{  where } S_t \triangleq\sum_{i = w+1}^t Z_i.
\end{equation}
Then: 
\begin{enumerate}
    \item $\bE[T_b] < \infty$
    \item There is a $H>0$, that does not depend on $b,q$, or $w$, so that 
    \begin{equation*}
        \bE[T_b] \leq \frac{b + w\rho + H\left(1+\sqrt{b} +\sqrt{w} \right)}{\rho^{*}(1-q/w)}.
    \end{equation*}
    \item If $\rho^* \to \rho$ as $q,w \to\infty$ so that $q = o(w)$, then 
    \begin{equation}
        \bE[T_b] \leq\frac{b}{\rho}(1+o(1)),
    \end{equation}
    where $o(1)$ is a vanishing term as $q,w,b \to \infty$ so that $q = o(w)$ and $w = o(b)$.
\end{enumerate}

\end{lemma1}

Now we start to prove Theorem \ref{thm:multi}
\begin{proof}
Define the stopping time 
\begin{equation*}
    \Tilde{T}_{b,A^\circ}^{multi} \triangleq \inf \{t>w| \Tilde{W}_{t,A^\circ} \geq b \}
\end{equation*}
with 
\begin{equation*}
    \Tilde{W}_{t,A^\circ} = \Tilde{W}_{t-1,A^\circ} + \Lambda_{t,A^\circ}^{\hat{\theta}_t^\circ}, \ t >w, \ \ \Tilde{W}_{w,A^\circ} = 0.
\end{equation*}
It is worth noting that the difference between $\Tilde{W}_{t,A^\circ}$ and $W_{t,A^\circ}$ lies in the iterative definition of the former, where the right-hand side of the equation is not restricted to the positive part only. This means that $T_{b,A^\circ}^{multi} \leq  \Tilde{T}_{b,A^\circ}^{multi}$. Fix the change magnitude $\Delta$ and change location $[k,j]$, $\theta\in\Theta$ is its post-change parameter.  If we want to show that 
\begin{equation}
    \label{eq: thm2 pf1}
    \sup_{a^{\circ}_1\dots,a^{\circ}_w}\bbE_{1,A^\circ}^\theta \left[T_{b,A^\circ}^{multi}|A_{[w]}^{\circ}  \right] \leq \frac{b}{I_{j,c_j}^{\Delta, [k,j]}}(1 + o(1)),
\end{equation}
as $q,w,b\to \infty$ and $q = o(w),w = o(b)$, we only to show 
\begin{equation*}
    \sup_{a^{\circ}_1\dots,a^{\circ}_w}\bbE_{1,A^\circ}^\theta \left[\Tilde{T}_{b,A^\circ}^{multi}|A_{[w]}^{\circ}\right] \leq \frac{b}{I_{j,c_j}^{\Delta, [k,j]}}(1 + o(1)),
\end{equation*}
as $q,w,b\to \infty$ and $q = o(w),w = o(b)$. In what follows, we fixed arbitrary $a^{\circ}_1,\dots,a^{\circ}_w$ in $0\cup[p]$ and we show 
\begin{equation}
\label{eq: thm2 pf2}
    \bbE_{1,A^\circ}^\theta \left[\Tilde{T}_{b,A^\circ}^{multi}|A_{[w]}^{\circ} \right] \leq \frac{b}{I_{j,c_j}^{\Delta, [k,j]}}(1 + o(1)).
\end{equation}
as $q,w,b\to \infty$ and $q = o(w),w = o(b)$.

According to Lemma \ref{lem:multi liminf}, we know that for $q$ large enough, there exist the positive lower bound of $\bbE_{1,A^{\circ}}^{\theta}[\Lambda_{t,A^{\circ}}^{\hat{\theta}_{t}^{\circ}}| \cF_{t-w-1}, A^{\circ}_{[w]}]$, we denote it as $J^q$ for $t\notin\sN$ and $K^q$ for $t\in\sN$. It is clear that $J^q$ converge to $I_{j,c_j}^{\Delta,[k,j]}$ and $K^q$ converge to $\frac{1}{p+1}\sum_{i=0}^p I_{i,c_i}^{\Delta,[k,j]}$ as $q\to \infty$.

Then using the following identifications in Lemma \ref{lem:ref}:
\begin{equation*}
    \begin{aligned}
        \bP &\leftrightarrow \bbP_{1,A^{\circ}}^{\theta}\left(\cdot|A_{[w]}^{\circ} \right), \\
        \cG_t&\leftrightarrow \cF_t, \\
        Z_t & \leftrightarrow \Lambda_{t,A^\circ}^{\hat{\theta}_t^\circ} ,\\
        T_b &\leftrightarrow \Tilde{T}_{b,A^\circ}^{multi},
    \end{aligned}
\end{equation*}
and set $\rho^*, \rho_*, \rho, \nu$ as follows
\begin{equation*}
    \begin{aligned}
        \rho^* &\leftrightarrow J^q,\\
        \rho_* &\leftrightarrow K^q, \\
        \rho & \leftrightarrow I_{j,c_j}^{\Delta,[k,j]}, \\
        \nu & \leftrightarrow V.
    \end{aligned}
\end{equation*}
Thus, we have  \eqref{eq: thm2 pf2} holds based on Lemma \ref{lem:ref}.  We further establish \eqref{eq: thm2 pf1}. Combining this with Lemma \ref{lem:multi IC}, Lemma \ref{lem:multi OC}, Theorem \ref{thm:general} and setting $b = \log\gamma$, the proof is complete.

\end{proof}

\section{Proof of Lemmas \ref{lem:max IC}-\ref{lem:max liminf} and Theorem \ref{thm:max}}
\label{app:max}
\textbf{In this section, we use the additional notations defined in Appendix \ref{app:notation}}
and we  prove Lemmas \ref{lem:max IC}-\ref{lem:max liminf} and Theorem \ref{thm:max} in Section \ref{sec: thm sub3}.


\subsection{Proof of Lemma \ref{lem:max IC}}
\label{app:max IC}
\begin{proof}
For each $l \in [p]$, we have 
\begin{equation*}
    e^{W_{t,A^{*}}^l} = \max\{e^{W_{t-1,A^{*}}^l} ,1\} \frac{f^{\hat{\theta}_t^*}_{a_t^*}[l](Y_l^{t,do(a_t^{*})})}{f^{\theta^0}_{a_t^*}[l](Y_l^{t,do(a_t^*)})}, \ t > w, \ e^{W_{w,A^{*}}} = 1. 
\end{equation*}
Define 
\begin{equation*}
    \cW_{t,A^*}^l = (\cW_{t,A^*}^l+1) \frac{f^{\hat{\theta}_t^*}_{a_t^*}[l](Y_l^{t,do(a_t^{*})})}{f^{\theta^0}_{a_t^*}[l](Y_l^{t,do(a_t^*)})}, \ t > w, \ \cW_{t,A^*}^l = 1.
\end{equation*}

It is easy to find that $e^{W_{b,A^*}^l} \leq \cW_{b,A^*}^l$ for $l \in [p]$.

Because $\hat{\theta}_t^{*}$ is $\cF_{t-1}$-measurable, we have
\begin{equation*}
    \bbE_{\infty,A^{*}}[\cW_{t,A^{*}}^l - t| \cF_{t-1}] = \cW_{t-1,A^{*}}^l + 1-t = \cW_{t-1,A^{*}}^l - (t-1), \quad  \text{for } l \in[p].
\end{equation*}
This means that $\{\cW_{t,A^{*}}^l - t| t \geq w\}$ is a martingale under $\bbP_{\infty,A^{*}}$, with respect to the filtration $\{\cF_t |t\in\bbN \}$. $\bbP_{\infty,A}(\cdot)$ denotes the probability measure under the pre-change condition with sequential intervention policy $A$.

Then using the Optional Sampling Theorem \citep{tartakovsky2014sequential}, we have, for each $l\in [p]$, 
\begin{equation}
\label{eq:lem4 pf1}
    \bbE_{\infty,A^{*}}[ \cW^l_{T_{b,A^*}^{max}, A^*} - T_{b,A^*}^{max}] = \cW_{w,A^*}^l - w = 1 - w.
\end{equation}

We also have that 
\begin{equation*}
    e^b \leq e^{\max_{l\in[p]}W^l_{T_{b,A^*}^{max},A^*}} \leq \max_{l\in[p]} \cW^l_{T_{b,A^*}^{max},A^*} \leq \sum_{l=1}^p \cW^l_{T_{b,A^*}^{max},A^*}.
\end{equation*}
Then take expectation under $\bbP_{\infty,A^*}$ and combine \eqref{eq:lem4 pf1}. We can get
\begin{equation*}
    e^b \leq \sum_{l=1}^p(\bbE_{\infty,A^*}[T_{b,A^*}^{max}] +1 -w) \leq p \bbE_{\infty,A^*}[T_{b,A^*}^{max}].
\end{equation*}
So we can set $b = \log\gamma + \log p$ to ensure $(A^*, T_{b,A^*}^{max}) \in \cC_{\gamma}$.

\end{proof}

\subsection{Proof of Lemma \ref{lem:max OC}}
\label{app:max OC}

\begin{proof}
For each $\tau \in \bbN$, we can define a new stopping time $T_{b,A^{*},\tau}^{k}$ as follows
\begin{equation*}
    T_{b,A^{*},\tau}^{k} = \inf\{t>\tau +w-1 | W_{t,A^{*}}^{\tau} > b \},
\end{equation*}
where 
\begin{equation*}
    W_{t,A^{*}}^{\tau} = \max\{W_{t-1,A^*}^{\tau} ,0\} + \Lambda_{t,A^*}^{\hat{\theta}_t^*}[k], \ t>\tau+w-1, \quad W_{\tau+w-1,A^*}^{\tau} = 0.
\end{equation*}

Define $x^+ = \max\{x,0\}$, for any $\tau \in \bbN$ and $\theta \in \Theta$, we have
\begin{equation*}
    \begin{aligned}
        \bbE_{\tau,A^{*}}^{\theta}\left[(T_{b,A^{*}}^{k} - \tau + 1)^+ | \cF_{\tau-1}\right] &\leq w + \bbE_{\tau,A^{*}}^{\theta}\left[(T_{b,A^{*}}^{k} - \tau + 1 + w)^+ | \cF_{\tau-1}\right] \\
        & \leq w +   \bbE_{\tau,A^{*}}^{\theta}\left[T_{b,A^{*},\tau}^{k} - \tau + 1 + w | \cF_{\tau-1}\right] \\
        & = \bbE_{\tau,A^{*}}^{\theta}\left[ \bbE_{\tau,A^{*}}^{\theta}\left[T_{b,A^{*},\tau}^{k} - \tau + 1  |A^{*}_{[\tau,\tau+w-1]}, \cF_{\tau-1}\right]  | \cF_{\tau-1}\right] \\
        &= \bbE_{\tau,A^{*}}^{\theta}\left[ \bbE_{\tau,A^{*}}^{\theta}\left[T_{b,A^{*},\tau}^{k} - \tau + 1  |A^{*}_{[\tau,\tau+w-1]}\right]  | \cF_{\tau-1}\right] \\
        &\leq \sup_{a_{\tau}^{*},\dots,a_{\tau+w-1}^{*}} \bbE_{\tau,A^{*}}^{\theta}\left[T_{b,A^{*},\tau}^{k} - \tau + 1  |A^{*}_{[\tau,\tau+w-1]}\right]\\
        & = \sup_{a_1^{*},\dots,a_w^{*}} \bbE_{1,A^{*}}^{\theta}\left[T_{b,A^{*}}^{k}  |A^{*}_{[w]}  \right],
    \end{aligned}
\end{equation*}
where $A^{*}_{[c,d]} \triangleq \{a_t^{*}| c\leq t\leq d \}$. The first inequality holds because $x^+ < y+(x-y)^+$ for $x,y>0$. The second inequality holds because $T_{b,A^{*}}^{k} \leq T_{b,A^{*},\tau}^{k}$ since $W_{t,A^{*}}^k \geq W_{t,A^{*}}^\tau$. The first equality follows the law of iterated expectation. The  second holds because $T_{b,A^{*},\tau}^{k}$ only depends on $A^{*}_{[\tau,\tau+w-1]}$. The last equality follows the definition of $T_{b,A^{*},\tau}^{k}$.
\end{proof}
\subsection{Proof of Lemma \ref{lem:max arm j}}
\label{app:max arm j}
\begin{proof}
We first recall the notation. Fix the change magnitude $\Delta$ and change location $[k,j]$, $\theta = \theta^{\Delta,[k,j]} = \{(\bmu_{Y,[p]\backslash i}^{\Delta,[k,j],do(i)},[\bSigma_Y^{\Delta,[k,j],do(i)}]_{[p]\backslash i,[p]\backslash i}) \}_{i=0}^p \in \Theta$ is the post-change parameter. $\bmu_{Y,[p]\backslash i}^{\Delta,[k,j],do(i)} \in \bbR^{|[p]\backslash i|}$ is the mean of $\bmY^{t,do(X_i^t=c_i)}_{[p]\backslash i}$ for $t \geq \tau$ and for convenience, we use $\bmu_Y^{\theta,do(i)}$ as its abbreviation. $[\bSigma_Y^{\Delta,[k,j],do(i)}]_{[p]\backslash i,[p]\backslash i} \in \bbR^{|[p]\backslash i| \times |[p]\backslash i|}$ is the covariance matrix of $\bmY^{t,do(X_i=c_i)}_{[p]\backslash i}$ for $t \geq \tau$, and for convenience, we use $\bSigma_Y^{\theta,do(i)}$ as its abbreviation. Since the dimensions of $\bSigma_Y^{\theta,do(i)}$ and $\bSigma_Y^{\Delta,[k,j],do(i)}$ may differ when \( i \neq 0 \), we by default index $\bSigma_Y^{\theta,do(i)}$  based on its corresponding indices in $\bSigma_Y^{\Delta,[k,j],do(i)}$ in subsequent steps. The same applies to $\bmu_Y^{\theta,do(i)}$ and $\bmu_Y^{\Delta,[k,j],do(i)}$ as well.  $\Tilde{\theta} = \{(\Tilde{\bmu}^i,\Tilde{\bSigma}^i )_{i=0}^p \} \in \Tilde{\Theta}$ is a parameter in the extend post-change parameter space. In the computation, we use MLE to find an optimal $\Tilde{\theta}\in\Tilde{\Theta}$ for estimating $\theta$, where $\Tilde{\bmu}^i$ is the estimate of $\bmu_Y^{\theta,do(i)}$, and $\Tilde{\bSigma}^i$ is the estimate of $\bSigma_Y^{\theta,do(i)}$. Similarly, we index their dimensions using the same way as $\bmu_Y^{\theta,do(i)}$ and $\bSigma_Y^{\theta,do(i)}$.

In this section, we consider $\bmC$ is the outcome of Algorithm \ref{alg: c_j}. Additionally, for $\Tilde{\theta} \in \Tilde{\Theta}$, we set $I_{i}^{\Tilde{\theta}}[l] \triangleq D(f_{i}^{\Tilde{\theta}}[l]||f_{i}^{\theta^0}[l] )$, $i\in 0\cup[p]$, $l\in[p]\backslash i$.  Then we can use $I_i^{\theta}[l]$ as the abbreviation of $I_{i,c_i}^{\Delta, [k,j]}[l]$ and $I^{\theta}[l] = \max_{i\in0\cup[p]} I_i^{\theta}[l]$. According to Proposition \ref{prop: Ia}, we have $I^{\theta}[l] = I_{j,c_j}^{\Delta,[k,j]}[l]$ for $l\in[p]$ when change occurs at location $[k,j]$. 

Then for any $\theta \in \Theta, \zeta \in \Tilde{\Theta}$ and $i\in 0\cup [p]$, $l\in[p]$, we set 
\begin{equation}
    B_i(\theta[l],\zeta[l])\triangleq \int \log\left(\frac{f_i^{\zeta}[l](x)}{f_i^{\theta^0}[l](x)} \right)f_i^{\theta}[l]dx = D(f_i^{\theta}[l]||f_i^{\theta^0}[l]) - D(f_i^{\theta}[l]||f_i^{\zeta}[l]) = I_i^{\theta}[l] - D(f_i^{\theta}[l]||f_i^{\zeta}[l]),
\end{equation}
where $\theta[l]$ and $\zeta[l]$ denote the $l$th pair parameter in $\theta$ and $\zeta$ respectively. It is clear that $B_i(\theta[l], \theta[l]) = I_i^{\theta}[l]$ and $I_i^\theta[l] \geq B_i(\theta[l],\zeta[l])$.
We also set $a^{*}(\zeta)$ as the selected node to intervene on according to \eqref{eq:inter max} by
replacing $\hat{f}_{i}^{t,\Delta,[k,j],*}[l] \doteq f_i^{\hat{\theta}_t^{*}}[l]$ by $f^{\zeta}_i[l]$,
for $t \notin \sN $.

We next define a neighbor of $\theta \in \Theta$, which is the post-change parameter corresponding to the change magnitude $\Delta$ and change location $[k,j]$. For a fixed $q \in\bbN$, define $\epsilon = \epsilon^q = \min\{ \delta/3, q^{-\eta/4}\}$, and $\cN_{\epsilon}^\theta$ as the $\epsilon$-neighbor set of $\theta$ which consists of $\Tilde{\theta}= \{(\Tilde{\bmu}^i,\Tilde{\bSigma}^i )_{i=0}^p \} $ that satisfies the following conditions,
\begin{align}
\label{eq:cN 1}
    |(\Tilde{\mu}^i_j)^2 - (\mu_{Y,j}^{\theta,do(i)})^2| < \epsilon&\text{ for }i \in 0\cup [p]\text{ and }j \in [p]\backslash i.\\
    |[\Tilde{\bSigma}^i]_{j,j}-[\bSigma_{Y}^{\theta,do(i)}]_{j,j}|<\epsilon &\text{ for }i \in 0\cup [p]\text{ and }j \in [p]\backslash i.\\
    \label{eq:cN 2}
    |\log\left([\Tilde{\bSigma}^i]_{j,j}\right)-\log\left([\bSigma_{Y}^{\theta,do(i)}]_{j,j}\right)|<\epsilon &\text{ for }i \in 0\cup [p]\text{ and }j \in [p]\backslash i.
\end{align}
Consider the bounded constant $M$ in $\Tilde{\Theta}$ is large enough, we can say $\cN_\epsilon^\theta \subset \Tilde{\Theta}$. Obviously, $\cN_{\epsilon}^{\theta}$ is not empty since at least $\theta \in \cN_{\epsilon}^{\theta}$. Then we provide two important properties of $\cN_q^\theta$.
The first is, if $\zeta \in \cN_{\epsilon}^\theta$,  we can find that,
\begin{equation}
\label{eq:neighbour 1}
     |I_i^{\theta}[l] - I_i^{\zeta}[l]| < \delta/2, 
\end{equation}
holds for any $i \in 0\cup [p]$ and $l \in [p]$. Combining this with Proposition \ref{prop: Ia}, we can further get $a^{*}(\zeta) = j$ when the change occurs at location $[k,j]$. 

The second property is,  for any $i\in0\cup[p]$ and $l \in [p]$
\begin{equation}
\label{eq:neighbour 2}
     \lim_{\epsilon \to 0}\max_{\zeta\in\cN_\epsilon^{\theta}} D(f_i^\theta[l]||f_i^\zeta[l]) = 0, 
\end{equation}
which can be directly derived from the definition of the KL divergence between two normal distributions.

By \eqref{eq:neighbour 1} and Proposition \ref{prop: Ia}, to prove Lemma \ref{lem:max arm j}, we only to need to prove

\begin{equation}
\label{eq:lem5 lim P}
    \lim_{q,w\to \infty} \bbP_{1,A^{*}}^\theta \left(\hat{\theta}_t^{*} \in \cN_{\epsilon}^\theta| \cF_{t-w-1}, A^{*}_{[w]} \right) = 1.
\end{equation}

Since $\cN_\epsilon^\theta \in \Tilde{\Theta}$, according to the MLE property of normal distribution, we can write the $i$th pair $(\hat{\bmu}_{Y}^{t,do(i),*}, \hat{\bSigma}_{Y}^{t,do(i),*})$ in $\hat{\theta}_t^{*}$ as
\begin{equation}
    \begin{aligned}
        \hat{\bmu}_{Y}^{t,do(i),*} &= \frac{\sum_{n=t-w}^{t-1} \bmY^{n,do(a_n^*)}_{[p]\backslash a_n^*}\bbI\{a_n^* = i\}}{\sum_{n=t-w}^{t-1} \bbI\{a_n^{*} = i\}}, \\
        \hat{\bSigma}_{Y}^{t,do(i),*} &= \frac{\sum_{n=t-w}^{t-1}(\bmY^{n,do(a_n^*)}_{[p]\backslash a_n^*} - \hat{\bmu}_{Y}^{t,do(i),*}) (\bmY^{n,do(a_n^*)}_{[p]\backslash a_n^*} - \hat{\bmu}_{Y}^{t,do(i),*})^T  \bbI\{a_n^* = i\}}{\sum_{n=t-w}^{t-1} \bbI\{a_n^{*} = i\}},
    \end{aligned}
\end{equation}

For convenience, we denote $n_i = \sum_{n=t-w}^{t-1} \bbI\{a_n^{*} = i\}$ for $i\in 0\cup[p]$. Since when $t\in\sN$, we uniformly random select intervention node, we have $n_i = \Omega(q^{\eta})$ as $q\to \infty$.

Combine \eqref{eq:cN 1} to \eqref{eq:cN 2}, we have that 
\begin{equation}
\label{eq:I II III}
\begin{aligned}
     \bbP_{1,A^*}^\theta \left(\hat{\theta}_t^{*} \in \cN_{\epsilon}^\theta| \cF_{t-w-1}, A^{*}_{[w]} \right) \geq 1 
     & - \sum_{i=0}^p\sum_{j=1,j\neq i}^p\underbrace{\bbP_{1,A^*}^\theta \left( |(\hat{\mu}_{Y,j}^{t,do(i),*})^2 - (\mu_{Y,j}^{\theta,do(i)})^2| \geq \epsilon\right)}_{I} \\ 
     & - \sum_{i=0}^p\sum_{j=1,j\neq i}^p\underbrace{\bbP_{1,A^*}^\theta \left( |[\hat{\bSigma}_{Y}^{t,do(i),*}]_{j,j}-[\bSigma_{Y}^{\theta,do(i)}]_{j,j}|\geq\epsilon\right)}_{II} \\ 
     & - \sum_{i=0}^p\sum_{j=1,j\neq i}^p\underbrace{\bbP_{1,A^*}^\theta \left( |\log\left([\hat{\bSigma}_{Y}^{t,do(i),*}]_{j,j}\right)-\log\left([\bSigma_{Y}^{\theta,do(i)}]_{j,j}\right)|\geq\epsilon\right)}_{III}.
\end{aligned}
\end{equation}
Next, we will prove term $I$ to term $III$ converge to $0$ as $q \to \infty$ using the centralization properties of the sample mean and sample covariance matrix of the normal distribution.

Consider the term $I$, using Markov inequality, we have
\begin{equation*}
\begin{aligned}
    \bbP_{1,A^*}^\theta \left( |(\hat{\mu}_{Y,j}^{t,do(i),*})^2 - (\mu_{Y,j}^{\theta,do(i)})^2| \geq \epsilon\right) &\leq \frac{1}{\epsilon^2}\left(Var\left((\hat{\mu}_{Y,j}^{t,do(i),*})^2\right) + \left(\bbE_{1,A^*}^\theta[(\hat{\mu}_{Y,j}^{t,do(i),*})^2] - (\mu_{Y,j}^{\theta,do(i)})^2\right)^2 \right) \\
    & =\frac{1}{\epsilon^2} \left(\frac{2[\bSigma_{Y}^{\theta,do(i)}]_{j,j}^2 + 4n_i[\bSigma_{Y}^{\theta,do(i)}]_{j,j} (\mu_{Y,j}^{\theta,do(i)})^2 }{n_i^2} +\frac{[\bSigma_Y^{\theta,do(i)}]_{j,j}^2}{n_i^2} \right)\\
    &= O(q^{-\eta/2})
\end{aligned}
\end{equation*}
converges to $0$ as $q\to \infty$.

Consider the term $II$, using the Markov inequality, we have
\begin{equation*}
    \begin{aligned}
        \bbP_{1,A^*}^\theta \left( |[\hat{\bSigma}_{Y}^{t,do(i),*}]_{j,j}-[\bSigma_{Y}^{\theta,do(i)}]_{j,j}|\geq\epsilon\right) & \leq \frac{1}{\epsilon^2} \left(Var\left(\hat{\bSigma}_{Y}^{t,do(i),*}]_{j,j} \right) + \left(\bbE_{1,A^{*}}^{\theta}\left[\hat{\bSigma}_{Y}^{t,do(i),*}]_{j,j} \right] - [\bSigma_{Y}^{\theta,do(i)}]_{j,j} \right)^2   \right)  \\
        & = \frac{1}{\epsilon^2} \left( \frac{(2n_i-2)[\bSigma_Y^{\theta,do(i)}]_{j,j}^2}{n_i^2}+\frac{[\bSigma_Y^{\theta,do(i)}]_{j,j}^2}{n_i^2}\right) \\
        & = O(q^{-\eta/2})
    \end{aligned}
\end{equation*}
converges to $0$ as $q\to \infty$.

Consider the term $III$, using the Markov inequality, we have
\begin{equation}
\label{eq:III 1}
    \begin{aligned}
    \bbP_{1,A^*}^\theta &\left(|\log\left([\hat{\bSigma}_{Y}^{t,do(i),*}]_{j,j}\right)-\log\left([\bSigma_{Y}^{\theta,do(i)}]_{j,j}\right)|\geq\epsilon\right)\\ 
    &\leq \frac{1}{\epsilon^2} \left(Var\left(\log\left([\hat{\bSigma}_{Y}^{t,do(i),*}]_{j,j}\right)\right) + \left(\bbE_{1,A^*}^\theta\left[\log\left([\hat{\bSigma}_{Y}^{t,do(i),*}]_{j,j}\right) \right] - \log\left([\bSigma_{Y}^{\theta,do(i)}]_{j,j}\right) \right)^2\right).
    \end{aligned}
\end{equation}
Since $\log\left(n_i \frac{[\hat{\bSigma}_{Y}^{t,do(i),*}]_{j,j}}{[\bSigma_{Y}^{\theta,do(i)}]_{j,j}} \right) \sim \log \chi^2(n_i-1)$, i.e., the log chi-squared distribution with $n_i-1$ degrees of freedom, we have
\begin{equation*}
    \begin{aligned}
        \bbE_{1,A^{*}}^{\theta} \left[\log\left([\hat{\bSigma}_{Y}^{t,do(i),*}]_{j,j}\right) \right] &= \log\left([\bSigma_{Y}^{\theta,do(i)}]_{j,j}\right) - \log n_i + \log 2 + \psi(\frac{n_i-1}{2}), \\
        Var\left(\log\left([\hat{\bSigma}_{Y}^{t,do(i),*}]_{j,j}\right) \right) & = \psi^{'}(\frac{n_i-1}{2}),
    \end{aligned}
\end{equation*}
where $\psi$ is the degamma function, i.e., $\psi(x) = \log x - \frac{1}{2x} + o(\frac{1}{x})$. $\psi^{'}$is the trigamma function, i.e., $\psi^{'}(x) = \frac{1}{x} + o(\frac{1}{x})$. Thus, we have 
\begin{equation}
\label{eq:III 2}
    \begin{aligned}
        \bbE_{1,A^*}^\theta\left[\log\left([\hat{\bSigma}_{Y}^{t,do(i),*}]_{j,j}\right) \right] - \log\left([\bSigma_{Y}^{\theta,do(i)}]_{j,j}\right) &= O(\frac{1}{n_i}), \\
        Var\left(\log\left([\hat{\bSigma}_{Y}^{t,do(i),*}]_{j,j}\right)\right) &= O(\frac{1}{n_i}).
    \end{aligned}
\end{equation}

Then combine \eqref{eq:III 1} and \eqref{eq:III 2}, we get 
\begin{equation*}
    \bbP_{1,A^*}^\theta \left(|\log\left([\hat{\bSigma}_{Y}^{t,do(i),*}]_{j,j}\right)-\log\left([\bSigma_{Y}^{\theta,do(i)}]_{j,j}\right)|\geq\epsilon\right) \leq O(q^{-\eta/2})
\end{equation*}
converges to $0$ as $q \to \infty$.

Now we have \eqref{eq:lem5 lim P}, and complete the proof.
\end{proof}

\subsection{Proof of Lemma \ref{lem:max liminf}}
\label{app:max liminf}
\begin{proof}

In this section, we use the same notations as the last section.

When $t \notin \sN$, for $l\in[p]$, by \eqref{eq:neighbour 1} and \eqref{eq:neighbour 2}, we have 
\begin{equation}
\label{eq:lem5 mainpf}
    \begin{aligned}
         &\quad \bbE_{1,A^{*}}^{\theta}[\Lambda_{t,A^{*}}^{\hat{\theta}_{t}^{*}}[l]| \cF_{t-w-1}, A^{*}_{[w]}] \\
        & = \int_{\Tilde{\Theta}} \bbE_{1,A^*}^\theta\left[\log\left(\frac{f^{\zeta}_{a^{*}(\zeta)}[l](Y^{t,do(a^{*}(\zeta))}_{l})}{f^{\theta^0}_{a^{*}(\zeta)}[l](Y^{t,do(a^{*}(\zeta))}_{l})} \right) |\hat{\theta}_t^{*} = \zeta, \cF_{t-w-1}, A^{*}_{[w]}\right]\bbP_{1,A^{*}}^\theta \left(\hat{\theta}_t^{*} = \zeta| \cF_{t-w-1}, A^{*}_{[w]} \right) d\zeta\\
        &= \int_{\Tilde{\Theta}} B_{a^{*}(\zeta)}(\theta[l],\zeta[l]) \bbP_{1,A^*}^\theta \left(\hat{\theta}_t^{*} = \zeta| \cF_{t-w-1}, A^{*}_{[w]} \right) d\zeta\\
        &=  \int_{\cN_\epsilon^{\theta}} \left(I^{\theta}[l] - D(f^{\theta}_{j}[l] ||f^\zeta_{j}[l]) \right) \bbP_{1,A^*}^\theta \left(\hat{\theta}_t^{*} = \zeta| \cF_{t-w-1}, A^{*}_{[w]} \right) d\zeta \\
        &\quad + \int_{\Tilde{\Theta}\backslash\cN_\epsilon^{\theta}} 
        B_{a^{*}(\zeta)}(\theta[l],\zeta[l])
        \bbP_{1,A^{*}}^\theta \left(\hat{\theta}_t^{*} = \zeta| \cF_{t-w-1}, A^{*}_{[w]} \right) d\zeta \\
        &\geq \left(I^{\theta}[l] - \max_{\zeta \in \cN_\epsilon^\theta} D(f^{\theta}_{j}[l] ||f^\zeta_{j}[l]) \right) \int_{\cN_\epsilon^{\theta}}  \bbP_{1,A^{*}}^\theta \left(\hat{\theta}_t^{*} = \zeta| \cF_{t-w-1}, A^{*}_{[w]} \right) d\zeta \\ 
        &\quad + \min_{\zeta\in\Tilde{\Theta}\backslash\cN_{\epsilon}^{\theta}}B_{a^{*}(\zeta)}(\theta[l],\zeta[l])\int_{\Tilde{\Theta}\backslash\cN_\epsilon^{\theta}} 
        \bbP_{1,A^{*}}^\theta \left(\hat{\theta}_t^{*} = \zeta| \cF_{t-w-1}, A^{*}_{[w]} \right) d\zeta \\
        &= I^\theta[l] - \max_{\zeta \in \cN_\epsilon^\theta} D(f^{\theta}_{j}[l] ||f^\zeta_{j}[l]) -  \\ 
        &\qquad \left( I^\theta[l] - \max_{\zeta \in \cN_\epsilon^\theta} D(f^{\theta}_{j}[l] ||f^\zeta_{j}[l]) - \min_{\zeta\in\Tilde{\Theta}\backslash\cN_{\epsilon}^{\theta}}B_{a^{*}(\zeta)}(\theta[l],\zeta[l])\right) \bbP_{1,A^{*}}^\theta \left(\hat{\theta}_t^{*} \in \Tilde{\Theta}\backslash\cN_{\epsilon}^\theta| \cF_{t-w-1}, A^{*}_{[w]} \right).
    \end{aligned}
\end{equation}
Since $\epsilon \to 0$ as $q\to \infty$, we have $\lim_{q\to\infty}\max_{\zeta \in \cN_\epsilon^\theta} D(f^{\theta}_{i}[l] ||f^\zeta_{i}[l]) \to 0$ and $I^{\theta}[l] \geq B_i(\theta[l],\zeta[l])$ holds for any $\zeta \in \Tilde{\Theta}$, $i\in 0\cup[p]$ and $l\in[p]$. 
 Combine \eqref{eq:lem5 lim P} with \eqref{eq:lem5 mainpf}, we can get 
\begin{equation*}
     \liminf_{q\to\infty}\bbE_{1,A^{*}}^{\theta}[\Lambda_{t,A^{*}}^{\hat{\theta}_{t}^{*}}[l]| \cF_{t-w-1}, A^{*}_{[w]}] \geq I^\theta[l]
\end{equation*}
holds for $t\notin\sN$. Recalling $I^\theta[l] = I^{\Delta,[k,j]}_{j,c_j}[l]$, we complete the proof of Lemma \ref{lem:max liminf} when $t\notin \sN$.

Then we prove Lemma \ref{lem:max liminf}  when $t\in\sN$. For $t\in\sN$, $l\in[p]$, similar to \eqref{eq:lem5 mainpf}, we have 
\begin{equation}
\label{eq:lem5 3}
    \begin{aligned}
           &\quad \bbE_{1,A^{*}}^{\theta}[\Lambda_{t,A^{*}}^{\hat{\theta}_{t}^{*}}[l]| \cF_{t-w-1}, A^{*}_{[w]}] \\
        & = \int_{\Tilde{\Theta}} \bbE_{1,A^{*}}^\theta\left[\log\left(\frac{f^{\zeta}_{a^{*}_t}[l](Y^{t,do(a^{*}_t)}_l)}{f^{\theta^0}_{a^{*}_t}[l](Y^{t,do(a^{*}_t)}_{l})} \right) |\hat{\theta}_t^{*} = \zeta, \cF_{t-w-1}, A^{*}_{[w]}\right]\bbP_{1,A^{*}}^\theta \left(\hat{\theta}_t^{*} = \zeta| \cF_{t-w-1}, A^{*}_{[w]} \right) d\zeta\\
        &= \int_{\Tilde{\Theta}} \sum_{i=0}^p\frac{1}{p+1}B_{i}(\theta[l],\zeta[l]) \bbP_{1,A^{*}}^\theta \left(\hat{\theta}_t^{*} = \zeta| \cF_{t-w-1}, A^{*}_{[w]} \right) d\zeta\\
        &= \frac{1}{p+1}\sum_{i=0}^p \left( \int_{\cN_\epsilon^{\theta}} \left(I^{\theta}_i[l] - D(f^{\theta}_{i}[l] ||f^\zeta_{i}[l]) \right) \bbP_{1,A^{*}}^\theta \left(\hat{\theta}_t^{*} = \zeta| \cF_{t-w-1}, A^{*}_{[w]} \right) d\zeta \right.\\
        &\quad \left. + \int_{\Tilde{\Theta}\backslash\cN_\epsilon^{\theta}} 
        B_{i}(\theta[l],\zeta[l])
        \bbP_{1,A^{*}}^\theta \left(\hat{\theta}_t^{*} = \zeta| \cF_{t-w-1}, A^{*}_{[w]} \right) d\zeta \right)\\
        &\geq\frac{1}{p+1}\sum_{i=0}^p\left(  \left(I^{\theta}_i[l] - \max_{\zeta \in \cN_\epsilon^\theta} D(f^{\theta}_{i}[l] ||f^\zeta_{i}[l]) \right) \int_{\cN_\epsilon^{\theta}}  \bbP_{1,A^{*}}^\theta \left(\hat{\theta}_t^{*} = \zeta| \cF_{t-w-1}, A^{*}_{[w]} \right) d\zeta \right.\\ 
        &\quad \left.+ \min_{\zeta\in\Tilde{\Theta}\backslash\cN_{\epsilon}^{\theta}}B_{i}(\theta[l],\zeta[l])\int_{\Tilde{\Theta}\backslash\cN_\epsilon^{\theta}} 
        \bbP_{1,A^{*}}^\theta \left(\hat{\theta}_t^{*} = \zeta| \cF_{t-w-1}, A^{*}_{[w]} \right) d\zeta\right) \\
        &= \frac{1}{p+1}\sum_{i=0}^p \left(I^\theta_i[l] - \max_{\zeta \in \cN_\epsilon^\theta} D(f^{\theta}_{i}[l] ||f^\zeta_{i}[l]) \right.\\ 
        &\left.\qquad \qquad\qquad - \left( I^\theta_i[l] - \max_{\zeta \in \cN_\epsilon^\theta} D(f^{\theta}_{i}[l] ||f^\zeta_{i}[l]) - \min_{\zeta\in\Tilde{\Theta}\backslash\cN_{\epsilon}^{\theta}}B_{i}(\theta[l],\zeta[l])\right) \bbP_{1,A^{*}}^\theta \left(\hat{\theta}_t^{*} \in \Tilde{\Theta}\backslash\cN_{\epsilon}^\theta| \cF_{t-w-1}, A^{*}_{[w]} \right) \right).
    \end{aligned}
\end{equation}
We already know that $\lim_{\epsilon \to 0} \max_{\zeta \in \cN_\epsilon^\theta} D(f^{\theta}_{i}[l] ||f^\zeta_{i}[l]) \to 0$. Combine \eqref{eq:lem5 lim P} and \eqref{eq:lem5 3}, we get that for $t\in \sN$
\begin{equation*}
    \liminf_{q\to\infty}\bbE_{1,A^{*}}^{\theta}[\Lambda_{t,A^{*}}^{\hat{\theta}_{t}^{*}}[l]| \cF_{t-w-1}, A^{*}_{[w]}] \geq \frac{1}{p+1}\sum_{i=0}^p I^\theta_i[l].
\end{equation*}
Recall that $I_i^\theta[l]$ is the abbreviation for $I_{i,c_i}^{\Delta,[k,j]}[l]$. So we complete the proof of  Lemma \ref{lem:max liminf} when $t \in \sN$.
\end{proof}

\subsection{Proof of Theorem \ref{thm:max}}
\label{app:max thm}
\begin{proof}

Fixed change magnitude $\Delta$ and change location $[k,j]$, $\theta\in\Theta$ is its post-change parameter. Using Lemma \ref{lem:max liminf}, Lemma \ref{lem:ref} and applying proof technique similar to that  employed in deriving \eqref{eq: thm2 pf1} in Appendix \ref{app:multi thm}, we can obtain:
\begin{equation*}
    \sup_{\nu_1\dots,\nu_w}\bbE_{1,A^\circ}^\theta \left[T_{b,A^\circ}^{k}|a_1^\circ =\nu_1 ,\dots, a_w^\circ = \nu_w  \right] \leq \frac{b}{I_{j,c_j}^{\Delta, [k,j]}}(1 + o(1)),
\end{equation*}
as $q,w,b\to \infty$ and $q = o(w),w = o(b)$. 

Using Lemma \ref{lem:max OC}, we get 
\begin{equation*}
     \cJ_{\theta}(A^{*}, T_{b,A^*}^k) \leq \frac{b}{I_{j,c_j}^{\Delta,[k,j]}}(1+o(1)),
\end{equation*}
as $q,w,b\to \infty$ and $q = o(w),w = o(b)$. 

Since we know that $T_{b,A^*}^{max} = \min_{l\in[p]} T_{b,A^*}^l$, we have 
\begin{equation*}
    \cJ_{\theta}(A^{*}, T_{b,A^*}^{max})\leq \cJ_{\theta}(A^{*}, T_{b,A^*}^k) \leq \frac{b}{I_{j,c_j}^{\Delta,[k,j]}}(1+o(1)),
\end{equation*}
as $q,w,b\to \infty$ and $q = o(w),w = o(b)$. Combining this with Lemma \ref{lem:max IC}, Theorem \ref{thm:general} and setting $b = \log\gamma +\log p$, we complete the proof.

\end{proof}

\section{Causal Structure in Case Study}
\label{app: dag}
\subsection{Case in Ecology}
\label{app:ecology}
\begin{figure}[h]
    \centering
    \includegraphics[width=0.95\linewidth]{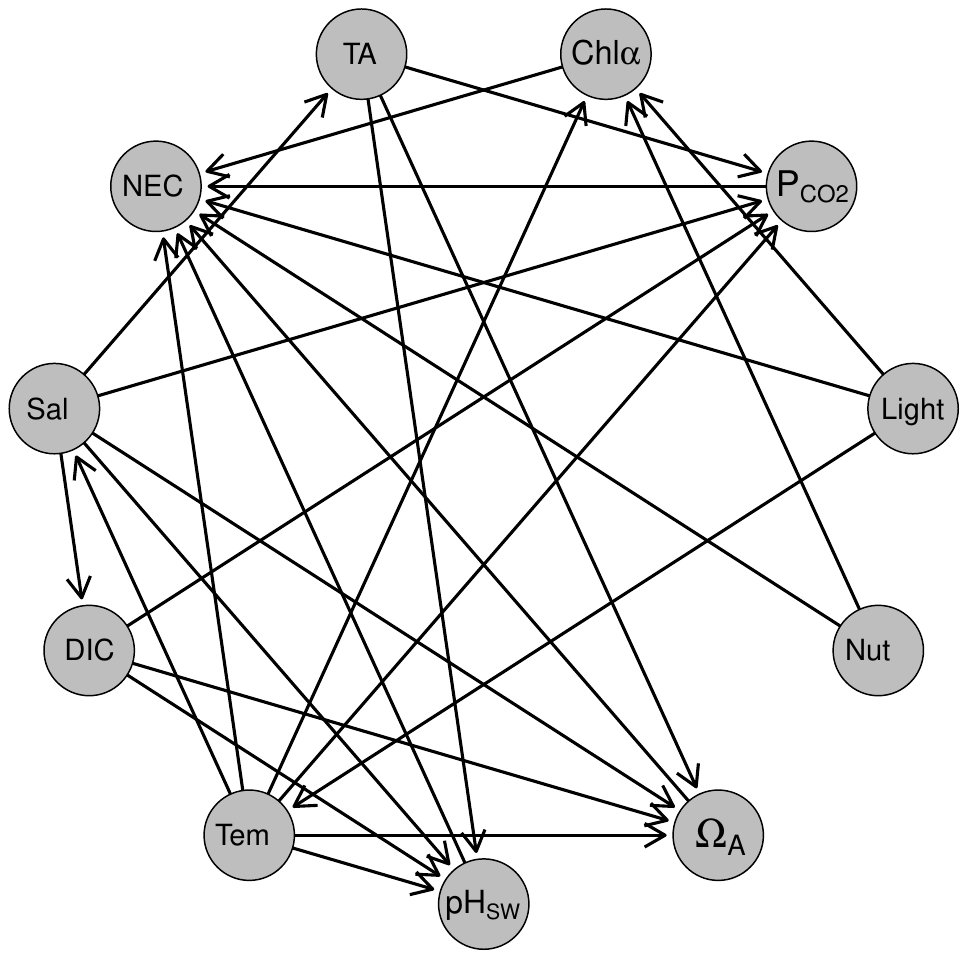}
    \caption{Causal structure of case in ecology.}
    \label{fig:case1dag}
\end{figure}

The DAG describing the causal structure between a set of environmental variables in Section \ref{sec:ecology} is given in Fig \ref{fig:case1dag}.  The variables included in the DAG are
\begin{itemize}
    \item Chl$\alpha$: sea surface chlorophyll a;
    \item Sal: sea surface salinity;
    \item TA: seawater total alkalinity;
    \item DIC: seawater dissolved inorganic carbon;
    \item $P_{CO2}$: seawater $P_{CO2}$;
    \item Tem: bottom temperature;
    \item NEC: net ecosystem calcification;
    \item Light: bottom light levels;
    \item Nut: PC1 of NH4, NiO2+NiO3, SiO4;
    \item $pH_{SW}$ : seawater pH;
    \item $\Omega_A$: seawater saturation with respect to aragonite.
\end{itemize}
See \cite{courtney2017environmental} for more details.

\subsection{Case in Psychology}
\label{app:psychology}
\begin{figure}[h]
    \centering
    \includegraphics[width=0.9\linewidth]{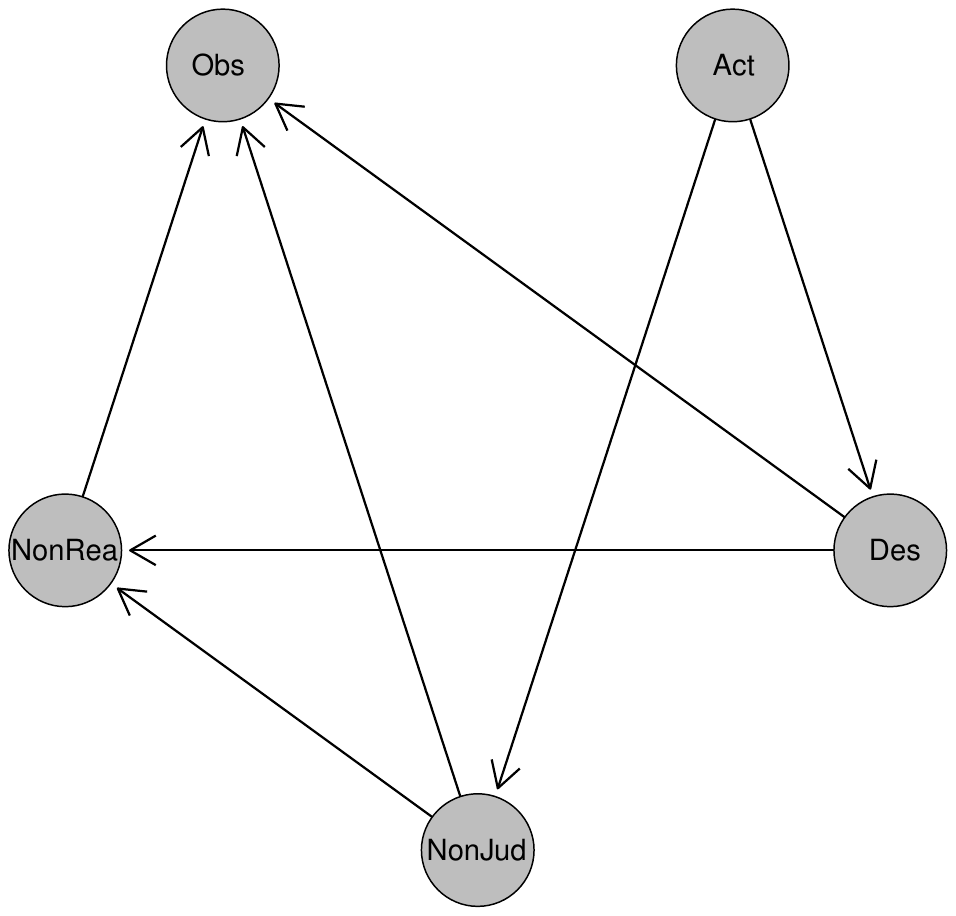}
    \caption{Causal structure of case in psychology.}
    \label{fig:case2dag}
\end{figure}

The DAG describing the causal structure between five factors in Section \ref{sec:psychology} is given in Fig \ref{fig:case2dag}.  The variables included in the DAG are
\begin{itemize}
    \item Obs: Observing;
    \item Act: Acting with awareness;
    \item NonRea: Nonreactivvity;
    \item Des: Describing;
    \item Nonjud: Nonjudging.
\end{itemize}
See \cite{heeren2021network} for more details.

\vskip 0.2in
\bibliography{main-jmlr.bbl}

\end{document}